\documentclass[review]{elsarticle}

\usepackage{lineno}

\usepackage{setspace}

\usepackage{amsmath,amssymb,amsfonts,bm}
\usepackage{caption}
\usepackage{multirow}
\usepackage{array}
\usepackage{subfigure}
\usepackage{booktabs}
\usepackage[ruled]{algorithm2e}

\usepackage[colorlinks, linkcolor=blue, anchorcolor=blue, citecolor=blue]{hyperref}
\usepackage{geometry}
\journal{Journal}

\begin{document}

\captionsetup[figure]{labelfont={bf},labelformat={default},labelsep=period,name={Fig.}}

\begin{frontmatter}

\title{Physics-Informed Deep Monte Carlo Quantile Regression method for Interval Multilevel Bayesian Network-based Satellite Heat Reliability Analysis}

\author[mymainaddress,mysecondaryaddress]{Xiaohu Zheng}

\author[mysecondaryaddress]{Wen Yao\corref{mycorrespondingauthor}}
\cortext[mycorrespondingauthor]{Corresponding author}
\ead{wendy0782@126.com}

\author[mysecondaryaddress]{Zhiqiang Gong\corref{mycorrespondingauthor}}
\ead{gongzhiqiang13@nudt.edu.cn}

\author[mysecondaryaddress]{Yunyang Zhang}

\author[mysecondaryaddress]{Xiaoya Zhang}

\address[mymainaddress]{College of Aerospace Science and Engineering, National University of Defense Technology, No. 109, Deya Road, Changsha 410073, China}
\address[mysecondaryaddress]{Defense Innovation Institute, Chinese Academy of Military Science, No. 53, Fengtai East Street, Beijing 100071, China}

\begin{abstract}
Temperature field reconstruction is essential for analyzing satellite heat reliability. As a representative machine learning model, the deep convolutional neural network (DCNN) is a powerful tool for reconstructing the satellite temperature field. However, DCNN needs a lot of labeled data to learn its parameters, which is contrary to the fact that actual satellite engineering can only acquire noisy unlabeled data. To solve the above problem, this paper proposes an unsupervised method, i.e., the physics-informed deep Monte Carlo quantile regression method, for reconstructing temperature field and quantifying the aleatoric uncertainty caused by data noise. For one thing, the proposed method combines a deep convolutional neural network with the known physics knowledge to reconstruct an accurate temperature field using only monitoring point temperatures. For another thing, the proposed method can quantify the aleatoric uncertainty by the Monte Carlo quantile regression. Based on the reconstructed temperature field and the quantified aleatoric uncertainty, this paper models an interval multilevel Bayesian Network to analyze satellite heat reliability. Two case studies are used to validate the proposed method.
\end{abstract}

\begin{keyword}
Physics-informed convolutional neural network \sep Unsupervised \sep Reliability analysis \sep Quantile regression \sep Aleatoric uncertainty \sep Interval multilevel Bayesian Network
\end{keyword}

\end{frontmatter}

\section{Introduction}\label{sec1}
As an important aircraft in aerospace engineering \cite{Yao2011, Yao2013}, the components' working power of satellite will fluctuate due to some uncertainty factors like the manufacture defects, the different orbital environments, etc. Thereby, it will lead to changes in the temperature fields of some satellite subsystems. If the temperature is too high, it will significantly influence the satellite performance and even cause the satellite to malfunction. Therefore, the heat reliability analysis is essential for analyzing the effects of uncertainty factors on the satellite system performance and minimizing the failure risk \cite{Zheng2019, Zheng2020, Yao2019}.

In actual satellite engineering, only a limited number of temperature monitoring sensors are placed on satellites to save expensive satellite payloads and rare space on circuit boards. However, only engineers who know the temperature at any point on the satellite component can perform an accurate heat reliability analysis. To solve this problem, the surrogate model needs to be built to reconstruct the temperature field of the entire satellite subsystem based on a limited number of monitoring point temperatures. Then, the temperature at any point on the satellite component is obtained according to the reconstructed temperature field. Many methods have been studied to construct the surrogate model in recent years, such as the polynomial chaos expansion method \cite{Oladyshkin2012, Pan2017, Xu2020, Zhang2021}, the Gaussian process regression method \cite{Veiga2019, Cao2021}, the support vector machine method \cite{Okabe2021, Lee2021, Khatibinia2013}, the Kriging method \cite{Zhang2021, Wang2021, Xiao2021}, etc. These methods are suitable for low-dimensional regression problems. However, the satellite temperature field reconstruction (TFR) is a high-dimensional image-to-image regression problem in this paper. As a representative machine learning model, the deep convolutional neural network (DCNN) has drawn a lot of researches in many fields \cite{Li2017, Zhang2020, Zhaoxiaoyu2020, Ronneberger2015}. Due to the convolution operation can reduce the amount of calculation by associating some neurons, DCNN can construct an accurate surrogate model for high-dimensional image-to-image regression problems. Thus, this paper uses DCNN to build a surrogate model for reconstructing the temperature field of the satellite subsystem.

In recent years, plenty of methods has been studied for constructing a surrogate model based on DCNN. The DCNN-based surrogate model construction methods mainly include the ship-ship collision risk classification method\cite{Zhang2020}, the time-frequency information signal prediction method\cite{Cao2021}, the associated remaining useful life estimation method\cite{Li2019}, the accident prediction method of highway-rail grade crossing\cite{Gao2021}, the sensor network modeling method\cite{Li2021}, the uncertainty quantification method of blood flow\cite{Ye2021}, etc. For these methods, the common feature is that they all need a lot of labeled data to learn the parameters of DCNN models. However, due to the high cost of acquiring data labels, some engineering problems only have a small amount of labeled data or even have just some unlabeled data. For satellite engineering, only some monitoring point temperatures can be acquired by a limited number of sensors, and the truth temperature field of the satellite subsystem is hard and expensive to obtain. \textbf{Therefore, one of the difficulties for the satellite TFR problem is constructing an accurate DCNN model by some monitoring point temperatures.}

To overcome the problem of insufficient labeled training data, DCNN combined with the known physical knowledge, such as partial differential equations, boundary and initial conditions, etc., results in a physics-informed DCNN model used to construct surrogate models in different problems. Shen et al.\cite{Shen2021} proposed a physics-informed DCNN model by a simple threshold model for bearing fault detection. For solving partial differential equations, Fang \cite{Fang2021A} developed a hybrid physics-informed DCNN model based on finite volume methods. Besides, Gao et al. \cite{Gao2021PhyGeoNet, Gao2021Super} proposed a novel physics-informed DCNN learning framework to solve partial differential equations on irregular domains. Combined with the topology optimization knowledge, Zhang et al. \cite{Zhang2021TONR} studied a topology optimization via neural reparameterization framework using physics-informed DCNN. In summary, the above problems make full use of the known physical knowledge to reduce the amount of labeled data when DCNN is used to construct a high-precision surrogate model. For the satellite TFR problem, the steady-state temperature field satisfies the Laplace equation and some boundary conditions \cite{Zhaoxiaoyu2020}. Besides, the adjacent temperature value in the steady-state temperature field will not mutate sharply, which means that the steady-state temperature field obeys the TV regularization rule \cite{Rudin1992,Gongzq2021}. \textbf{Based on the known physical knowledge, this paper proposes a physics-informed DCNN model to reconstruct the temperature field of the satellite subsystem by some monitoring point temperatures.}

In actual satellite engineering, the monitoring point temperatures are measured by temperature sensors. However, the possible measure drift of the sensor will lead to data noise existing in the monitoring point temperatures. Thus, when the physics-informed DCNN model reconstructs the temperature field of the satellite subsystem by the monitoring point temperatures with noises, there is aleatoric uncertainty in the reconstructed temperature field. If the aleatoric uncertainty is ignored, the reconstructed temperature field will not be accurate. Sequentially, the final satellite heat reliability analysis results will also be unbelievable. \textbf{Thereby, another difficulty for the satellite TFR problem is quantifying the aleatoric uncertainty caused by the data noise.} Some researchers have studied the aleatoric uncertainty quantification method. Based on a Gaussian maximum-likelihood formulation of a neural network, Nix and Weigend \cite{Nix1994} proposed an architecture for a network with one output unit and one variance unit, where the variance unit is used to quantify aleatoric uncertainty. Based on this method, Kendall and Gal \cite{Kendall2017} studied aleatoric uncertainty quantification for classification Tasks by DCNN. Both the above two approaches are limited by Gaussian assumption \cite{Nix1994}. Combined with quantile regression, Tagasovska and Lopez-Paz \cite{Tagasovska2019} proposed a simultaneous quantile regression (SQR) method to estimate aleatoric uncertainty. However, the SQR method is not suitable for the image-to-image regression problem. \textbf{For the satellite TFR problem, this paper proposes a Monte Carlo quantile regression method to quantify the aleatoric uncertainty caused by data noise based on quantile regression.}

In summary, the proposed physics-informed DCNN model combined with the proposed Deep MC-QR method results in a physics-informed deep Monte Carlo quantile regression (Deep MC-QR) method. The main innovations of the proposed physics-informed Deep MC-QR method are twofold: \textbf{(1) This paper proposes a physics-informed DCNN model to construct a satellite TFR surrogate model using only monitoring point temperatures.} \textbf{(2) This paper proposes a Deep MC-QR method to quantify the aleatoric uncertainty caused by data noise based on the Monte Carlo quantile regression.} In this paper, the parameters involved in the physics-informed Deep MC-QR method are learned by the proposed MC iterative training algorithm. After that, the satellite subsystem temperature field and the corresponding aleatoric uncertainty can be estimated by the proposed Satellite TFR and aleatoric uncertainty quantification algorithm. 

This paper constructs the interval temperature field using the reconstructed satellite subsystem temperature field and the corresponding aleatoric uncertainty to make the satellite heat reliability analysis more believable. According to the work state thresholds of satellite components, the normal probability intervals of satellite components can be obtained by implementing the Monte Carlo simulation on the trained physics-informed Deep MC-QR model. Many methods have been studied for system reliability inference in recent years, such as Bayesian Network (BN) \cite{Zheng2020, Mi2017Reliability, Mi2018, Mi2020}, Universal Generate Function \cite{Levitin2004, Mi2015}, Fault Tree Analysis \cite{Jung2020, Ding2021}, Binary Decision Diagram \cite{Reed2017, Kawahara2019}, etc. This paper chooses BN to infer the satellite heat reliability. \textbf{According to the components' normal probability intervals and the satellite subsystem hierarchical structure, the heat reliability inference model is built to be an interval multilevel BN.} Then, this paper derives the normal probability interval calculation formula for the child node whose logical relationship is series or parallel, based on which the interval multilevel BN can infer the satellite heat reliability.

The rest of this paper is organized as follows. In section \ref{sec2}, the theoretical bases including DCNN and quantile regression are introduced. Then, the physics-informed Deep MC-QR method for satellite TFR is proposed in section \ref{sec3}. Using the reconstructed temperature field and quantified aleatoric uncertainty, the satellite heat reliability analysis method is proposed based on interval multilevel BN in section \ref{sec4}. In section \ref{sec5}, the first case study is used to validate that the proposed physics-informed Deep MC-QR method can accurately reconstruct satellite temperature field and quantify the aleatoric uncertainty, and the proposed methods are applied to the heat reliability analysis for a satellite subsystem in the second case study.

\section{Theoretical bases}\label{sec2}
\subsection{Deep convolutional neural network}\label{sec21}
\paragraph{\textbf{Definition of DCNN}} DCNN is a deep neural network that includes convolutional layers \cite{Goodfellow2016}. The DCNN adopted in this paper uses the two-dimensional convolutional layer. The $l\text{th}$ two-dimensional convolutional layer performs convolution operations on the $l\text{th}$ two-dimensional layer input $\bm{I}^l$ and the convolution kernel $\bm{K}^l$ and adds a scalar bias $b^l$ to obtain the output $\bm{S}^l$, i.e.,
\begin{equation}\label{conv2d}
\begin{aligned}
& \bm{S}^l\left( i,j \right)=\left( \bm{I}^l*\bm{K}^l \right)\left( i,j \right)+b^l \\ 
& \qquad\quad =\sum\limits_{u}{\sum\limits_{v}{\bm{I^l}\left( u,v \right)\bm{K}^l\left( i-u,j-v \right)}}+b^l, \\ 
\end{aligned}
\end{equation}
where $\bm{S}^l\left( i,j \right)$ represents the element in the $i\text{th}$ row and $j\text{th}$ column of $\bm{S}^l$, $*$ denotes the convolution operation. Then, the output $\bm{S}^l$ is input to the nonlinear activation layer to get the activation value $ \bm{H}^l$, i.e.,
\begin{equation}
\bm{H}^l\left( i,j \right)=g\left[\bm{S}^l\left( i,j \right)\right],
\end{equation}
where $g$ is the nonlinear activation function \cite{Nair2010}. In addition, DCNN usually uses a pooling layer (e.g., max-pooling \cite{Zhou1988}) to help the representation of layer input be approximately invariant\cite{Goodfellow2016}.

For a DCNN, the parameters $ \bm \theta=\left(\bm{K}, \bm{b}\right)$ need to be learned by minimizing the loss function $ \mathcal{L}\left(\theta\right) $, i.e.,
\begin{equation}\label{Param_update_DCNN}
\bm{\theta }^* = \arg \min \mathcal{L}\left(\bm{\theta}\right),
\end{equation}
where $ \mathcal{L}\left(\theta\right) $ can be the mean absolute error or the mean square error\cite{Girshick2015}.

\begin{figure*}[!htb]
	\centering
	{\includegraphics[scale=0.4]{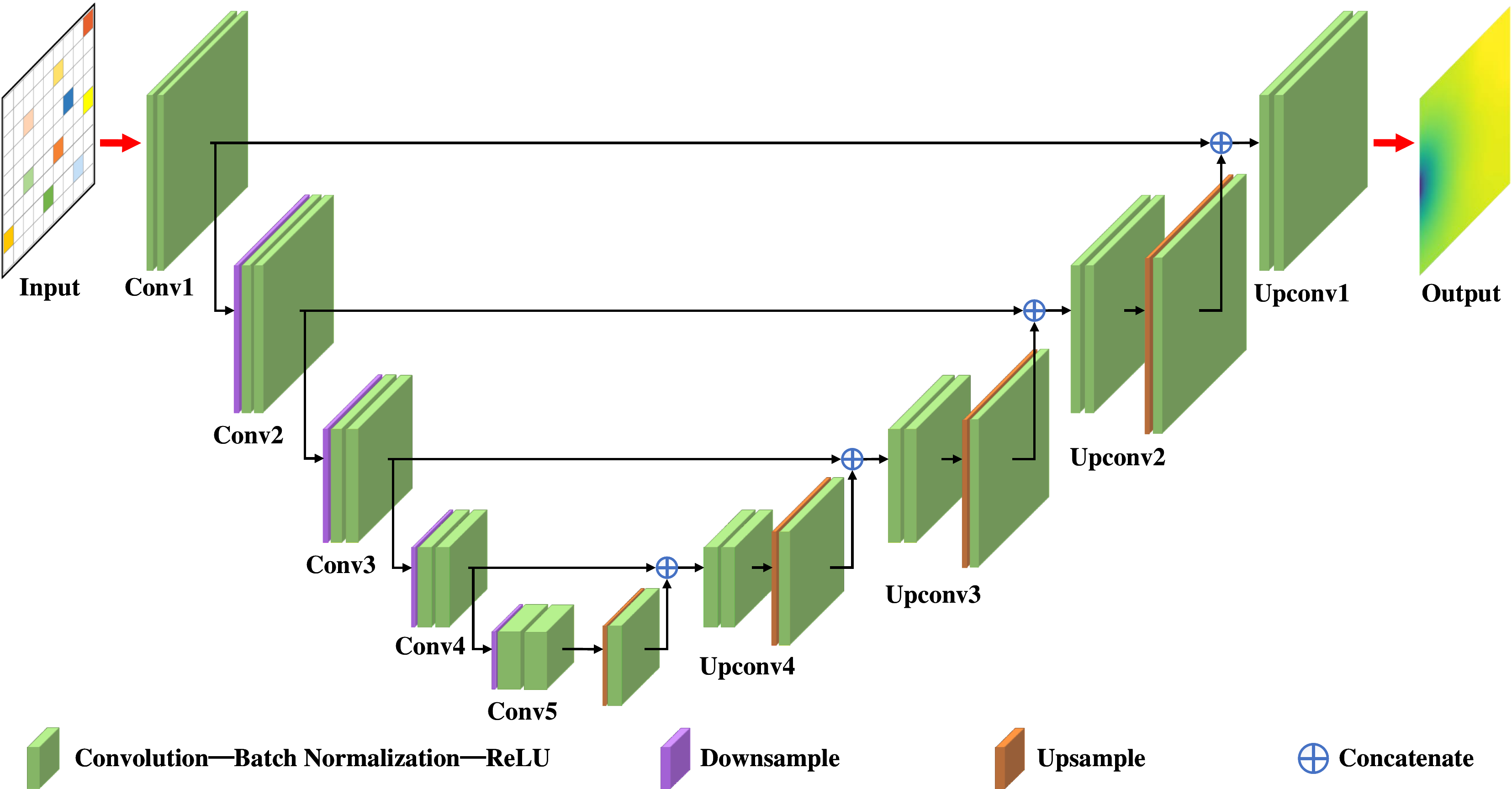}}
	\caption{The architecture of U-net.}\label{U-net}
\end{figure*}

\paragraph{\textbf{U-net}} This paper uses a classical DCNN U-net \cite{Ronneberger2015}, which can process large images and capture the detailed characteristics of the images. The architecture of U-net is shown in Fig.\ref{U-net}, including two parts, i.e., the encoder part (Conv1, Conv2, Conv3, Conv4, Conv5) and the decoder part (Upconv1, Upconv2, Upconv3, Upconv4), where the encoder part extracts the image's feature, and the decoder part gradually samples the feature image to the target output image's size. The green cuboid contains convolution calculation, batch normalization, and ReLU \cite{Nair2010}. The purple cuboid is the downsample achieved by the max-pooling operation \cite{Zhou1988}. The orange cuboid is the upsample implemented by the deconvolution operation \cite{Zeiler2011}. The other detailed information of the U-net can refer to this document \cite{Ronneberger2015}. 

\subsection{Quantile Regression}\label{sec22}
For a stochastic system $W=f\left(\bm{V}\right)$ with the random input variable $\bm{V}$, supposed that the cumulative distribution function of the output $W$ taking the value $w$ is  
\begin{equation}
{{F}_{W|\bm{V}=\bm{v}}}\left( w|\bm{v} \right)=P\left( W\le w \right).
\end{equation}
Thus, the $\tau$ quantile of the output $W$ is
\begin{equation}
{{Q}_{\tau}}\left( W|\bm{V}=\bm{v} \right)=\arg \inf \left\{ w\in \mathbb{R};{{F}_{W|\bm{V}=\bm{v}}}\left( w|\bm{v} \right)>\tau\right\},
\end{equation}
where $ \tau $ ($0\le\tau\le 1$) is the quantile level. If the inverse function of $ {{F}_{W|\bm{V}=\bm{v}}}\left( w|\bm{v} \right) $ is $ F_{W|\bm{V}=\bm{v}}^{-1}\left( \tau \right) $, 
\begin{equation}
{{Q}_{\tau}}\left( W|\bm{V}=\bm{v} \right)=F_{W|\bm{V}=\bm{v}}^{-1}\left(\tau\right).
\end{equation}
For the quantile level $\tau$, the goal of quantile regression is to build a surrogate model $\hat{W}={{\hat{f}}_{\tau}}\left( \bm{V} \right)$ of $W=f\left(\bm{V}\right)$. Based on the training data set $\left\{ \left( {{\bm{v}}_{i}},{{w}_{i}} \right)|i=1,2,\cdots ,N \right\}$, the pinball loss function \cite{Takeuchi2006, Tagasovska2019} was proposed, i.e., 
\begin{equation}\label{pinball_loss}
\begin{aligned}
& {{l}_{\tau}}\left[ {{w}_{i}},{{{\hat{f}}}_{\tau}}\left( {{\bm{v}}_{i}} \right) \right]=\left\{ \begin{matrix}
\tau\left( {{w}_{i}}-{{{\hat{w}}}_{i}} \right), & {{w}_{i}}\ge {{{\hat{w}}}_{i}}  \\
\left( 1-\tau \right)\left( {{{\hat{w}}}_{i}}-{{w}_{i}} \right), & {{w}_{i}}<{{{\hat{w}}}_{i}}  \\
\end{matrix} \right. \\ 
& \mathcal{L}\left[ {{w}_{i}},{{{\hat{f}}}_{\tau}}\left( \bm{v}_i \right) \right]=\frac{1}{N}\sum\limits_{i=1}^{N}{{{l}_{\tau}}\left[ {{w}_{i}},{{{\hat{f}}}_{\tau}}\left( {{\bm{v}}_{i}} \right) \right]}, \\ 
\end{aligned}
\end{equation}
where ${{\hat{w}}_{i}}={{\hat{f}}_{\tau}}\left( {{\bm{v}}_{i}} \right)$ is the prediction corresponding to the input $ \bm{v}_{i} $. By minimizing the loss function, i.e.,
\begin{equation}
{{\hat{f}}_{\tau}}\left( \bm{V} \right)= \arg \min \mathcal{L}\left[w,{{{\hat{f}}}_{\tau}}\left( \bm{v} \right) \right],
\end{equation}
the surrogate model $\hat{W}={{\hat{f}}_{\tau}}\left( \bm{V} \right)$ can be built to approximate $W=f\left(\bm{V}\right)$.

\section{Physics-informed Deep MC-QR method for satellite TFR}\label{sec3}
\subsection{Satellite TFR problem}\label{sec31}
\begin{figure*}[htb]
	\centering
	\subfigure [Satellite subsystem component layout]
	{\includegraphics[scale=0.7]{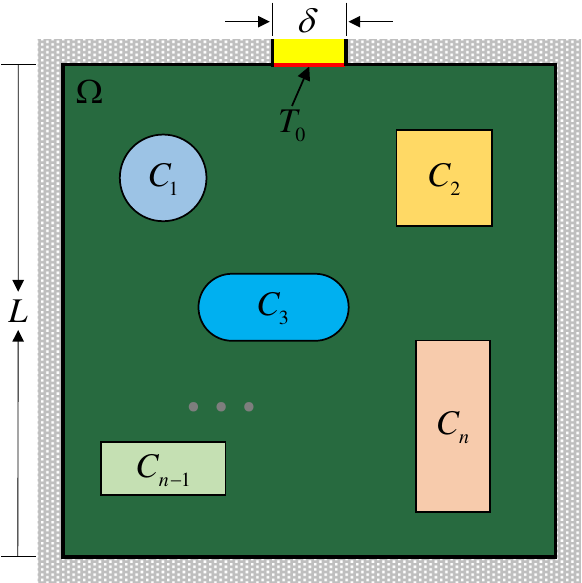}\label{TFR}}
	\hspace{10mm}
	\subfigure [Temperature monitoring point position]
	{\includegraphics[scale=0.7]{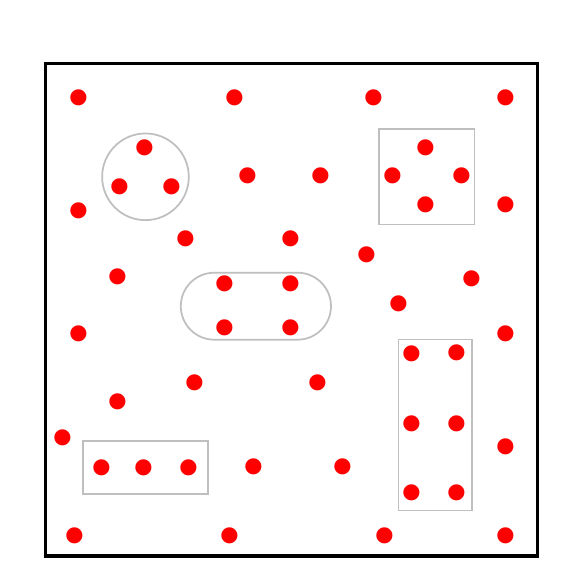}\label{MP}}
	\caption{Simplified modeling of a satellite subsystem with $n$ components in a two-dimensional square area.}\label{TFR_MP}
\end{figure*}

As shown in Fig.\ref{TFR}, a satellite subsystem with $n$ components $\left\{C_s|s=1,2,\cdots,n\right\}$ is simplified into a two-dimensional square area $ \Omega $ with side length $L$, where the powers of $n$ components are $\left\{P_s|s=1,2,\cdots,n\right\}$. The gray grid shaded area is adiabatic. Besides, the yellow rectangle with width $\delta$ at the top is the position of the heat sink. The temperature of each point at the junction (red line, denoted as $ {\Omega }_{BC} $) of square area $ \Omega $ and heat sink is a constant $ T_0 $. In Fig.\ref{MP}, the red points are the positions of temperature monitoring sensors. It is noted that the number of temperature monitoring sensors and their locations need to be determined according to the studied specific satellite subsystem.

For the heat conduction of two-dimensional square area $ \Omega $, its steady state temperature field $ \bm{T} $ meets the Laplace equation \cite{Zhaoxiaoyu2020, Gongzq2021}, i.e.,
\begin{equation}\label{laplace}
\frac{{{\partial }^{2}}\bm{T}\left( x,y \right)}{\partial {{x}^{2}}}+\frac{{{\partial }^{2}}\bm{T}\left( x,y \right)}{\partial {{y}^{2}}}+\bm{\varphi }\left( x,y \right)=0, \quad \left( x,y \right)\in \Omega,
\end{equation}
and the boundary condition
\begin{equation}\label{BC}
\bm{T}\left( x,y \right)={{T}_{0}},\quad \left( x,y \right)\in {{\Omega }_{BC}}
\end{equation}
where $\left( x,y \right)$ denotes the position coordinate and $\bm{\varphi }$ is the heat source intensity distribution determined by the component power. For the satellite TFR problem, the heat source intensity distribution $\bm{\varphi }$ is unknown. Thus, the satellite TFR problem only considers the heat conduction in areas without components. The Laplace equation in Eq.(\ref{laplace}) can be simplified to
\begin{equation}\label{laplace_nmp}
\frac{{{\partial }^{2}}\bm{T}\left( x,y \right)}{\partial {{x}^{2}}}+\frac{{{\partial }^{2}}\bm{T}\left( x,y \right)}{\partial {{y}^{2}}}=0, \quad \left( x,y \right)\in {{\Omega }_{NC}},
\end{equation}
where ${\Omega }_{NC}$ denotes the area without components.

\begin{figure*}[!htb]
	\centering
	{\includegraphics[scale=0.7]{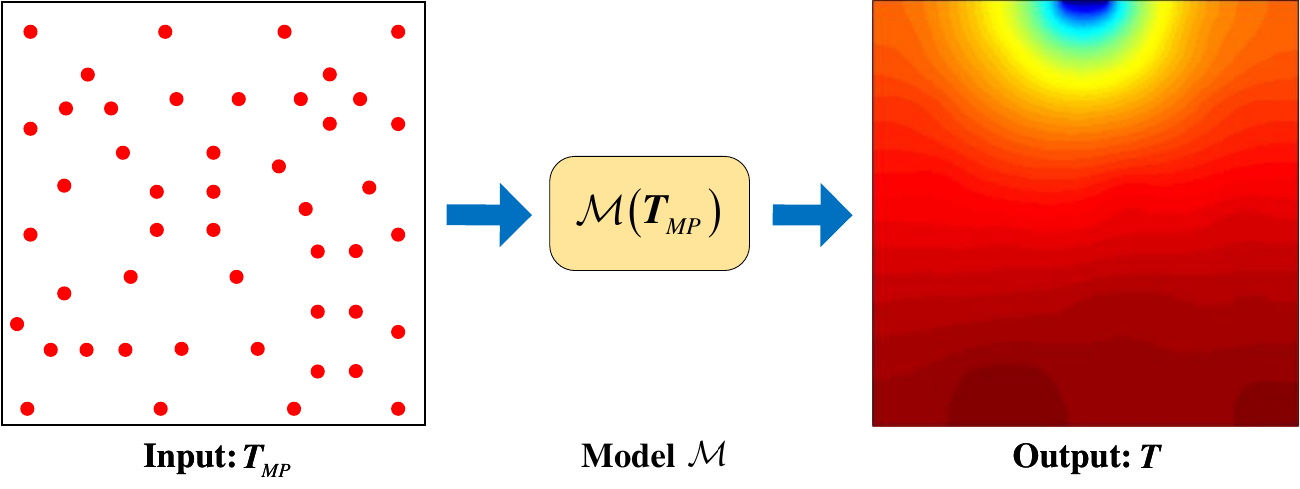}}
	\caption{The satellite TFR problem.}\label{PT_map}
\end{figure*}

As shown in Fig.\ref{PT_map}, the goal of satellite TFR problem is to build a surrogate model $\mathcal{M}\left(\bm{T}_{MP}\right)$ that can reconstruct the temperature field of the satellite subsystem (Fig.\ref{TFR}) by the monitoring point (MP) temperature image $ \bm{T}_{MP} $ (Fig.\ref{MP}). For the MP temperature image $ \bm{T}_{MP} $, the values of the red points are the temperature monitoring sensors' values, and the other points' values are all zero.

\subsection{Physics-informed Deep MC-QR model}\label{sec32}
\begin{figure*}[!htb]
	\centering
	{\includegraphics[scale=0.32]{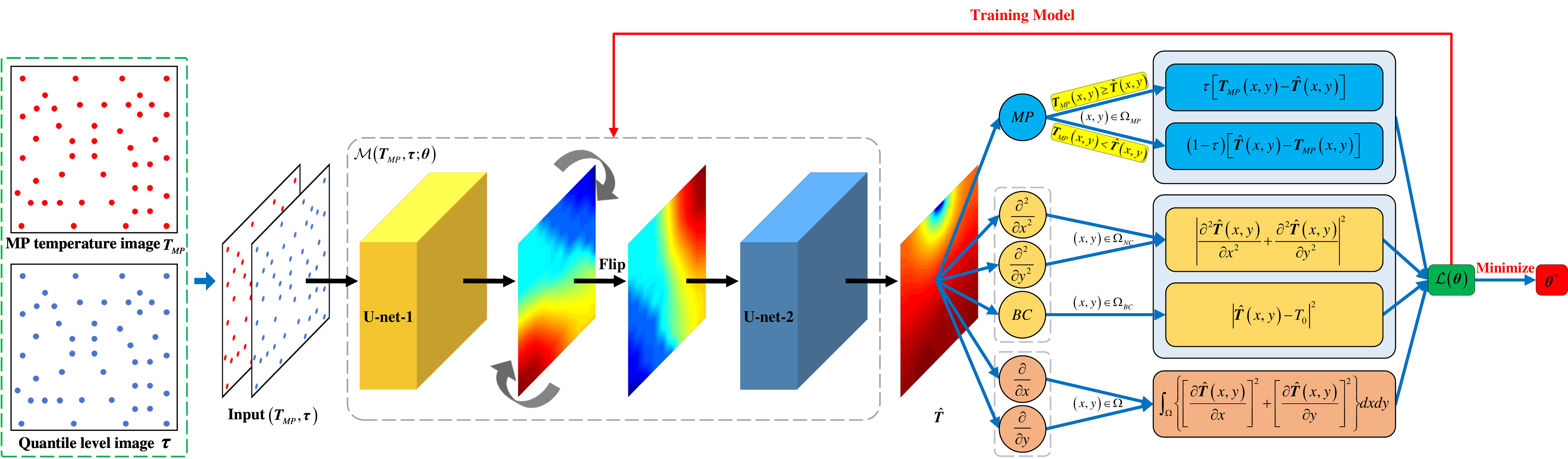}}
	\caption{The schematic of physics-informed Deep Monte Carlo quantile regression model.}\label{MCQR_DCNN}
\end{figure*}

For the satellite TFR problem, this section proposes a physics-informed Deep MC-QR model to reconstruct the temperature field and quantify the aleatoric uncertainty caused by data noise, as shown in Fig.\ref{MCQR_DCNN}. The input $\left( {{\bm{T}}_{MP}},\bm{\tau } \right)$ consists of the MP temperature image $ \bm{T}_{MP} $ and the quantile level image $\bm{\tau}$. In this paper, the MP temperature image $ \bm{T}_{MP} $ is discretized to be a ${{H}_{\Omega }}\times {{W}_{\Omega }}$ two-dimensional array. Besides, the quantile level image ${{H}_{\Omega }}\times {{W}_{\Omega }}$ is also a two-dimensional array $\bm{\tau}$. If $ \left ( x,y \right ) \in {\Omega}_{MP} $, the corresponding position's value $ \bm{\tau}\left(x,y\right) $ is equal to $ \tau $ ($\tau \sim U\left ( 0, 1 \right ) $), i.e.,
\begin{equation}\label{q_image}
\forall \left ( x,y \right ) \in {\Omega}_{MP}, \quad \tau \sim U\left ( 0, 1 \right ), \quad \bm{\tau}\left(x,y\right) = {\tau},
\end{equation}
where $ {\Omega }_{MP} $ denotes the set of temperature monitoring sensors' positions in the discretized temperature image $ \bm{T}_{MP} $. Otherwise, the rest elements of the array $\bm{\tau}$ are equal to zero. As shown in Fig.\ref{MCQR_DCNN}, the values of all blue points in the quantile level image $\bm{\tau}$ are equal to $\tau$, and the other points' values are zero.

In Fig.\ref{MCQR_DCNN}, the first DCNN U-net-1 extracts the feature map of input $\left( {{\bm{T}}_{MP}},\bm{\tau } \right)$. Besides, the feature map is flipped diagonally to ensure the reconstruction accuracy of temperature field in the upside and right-side boundaries \cite{Gongzq2021}. Then, the second DCNN U-net-2 uses the flipped feature map to predict the temperature field $\hat{\bm{T}}$ of the satellite subsystem. Therefore, two U-net models (U-net-1 and U-net-2) and diagonal flip operation build the physics-informed Deep MC-QR model $\mathcal{M}\left(\bm{T}_{MP}, \bm{\tau}; \bm{\theta}\right)$, where $\bm{\theta}$ are the parameters of two U-net models. The training of model $\mathcal{M}\left(\bm{T}_{MP}, \bm{\tau}; \bm{\theta}\right)$ is informed by the quantile MP temperature error, the Laplace equation (Eq.(\ref{laplace_nmp})), the boundary condition (Eq.(\ref{BC})), and the total variation (TV) regularization \cite{Rudin1992, Mahendran2015}. 

Supposed that $N_{MP}$ temperature monitoring sensors are placed at the appropriate positions of satellite subsystem in Fig.\ref{TFR}. Given $N$ MP temperature images $\left\{ \bm{T}^i_{MP}|i=1,2,\cdots,N \right\}$, the parameters $\bm{\theta}$ are learned by minimizing the proposed physics-informed loss function $\mathcal{L}\left(\theta\right)$, consisting of the quantile MP temperature mean error $ \mathcal{L}_{\tau}\left( \bm{\theta} \right) $, the Laplace equation mean squared error $ \mathcal{L}_{LE}\left( \bm{\theta} \right) $, the boundary condition mean squared error $ \mathcal{L}_{BC}\left( \bm{\theta} \right) $ and the TV regularization $ \mathcal{L}_{TV}\left( \bm{\theta} \right) $.
\begin{itemize}
	\item \textbf{Quantile MP temperature mean error $ \mathcal{L}_{\tau}\left( \bm{\theta} \right) $}
\end{itemize}

For the $i\text{th}$ MP temperature image $ \bm{T}^{i}_{MP} $, the predicted temperature field $\hat{\bm{T}}$ of the satellite subsystem is
\begin{equation}\label{M_tau_pre}
{\hat{\bm{T}}_i}=\mathcal{M}\left(\bm{T}^{i}_{MP}, \bm{\tau}_{i}; \bm{\theta}\right).
\end{equation}
For $ \left ( x,y \right ) \in {\Omega}_{MP} $, the error $ {{l}_{{\tau}_i}}\left( x,y;\bm{\theta} \right) $ is
\begin{equation}\label{QMPT_loss_xy}
\begin{aligned}
& {{l}_{{\tau}_i}}\left( x,y;\bm{\theta} \right)=\left\{ \begin{matrix}
{\tau}_i\left[ {{\bm{T}^i_{MP}}\left(x,y\right)}-{{\hat{\bm{T}_i}}\left( x,y \right)} \right], & {{\bm{T}^i_{MP}}\left(x,y\right)}\ge {{\hat{\bm{T}_i}}\left( x,y \right)}  \\
\left( 1-{\tau}_i\right)\left( {{\hat{\bm{T}_i}}\left( x,y \right)}-{{\bm{T}^i_{MP}}\left(x,y\right)} \right), & {{\bm{T}^i_{MP}}\left(x,y\right)}<{{\hat{\bm{T}_i}}\left( x,y \right)}  \\
\end{matrix} \right. , \\
\end{aligned}
\end{equation}
where ${\tau}_i \sim U\left ( 0, 1 \right ) $. For $N$ MP temperature images $\left\{ \bm{T}^i_{MP}|i=1,2,\cdots,N \right\}$, the quantile MP temperature mean error $ \mathcal{L}_{\tau}\left( \bm{\theta} \right) $ is
\begin{equation}\label{QMPT_loss}
\mathcal{L}_{\tau}\left( \bm{\theta} \right)=\frac{1}{N}\frac{1}{N_{MP}}\sum\limits_{i=1}^{N}{\sum_{\left ( x,y \right ) \in {\Omega}_{MP}}  {{{l}_{{\tau}_i}}\left( x,y;\bm{\theta} \right)}}.
\end{equation}

\begin{itemize}
	\item \textbf{Laplace equation mean squared error $ \mathcal{L}_{LE}\left( \bm{\theta} \right) $}
\end{itemize}

Suppose that each discretized MP temperature image $ \bm{T}^{i}_{MP} $ has $N_{NC}$ points satisfying the condition $ \left( x,y \right)\in {{\Omega }_{NC}} $. According to Eq.(\ref{laplace_nmp}), the Laplace equation mean squared error $ \mathcal{L}_{LE}\left( \bm{\theta} \right) $ for $N$ MP temperature images $\left\{ \bm{T}^i_{MP}|i=1,2,\cdots,N \right\}$ is
\begin{equation}\label{LE_loss}
{{\mathcal{L}}_{LE}}\left( \bm{\theta } \right)=\frac{1}{N}\frac{1}{{{N}_{NC}}}\sum\limits_{i=1}^{N}{\sum\limits_{\left( x,y \right)\in {{\Omega }_{NC}}}{{{\left| \frac{{{\partial }^{2}}\hat{\bm{T}}_i\left( x,y \right)}{\partial {{x}^{2}}}+\frac{{{\partial }^{2}}\hat{\bm{T}}_i\left( x,y \right)}{\partial {{y}^{2}}} \right|}^{2}}}}.
\end{equation}
By discretizing the Laplace equation Eq.(\ref{laplace_nmp}), the first part $ {{\partial }^{2}}\hat{\bm{T}}_i\left( x_j,y_k \right)/{\partial {{x}_j^{2}}} $ for position coordinate $\left(x_j,y_k\right)$ can be
\begin{equation}\label{le_X_discre}
\frac{{{\partial }^{2}}\hat{\bm{T}}_i\left( x_j,y_k \right)}{\partial {{x}_j^{2}}}=\frac{{\left[ \hat{\bm{T}}_i\left( {{x}_{j+1}},{{y}_{k}} \right)-\hat{\bm{T}}_i\left( {{x}_{j}},{{y}_{k}} \right) \right]}/{\left( {{x}_{j+1}}-{{x}_{j}} \right)-{\left[ \hat{\bm{T}}_i\left( {{x}_{j}},{{y}_{k}} \right)-\hat{\bm{T}}_i\left( {{x}_{j-1}},{{y}_{k}} \right) \right]}/{\left( {{x}_{j}}-{{x}_{j-1}} \right)}\;}\;}{{{x}_{j}}-{{x}_{j-1}}}.
\end{equation}
Denoted that $ \Delta {{x}_{j}}={{x}_{j}}-{{x}_{j-1}} $ and $ \Delta {{\hat{\bm{T}}}^i_{j}}=\hat{\bm{T}}_i\left( {{x}_{j}},{{y}_{k}} \right)-\hat{\bm{T}}_i\left( {{x}_{j-1}},{{y}_{k}} \right) $. Then, Eq.(\ref{le_X_discre}) can be
\begin{equation}\label{le_X_discre_simple}
\begin{aligned}
& \frac{{{\partial }^{2}}\hat{\bm{T}}_i\left( x_j,y_k \right)}{\partial {{x}_j^{2}}}=\frac{{\Delta {{{\hat{\bm{T}}}}^i_{j+1}}}/{\Delta {{x}_{j+1}}-{\Delta {{{\hat{\bm{T}}}}^i_{j}}}/{\Delta {{x}_{j}}}\;}\;}{\Delta {{x}_{j}}} \\ 
& \qquad\qquad\quad\;\,=\frac{\Delta {{{\hat{\bm{T}}}}^i_{j+1}}\Delta {{x}_{j}}-\Delta {{{\hat{\bm{T}}}}^i_{j}}\Delta {{x}_{j+1}}}{\Delta {{x}_{j+1}}{{\Delta }^{2}}{{x}_{j}}}. \\ 
\end{aligned}
\end{equation}
Like Eq.(\ref{le_X_discre_simple}), the second part $ {{\partial }^{2}}\hat{\bm{T}}_i\left( x_j,y_k \right)/{\partial {{y}_k^{2}}} $ of Eq.(\ref{laplace_nmp}) can be
\begin{equation}\label{le_Y_discre_simple}
\begin{aligned}
& \frac{{{\partial }^{2}}\hat{\bm{T}}_i\left( {{x}_{j}},{{y}_{k}} \right)}{y_{k}^{2}}=\frac{{\Delta {{{\hat{\bm{T}}}}^i_{k+1}}}/{\Delta {{y}_{k+1}}-{\Delta {{{\hat{\bm{T}}}}^i_{k}}}/{\Delta {{y}_{k}}}\;}\;}{\Delta {{y}_{k}}} \\ 
& \qquad\qquad\quad\;\,=\frac{\Delta {{{\hat{\bm{T}}}}^i_{k+1}}\Delta {{y}_{k}}-\Delta {{{\hat{\bm{T}}}}^i_{k}}\Delta {{y}_{k+1}}}{\Delta {{y}_{k+1}}{{\Delta }^{2}}{{y}_{k}}}, \\ 
\end{aligned}
\end{equation}
where $ \Delta {{y}_{k}}={{y}_{k}}-{{y}_{k-1}} $ and $ \Delta {{\hat{\bm{T}}}^i_{k}}=\hat{\bm{T}}_i\left( {{x}_{j}},{{y}_{k}} \right)-\hat{\bm{T}}_i\left( {{x}_{j}},{{y}_{k-1}} \right) $. Refer to Eqs.(\ref{le_X_discre_simple}) and (\ref{le_Y_discre_simple}), Eq.(\ref{LE_loss}) is further transformed into
\begin{equation}\label{LE_loss_simple}
{{\mathcal{L}}_{LE}}\left( \bm{\theta } \right)=\frac{1}{N}\frac{1}{{{N}_{NC}}}\sum\limits_{i=1}^{N}{\sum\limits_{\left( x,y \right)\in {{\Omega }_{NC}}}{{{\left| \frac{\Delta {{{\hat{\bm{T}}}}^i_{j+1}}\Delta {{x}_{j}}-\Delta {{{\hat{\bm{T}}}}^i_{j}}\Delta {{x}_{j+1}}}{\Delta {{x}_{j+1}}{{\Delta }^{2}}{{x}_{j}}}+\frac{\Delta {{{\hat{\bm{T}}}}^i_{k+1}}\Delta {{y}_{k}}-\Delta {{{\hat{\bm{T}}}}^i_{k}}\Delta {{y}_{k+1}}}{\Delta {{y}_{k+1}}{{\Delta }^{2}}{{y}_{k}}} \right|}^{2}}}}.
\end{equation}

\begin{itemize}
	\item \textbf{Boundary condition mean squared error $ \mathcal{L}_{BC}\left( \bm{\theta} \right) $}
\end{itemize}

Assume that there are $N_{BC}$ points $\left(x,y\right)$ in the area ${{\Omega }_{BC}}$ of each discretized  MP temperature image $ \bm{T}^{i}_{MP} $. Refer to Eq.(\ref{BC}), the boundary condition mean squared error $ \mathcal{L}_{BC}\left( \bm{\theta} \right) $ for $N$ MP temperature images $\left\{ \bm{T}^i_{MP}|i=1,2,\cdots,N \right\}$ is
\begin{equation}\label{bc_loss}
{{\mathcal{L}}_{BC}}\left( \bm{\theta } \right)=\frac{1}{N}\frac{1}{{{N}_{BC}}}\sum\limits_{i=1}^{N}{\sum\limits_{\left( x,y \right)\in {{\Omega }_{BC}}}{{{\left| {{{\hat{\bm{T}}}}_{i}}\left( x,y \right)-{{T}_{0}} \right|}^{2}}}}.
\end{equation}

\begin{itemize}
	\item \textbf{TV regularization $ \mathcal{L}_{TV}\left( \bm{\theta} \right) $}
\end{itemize}

TV regularization \cite{Rudin1992,Gongzq2021}, as shown in Eq.(\ref{tv}), can maintain the smoothness of the image, 
\begin{equation}\label{tv}
{{\mathcal{R}}_{{{V}^{\beta }}}}=\int_{\Omega }{{{\left\{ {{\left[ \frac{\partial f\left( u,v \right)}{\partial u} \right]}^{2}}+{{\left[ \frac{\partial f\left( u,v \right)}{\partial v} \right]}^{2}} \right\}}^{\frac{\beta }{2}}}}dudv,
\end{equation}
where $ f\left(u,v\right) $ is a continuous function. Based on the property that the steady-state temperature field of the satellite subsystem will not mutate sharply, this paper adopts TV regularization to assist the training of model $\mathcal{M}\left(\bm{T}_{MP}, \bm{\tau}; \bm{\theta}\right)$. By the finite-difference approximation, the TV regularization $ \mathcal{L}_{TV}\left( \bm{\theta} \right) $ ($\beta=2$) for $N$ discretized MP temperature images $\left\{ \bm{T}^i_{MP}|i=1,2,\cdots,N \right\}$ is,
\begin{equation}\label{tv_loss}
\begin{aligned}
& \mathcal{L}_{TV}^{x}\left( \bm{\theta } \right)=\frac{1}{{{H}_{\Omega }}\times \left( {{W}_{\Omega }}-1 \right)}\sum\limits_{k=1}^{{{H}_{\Omega }}}{\sum\limits_{j=1}^{{{W}_{\Omega }}-1}{{{\left[ {{{\bm{\hat{T}}}}_{i}}\left( {{x}_{j+1}},{{y}_{k}} \right)-{{{\bm{\hat{T}}}}_{i}}\left( {{x}_{j}},{{y}_{k}} \right) \right]}^{2}}}}, \\ 
& \mathcal{L}_{TV}^{y}\left( \bm{\theta } \right)=\frac{1}{\left( {{H}_{\Omega }}-1 \right)\times {{W}_{\Omega }}}\sum\limits_{j=1}^{{{W}_{\Omega }}}{\sum\limits_{k=1}^{{{H}_{\Omega }}-1}{{{\left[ {{{\bm{\hat{T}}}}_{i}}\left( {{x}_{j}},{{y}_{k+1}} \right)-{{{\bm{\hat{T}}}}_{i}}\left( {{x}_{j}},{{y}_{k}} \right) \right]}^{2}}}}, \\ 
& {{\mathcal{L}}_{TV}}\left( \bm{\theta } \right)=\frac{1}{N}\sum\limits_{i=1}^{N}{\left[ \mathcal{L}_{TV}^{x}\left( \bm{\theta } \right)+\mathcal{L}_{TV}^{y}\left( \bm{\theta } \right) \right]}. \\ 
\end{aligned}
\end{equation}

Based on the quantile MP temperature mean error $ \mathcal{L}_{\tau}\left( \bm{\theta} \right) $, the Laplace equation mean squared error $ \mathcal{L}_{LE}\left( \bm{\theta} \right) $, the boundary condition mean squared error $ \mathcal{L}_{BC}\left( \bm{\theta} \right) $, and the TV regularization $ \mathcal{L}_{TV}\left( \bm{\theta} \right) $, the proposed physics-informed loss function $\mathcal{L}\left(\theta\right)$ for learning the parameters $\bm{\theta}$ of model $\mathcal{M}\left(\bm{T}_{MP}, \bm{\tau}; \bm{\theta}\right)$ is
\begin{equation}\label{PI_Loss}
\mathcal{L}\left( \bm{\theta } \right)={\alpha }_{\tau}{{\mathcal{L}}_{\tau }}\left( \bm{\theta } \right)+{{\alpha }_{LE}}{{\mathcal{L}}_{LE}}\left( \bm{\theta } \right)+{{\alpha }_{BC}}{{\mathcal{L}}_{BC}}\left( \bm{\theta } \right)+{{\alpha }_{TV}}{{\mathcal{L}}_{TV}}\left( \bm{\theta } \right),
\end{equation}
where $ {\alpha }_{\tau} $, $ {\alpha }_{LE} $, $ {\alpha }_{BC} $, and $ {\alpha }_{TV} $ are hyperparameters.

\subsection{Physics-informed Deep MC-QR model MC iterative training algorithm}\label{sec33}
\begin{figure*}[!htb]
	\centering
	{\includegraphics[scale=0.7]{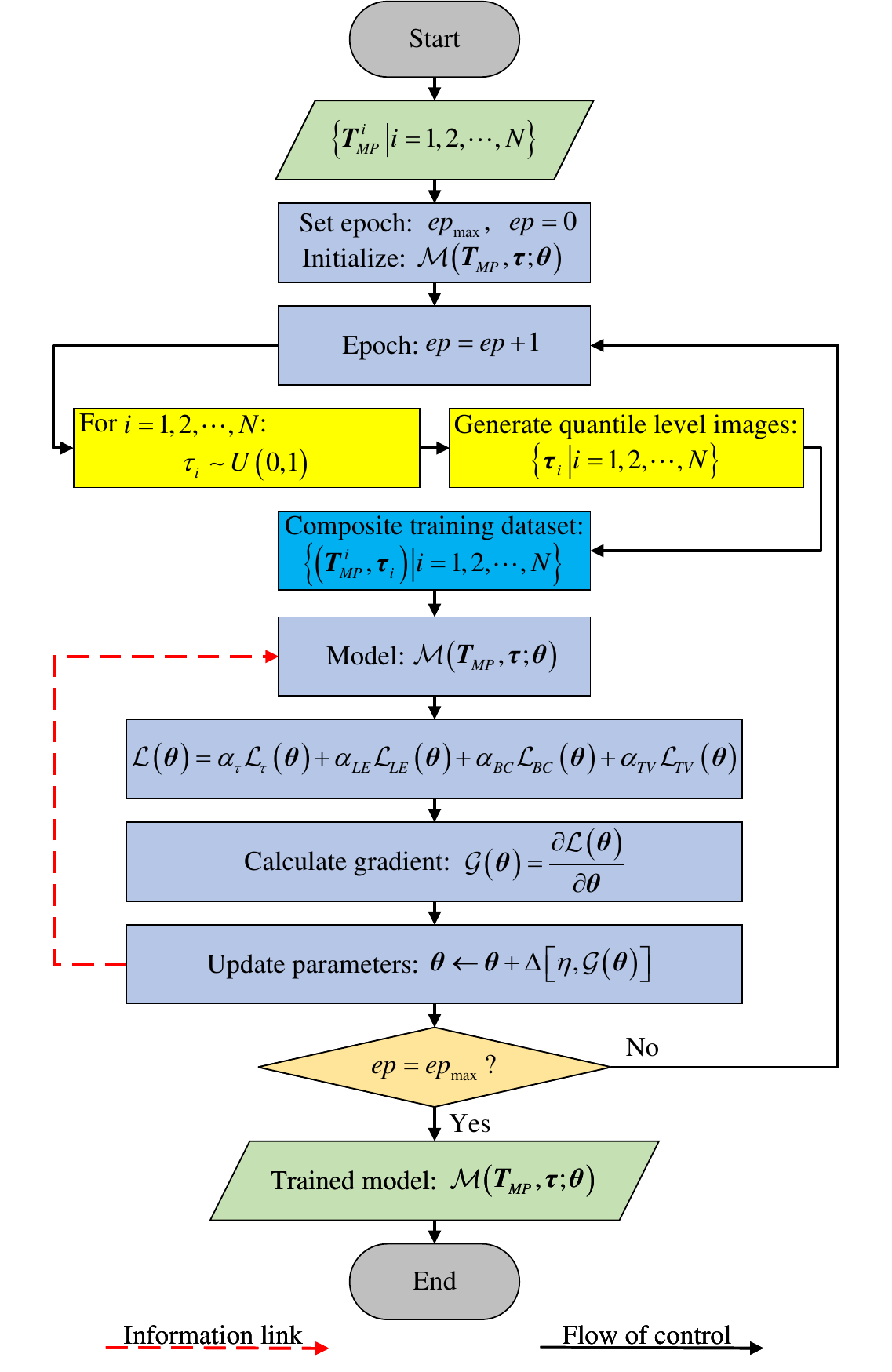}}
	\caption{The training flowchart of MC iterative training algorithm}\label{model_train}
\end{figure*}

This section proposes a MC iterative training algorithm for learning the parameters $\bm{\theta}$ of physics-informed Deep MC-QR model $\mathcal{M}\left(\bm{T}_{MP}, \bm{\tau}; \bm{\theta}\right)$, and its flowchart is shown in Fig.\ref{model_train}. 

Based on $N$ discretized MP temperature images $\left\{ \bm{T}^i_{MP}|i=1,2,\cdots,N \right\}$, the corresponding quantile level images $\left\{ {{\bm{\tau }}_{i}}\left| i=1,2,\cdots ,N \right. \right\}$ can be obtained by Eq.(\ref{q_image}) for the $ep\text{th}$ ($ ep=1,2,\cdots ,ep_{max} $) epoch. For the quantile level image $ {{\bm{\tau }}_{i}} $ corresponding to the $i\text{th}$ discretized MP temperature image, the quantile level ${\tau}_i$ is randomly sampled from the uniform distribution $U\left ( 0, 1 \right )$ in each model training epoch. The training dataset $\left\{ \left( \bm{T}_{MP}^{i},{{\bm{\tau }}_{i}} \right)\left| i=1,2,\cdots ,N \right. \right\}$ is composited using the discretized MP temperature images $\left\{ \bm{T}^i_{MP}|i=1,2,\cdots,N \right\}$ and the quantile level images $\left\{ {{\bm{\tau }}_{i}}\left| i=1,2,\cdots ,N \right. \right\}$. Then, the parameters $\bm{\theta}$ of model $\mathcal{M}\left(\bm{T}_{MP}, \bm{\tau}; \bm{\theta}\right)$ are learned by minimizing the proposed physics-informed loss function $\mathcal{L}\left(\theta\right)$, i.e., 
\begin{equation}\label{arg_min}
{\bm{\theta }}=\underset{\bm{\theta }}{\mathop{\arg }}\,\min \left[ {\alpha }_{\tau}{{\mathcal{L}}_{\tau }}\left( \bm{\theta } \right)+{{\alpha }_{LE}}{{\mathcal{L}}_{LE}}\left( \bm{\theta } \right)+{{\alpha }_{BC}}{{\mathcal{L}}_{BC}}\left( \bm{\theta } \right)+{{\alpha }_{TV}}{{\mathcal{L}}_{TV}}\left( \bm{\theta } \right) \right].
\end{equation}

The proposed MC iterative training algorithm uses the Adam algorithm \cite{Kingma2014} to iteratively solve Eq.(\ref{arg_min}). Based on the composited training dataset $\left\{ \left( \bm{T}_{MP}^{i},{{\bm{\tau }}_{i}} \right)\left| i=1,2,\cdots ,N \right. \right\}$, the proposed physics-informed loss function $\mathcal{L}\left(\bm{\theta}\right)$ is calculated by Eq.(\ref{PI_Loss}). Therefore, the gradient $ \mathcal{G}\left(\bm{\theta}\right) $ is derived by the chain rule for differentiating compositions of functions using automatic differentiation \cite{Baydin2015}, i.e.,
\begin{equation}\label{Gradient}
\mathcal{G}\left( \bm{\theta } \right)=\frac{\partial \mathcal{L}\left( \bm{\theta } \right)}{\partial \left( \bm{\theta } \right)}.
\end{equation}
In the training process, the parameters $ \bm{\theta} $ is updated iteratively as follows:
\begin{equation}\label{update_parm}
\bm{\theta }\leftarrow \bm{\theta }+\Delta \left[ \eta ,\mathcal{G}\left( \bm{\theta } \right) \right],
\end{equation}
where $ \eta $ is the learning rate, and $ \Delta \left( \cdot  \right) $ is the calculation update operator determined by the Adam algorithm \cite{Kingma2014}. In summary, the pseudo code for training the physics-informed Deep MC-QR model $\mathcal{M}\left(\bm{T}_{MP}, \bm{\tau}; \bm{\theta}\right)$ is shown in \textbf{Algorithm} \textbf{\ref{algorithm1}}. In particular, different from the existing DCNN training methods, the key step of the MC iterative training algorithm is that the quantile level images $\left\{ {{\bm{\tau }}_{i}}\left| i=1,2,\cdots ,N \right. \right\}$ are randomly generated by Eq.(\ref{q_image}) in each training epoch.

\begin{algorithm}[!t]
	\caption{Physics-informed Deep MC-QR model MC iterative training algorithm.}
	\label{algorithm1}
	\LinesNumbered
	\KwIn{\\ 
		\qquad (1) Learning rate $\eta$; \\
		\qquad (2) Maximum training epoch $ ep_{max} $; \\
		\qquad (3) Hyperparameters: $ {\alpha }_{\tau} $, $ {\alpha }_{LE} $, $ {\alpha }_{BC} $, $ {\alpha }_{TV} $; \\
		\qquad (4) $N$ discretized MP temperature images $\left\{ \bm{T}^i_{MP}|i=1,2,\cdots,N \right\}$.
	}
	\KwOut{\\ \qquad Trained physics-informed Deep MC-QR model $\mathcal{M}\left(\bm{T}_{MP}, \bm{\tau}; \bm{\theta}\right)$.}

	Initialize the physics-informed Deep MC-QR model $\mathcal{M}\left(\bm{T}_{MP}, \bm{\tau}; \bm{\theta}\right)$; \\ 
	\For{$ep=1:ep_{max}$}{
		\For{$i=1:N$}{
			Randomly take a sample ${\tau}_i$ from uniform distribution $U\left ( 0, 1 \right )$; \\
			Create a ${{H}_{\Omega }}\times {{W}_{\Omega }}$ two-dimensional array $\bm{\tau}_i$ with all zero elements; \\
			Perform $ \bm{\tau}_i\left(x,y\right) = {\tau}_i$ for $ \forall \left( x,y \right) \in {{\Omega}_{MP}} $; \\
			Composite training data $ \left( \bm{T}_{MP}^{i},{{\bm{\tau }}_{i}} \right) $;\\
			Estimate the reconstructed temperature field ${{\hat{\bm{T}}}_{i}}=\mathcal{M}\left(\bm{T}^i_{MP}, \bm{\tau}_i; \bm{\theta}\right)$.
		}
		Calculate the quantile MP temperature mean error $ \mathcal{L}_{\tau}\left( \bm{\theta} \right) $; \\
		Calculate the Laplace equation mean squared error $ \mathcal{L}_{LE}\left( \bm{\theta} \right) $; \\
		Calculate the boundary condition mean squared error $ \mathcal{L}_{BC}\left( \bm{\theta} \right) $; \\
		Calculate the TV regularization $ \mathcal{L}_{TV}\left( \bm{\theta} \right) $; \\
		Calculate the proposed physics-informed loss function $\mathcal{L}\left( \bm{\theta } \right)={\alpha }_{\tau}{{\mathcal{L}}_{\tau }}\left( \bm{\theta } \right)+{{\alpha }_{LE}}{{\mathcal{L}}_{LE}}\left( \bm{\theta } \right)+{{\alpha }_{BC}}{{\mathcal{L}}_{BC}}\left( \bm{\theta } \right)+{{\alpha }_{TV}}{{\mathcal{L}}_{TV}}\left( \bm{\theta } \right)$; \\
		Calculate the gradient $\mathcal{G}\left( \bm{\theta } \right)$ of the cost function $\mathcal{J}\left( \bm{\theta } \right)$; \\
		Calculate the calculation update operator $\Delta \left[ \eta ,\mathcal{G}\left( \bm{\theta } \right) \right]$; \\
		Update the paramters $\bm{\theta }\leftarrow \bm{\theta }+\Delta \left[ \eta ,\mathcal{G}\left( \bm{\theta } \right) \right]$.
	}
\end{algorithm}

\subsection{Satellite TFR and aleatoric uncertainty quantification algorithm}\label{sec34}
Refer to section \ref{sec33}, the trained physics-informed Deep MC-QR model $\mathcal{M}\left(\bm{T}_{MP}, \bm{\tau}; \bm{\theta}\right)$ can be obtained by the proposed MC iterative training algorithm. For the MP temperature image $ \bm{T}^{pre}_{MP} $, $ N_{pre} $ quantile levels $\left\{ {\tau}_{\mathcal{P} }^{pre}|\mathcal{P}=1,2,\cdots,N_{pre} \right\}$ are sampled from uniform distribution $U\left(0,1\right)$. Then, $ N_{pre} $ quantile level images $\left\{ \bm{\tau}_{\mathcal{P} }^{pre}|\mathcal{P}=1,2,\cdots,N_{pre} \right\}$ are generated by Eq.(\ref{q_image}), based  on which the prediction dataset $ \mathcal{D}_{pre}  $, i.e.,
\begin{equation}
\mathcal{D}_{pre}=\left\{ \left(\bm{T}^{pre}_{MP},\bm{\tau}_{\mathcal{P} }^{pre}\right)|\mathcal{P}=1,2,\cdots,N_{pre} \right\}
\end{equation}
are composited for the MP temperature image $ \bm{T}^{pre}_{MP} $. For $ \mathcal{P}=1,2,\cdots,N_{pre} $, the temperature field $ \hat{\bm{T}}_{\mathcal{P} }^{pre} $ is predicted by the trained physics-informed Deep MC-QR model $\mathcal{M}\left(\bm{T}_{MP}, \bm{\tau}; \bm{\theta}\right)$, i.e., $\hat{\bm{T}}_{\mathcal{P} }^{pre}=\mathcal{M}\left(\bm{T}^{pre}_{MP}, \bm{\tau}_{\mathcal{P} }^{pre}; \bm{\theta}\right)$. Thereby, the reconstructed temperature field's approximation $\hat{\bm{T}}_{pre}$ is predicted by calculating the mean of $ N_{pre} $ results $ \hat{\bm{T}}_{\mathcal{P} }^{pre} $ ($\mathcal{P}=1,2,\cdots,N_{pre}$), i.e.,
\begin{equation}\label{T_pre}
{{\hat{\bm{T}}}_{pre}}=\frac{1}{N_{pre}}\sum\limits_{\mathcal{P}=1}^{{{N}_{pre}}}{\hat{\bm{T}}_{\mathcal{P}}^{pre}}=\frac{1}{N_{pre}}\sum\limits_{\mathcal{P}=1}^{{{N}_{pre}}}{\mathcal{M}\left( \bm{T}_{MP}^{pre},\bm{\tau }_{\mathcal{P}}^{pre};\bm{\theta } \right)}.
\end{equation}
Besides, the aleatoric uncertainty $\bm{\sigma}_{pre}$ of the reconstructed temperature field $\hat{\bm{T}}_{pre}$ can be quantified by calculating the standard deviation of $ N_{pre} $ results $ \hat{\bm{T}}_{\mathcal{P} }^{pre} $ ($\mathcal{P}=1,2,\cdots,N_{pre}$), i.e.,
\begin{equation}\label{U_alea}
\begin{aligned}
& {{\bm{\sigma }}_{pre}}={{\left[ \frac{1}{N_{pre}}\sum\limits_{\mathcal{P}=1}^{{{N}_{pre}}}{{{\left( \hat{\bm{T}}_{\mathcal{P}}^{pre} \right)}^{2}}-{{\left( {{{\hat{\bm{T}}}}_{pre}} \right)}^{2}}} \right]}^{\frac{1}{2}}} \\ 
& \quad\;\,\,\, ={{\left\{ \frac{1}{N_{pre}}{{\sum\limits_{\mathcal{P}=1}^{{{N}_{pre}}}{{{\left[ \mathcal{M}\left( \bm{T}_{MP}^{pre},\bm{\tau }_{\mathcal{P}}^{pre};\bm{\theta } \right) \right]}^{2}}-\left[ \frac{1}{N_{pre}}\sum\limits_{\mathcal{P}=1}^{{{N}_{pre}}}{\mathcal{M}\left( \bm{T}_{MP}^{pre},\bm{\tau }_{\mathcal{P}}^{pre};\bm{\theta } \right)} \right]}}^{2}} \right\}}^{\frac{1}{2}}} \\ 
& \quad\;\,\,\,=\frac{1}{{{N}_{pre}^{2}}}{{\left\{ {{\sum\limits_{\mathcal{P}=1}^{{{N}_{pre}}}{{{\left[ \mathcal{M}\left( \bm{T}_{MP}^{pre},\bm{\tau }_{\mathcal{P}}^{pre};\bm{\theta } \right) \right]}^{2}}-\frac{1}{N_{pre}}\left[ \sum\limits_{\mathcal{P}=1}^{{{N}_{pre}}}{\mathcal{M}\left( \bm{T}_{MP}^{pre},\bm{\tau }_{\mathcal{P}}^{pre};\bm{\theta } \right)} \right]}}^{2}} \right\}}^{\frac{1}{2}}}. \\ 
\end{aligned}
\end{equation}

In summary, the pseudo code for predicting the reconstructed satellite subsystem temperature field and quantifying the aleatoric uncertainty is shown in the following \textbf{Algorithm} \textbf{\ref{algorithm2}}.

\begin{algorithm}[!htb]
	\caption{Satellite TFR and aleatoric uncertainty quantification algorithm.}
	\label{algorithm2}
	\LinesNumbered
	\KwIn{\\ 
		\qquad (1) MP temperature image $\bm{T}^{pre}_{MP}$; \\
		\qquad (2) The number $ N_{pre} $ of quantile levels; \\
		\qquad (3) Trained physics-informed Deep MC-QR model $\mathcal{M}\left(\bm{T}_{MP}, \bm{\tau}; \bm{\theta}\right)$. \\
	}
	\KwOut{\\ 
		\qquad (1) Reconstructed satellite subsystem temperature field $ {{\hat{\bm{T}}}_{pre}} $; \\
		\qquad (2) Quantified aleatoric uncertainty $ \hat{\bm{\sigma}}_{pre} $. \\
	}
	
	\For{$\mathcal{P}=1:N_{pre}$}{
		Randomly take a sample ${\tau}^{pre}_{\mathcal{P}}$ from uniform distribution $U\left ( 0, 1 \right )$; \\
		Create a ${{H}_{\Omega }}\times {{W}_{\Omega }}$ two-dimensional array $\bm{\tau}^{pre}_{\mathcal{P}}$ with all zero elements; \\
		Perform $\bm{\tau}^{pre}_{\mathcal{P}}\left(x,y\right) = {\tau}^{pre}_{\mathcal{P}}$ for $ \forall \left( x,y \right) \in {{\Omega}_{MP}} $; \\
		Composite training data $ \left( \bm{T}^{pre}_{MP},\bm{\tau}^{pre}_{\mathcal{P}} \right) $;\\
		Predict the reconstructed temperature field ${{\hat{\bm{T}}}^{pre}_{\mathcal{P}}}=\mathcal{M}\left(\bm{T}^{pre}_{MP},\bm{\tau}^{pre}_{\mathcal{P}}; \bm{\theta}\right)$.
	}
	Calculate the mean $ {{\hat{\bm{T}}}_{pre}} $ of $ \left\{{{\hat{\bm{T}}}^{pre}_{\mathcal{P}}}|\mathcal{P}=1,2,\cdots,N_{pre}\right\} $ by Eq.(\ref{T_pre}); \\
	Calculate the standard deviation $ \hat{\bm{\sigma}}_{pre} $ of $ \left\{{{\hat{\bm{\sigma}}}^{pre}_{\mathcal{P}}}|\mathcal{P}=1,2,\cdots,N_{pre}\right\} $ by Eq.(\ref{U_alea}). \\
\end{algorithm}

Given the test dataset $ \left\{\left(\bm{T}^{pre}_{MP,t}, \bm{T}_{test}^{t}\right)|t=1,2,\cdots,N_{test}\right\} $, the reconstructed temperature field's approximations $ \left\{\hat{\bm{T}}^t_{pre}|t=1,2,\cdots,N_{MCS}\right\} $ can be obtained by the \textbf{Algorithm} \ref{algorithm2}. In this paper, the prediction performance of the trained model $\mathcal{M}\left(\bm{T}_{MP}, \bm{\tau}; \bm{\theta}\right)$ is evaluated by the following four indicators, i.e.,
\begin{itemize}
	\item the average of root mean square error (RMSE)
		  \begin{equation}\label{e_RMSE}
		  \overline{RMSE}=\frac{1}{{{N}_{test}}}\sum\limits_{t=1}^{{{N}_{test}}}{\sqrt{\frac{1}{{{H}_{\Omega }}\times {{W}_{\Omega }}}\sum\limits_{\left( x,y \right)\in \Omega }{{{\left[ \bm{T}_{test}^{t}\left( x,y \right)-\bm{\hat{T}}_{pre}^{t}\left( x,y \right) \right]}^{2}}}}},
		  \end{equation}	
	\item the average of mean absolute error (MAE)
		  \begin{equation}\label{e_MAE}
		  \overline{MAE}=\frac{1}{{{N}_{test}}}\sum\limits_{t=1}^{{{N}_{test}}}{\frac{1}{{{H}_{\Omega }}\times {{W}_{\Omega }}}\sum\limits_{\left( x,y \right)\in \Omega }{\left| \bm{T}_{test}^{t}\left( x,y \right)-\bm{\hat{T}}_{pre}^{t}\left( x,y \right) \right|}},
		  \end{equation}
	\item the average of mean relative error (MRE)
		  \begin{equation}\label{e_MRE}
		  \overline{MRE}=\frac{1}{{{N}_{test}}}\sum\limits_{t=1}^{{{N}_{test}}}{\frac{1}{{{H}_{\Omega }}\times {{W}_{\Omega }}}\sum\limits_{\left( x,y \right)\in \Omega }{\frac{\left| \bm{T}_{test}^{t}\left( x,y \right)-\bm{\hat{T}}_{pre}^{t}\left( x,y \right) \right|}{\bm{T}_{test}^{t}\left( x,y \right)}}},
		  \end{equation}
    \item the average of R square ($R^2$)
	      \begin{equation}\label{R2}
	      \begin{aligned}
	      & \bm{\bar{T}}_{pre}^{t}=\frac{1}{{{N}_{test}}}\sum\limits_{t=1}^{{{N}_{test}}}{\bm{\hat{T}}_{pre}^{t}}, \\ 
	      & \overline{{{R}^{2}}}=\frac{1}{{{N}_{test}}}\sum\limits_{t=1}^{{{N}_{test}}}{\left\{ 1-{\frac{\sum\limits_{\left( x,y \right)\in \Omega }{{\left[ \bm{T}_{test}^{t}\left( x,y \right)-\bm{\hat{T}}_{pre}^{t}\left( x,y \right) \right]}^{2}}} {\sum\limits_{\left( x,y \right)\in \Omega }{{\left[ \bm{T}_{test}^{t}\left( x,y \right)-\bm{\bar{T}}_{pre}^{t}\left( x,y \right) \right]}^{2}}}} \right\}}. \\ 
	      \end{aligned}
	      \end{equation}
\end{itemize}
For the first three indicators $ \overline{RMSE} $, $ \overline{MAE} $ and $ \overline{MRE} $, the closer their values are to 0, the higher the accuracy of the trained model $\mathcal{M}\left(\bm{T}_{MP}, \bm{\tau}; \bm{\theta}\right)$. For the average of R square, ${{R}^{2}}\to 1$ means that the trained model $\mathcal{M}\left(\bm{T}_{MP}, \bm{\tau}; \bm{\theta}\right)$ can estimate the reconstructed satellite subsystem temperature field accurately.

\section{Satellite heat reliability analysis based on interval multilevel BN}\label{sec4}
\subsection{Aleatoric uncertainty-based satellite component normal probability interval estimation}\label{sec41}
Due to the approximation $\hat{\bm{T}}_{pre}$ with the aleatoric uncertainty $\bm{\sigma}_{pre}$, the interval temperature field $\widetilde{\bm{T}}_{pre}$ is used to represent the reconstructed temperature field of the MP temperature image $ \bm{T}^{pre}_{MP} $, i.e.,
\begin{equation}\label{interval_T}
\begin{aligned}
& {{{\hat{\bm{T}}}}_{+}}={{{\hat{\bm{T}}}}_{pre}}+\lambda {{\bm{\sigma }}_{pre}}, \\ 
& {{{\hat{\bm{T}}}}_{-}}={{{\hat{\bm{T}}}}_{pre}}-\lambda {{\bm{\sigma }}_{pre}}, \\ 
& {{{\bm{\widetilde{T}}}}_{pre}}=\left[ {{{\hat{\bm{T}}}}_{-}},{{{\hat{\bm{T}}}}_{+}} \right], \\ 
\end{aligned}
\end{equation}
where $ \lambda $ is the hyperparameter. In Eq.(\ref{interval_T}), any element in $ {{{\hat{\bm{T}}}}_{+}} $ and $ {{{\hat{\bm{T}}}}_{-}} $ is greater than or equal to zero, i.e., $\forall \left( x,y \right)\in{\Omega}$, $ {{{\hat{\bm{T}}}}_{+}}\left( x,y \right)\ge0 $ and ${{{\hat{\bm{T}}}}_{-}}\left( x,y \right)\ge0$. Define that $ {\Omega}_{C_s} $ represents the area where the component $ C_s $ ($ s=1,2,\cdots,n $) is located in $\Omega$ (Fig.\ref{TFR}). The interval highest temperature $ \widetilde{T}_s $ in the area ${\Omega}_{C_s}$ is
\begin{equation}\label{Ts}
{{\widetilde{T}}_{s}}=\left[ \underset{\left( x,y \right)\in {{\Omega }_{{{C}_{s}}}}}{\mathop{\max }}\,{{{\hat{\bm{T}}}}_{-}}\left( x,y \right),\underset{\left( x,y \right)\in {{\Omega }_{{{C}_{s}}}}}{\mathop{\max }}\,{{{\hat{\bm{T}}}}_{+}}\left( x,y \right) \right].
\end{equation}

The temperature field of the satellite subsystem will change with the different working power of the components, thereby affecting each component's operating performance. In this paper, ${{C}_{s}}=0$ and ${{C}_{s}}=1$  indicate the component failure and normal, respectively. The working state of component $ C_s $ depends on the interval highest temperature $ T_s $ in the area ${\Omega}_{C_s}$, and the definitions of two working states are
\begin{equation}\label{state_defin}
\left\{ \begin{aligned}
& {{C}_{s}}=0,\qquad {{T}_{s}}\ge T_{lim}^{C_s} \\ 
& {{C}_{s}}=1,\qquad {{T}_{s}}<T_{lim}^{C_s} \\ 
\end{aligned} \right.,
\end{equation}
where $ T_{lim}^{C_s} $ is working state threshold for the component $ C_s $.

Given $ N_{MCS} $ MP temperature images $ \left\{\bm{T}^{pre}_{MP,m}|m=1,2,\cdots,N_{MCS}\right\} $, the reconstructed temperature field's approximations $ \left\{\hat{\bm{T}}^m_{pre}|m=1,2,\cdots,N_{MCS}\right\} $ and the aleatoric uncertainty $ \left\{\bm{\sigma}^m_{pre}|m=1,2,\cdots,N_{MCS}\right\} $ are obtained by Eqs.(\ref{T_pre}) and (\ref{U_alea}). Then, $ N_{MCS} $ interval highest temperatures $ \left\{\widetilde{T}^m_s|m=1,2,\cdots,N_{MCS}\right\} $ are calculated using Eqs.(\ref{interval_T}) and (\ref{Ts}). Thus, the interval highest temperatures' lower bound set $ {\mathcal{T}}^{lower}_s $ and upper bound set $ {\mathcal{T}}^{upper}_s $ respectively are
\begin{equation}\label{Ts_L_U}
\begin{aligned}
& {\mathcal{T}}^{lower}_s= \left\{\widetilde{T}^{-m}_s|m=1,2,\cdots,N_{MCS}\right\}, \\
& {\mathcal{T}}^{upper}_s= \left\{\widetilde{T}^{+m}_s|m=1,2,\cdots,N_{MCS}\right\}, \\
\end{aligned}
\end{equation}
where $ \widetilde{T}^{-m}_{s} $ and $ \widetilde{T}^{+m}_{s} $ are the lower bound and the upper bound of interval highest temperature $ \widetilde{T}^m_s $, respectively. This section defines a set $ {\mathcal{T}}^{\forall}_s $, i.e.,
\begin{equation}\label{T_condition}
{\mathcal{T}}^{\forall}_s = \left\{T^{m}_s|T^{m}_s \in {{\widetilde{T}}^m_{s}},m=1,2,\cdots,N_{MCS}\right\},
\end{equation}
where the set $ {\mathcal{T}}^{\forall}_s $ satisfies two conditions $ {\mathcal{T}}^{\forall}_s \ne {\mathcal{T}}^{lower}_s $ and $ {\mathcal{T}}^{\forall}_s \ne {\mathcal{T}}^{upper}_s $. Suppose that  $ {{N}^{-}_{{{C}_{s}}=1}} $, $ {{N}^{+}_{{{C}_{s}}=1}} $ and $ {{N}^{\forall}_{{{C}_{s}}=1}} $ represent the number of elements satisfying the conditions $ \widetilde{T}^{-m} < T_{lim}^{C_s} $, $ \widetilde{T}^{+m} < T_{lim}^{C_s} $ and $ T^{m}_s < T_{lim}^{C_s} $, respectively. Apparently, there is an inequality relationship
\begin{equation}\label{inequal_N}
{{N}^{+}_{{{C}_{s}}=1}} \le {{N}^{\forall}_{{{C}_{s}}=1}} \le {{N}^{-}_{{{C}_{s}}=1}}.
\end{equation}
According to Eq.(\ref{state_defin}), three normal probabilities $ {\Pr}^{-} \left( {{C}_{s}}=1 \right) $, $ {\Pr}^{+} \left( {{C}_{s}}=1 \right) $ and $ {\Pr}^{\forall} \left( {{C}_{s}}=1 \right) $ are
\begin{equation}\label{Pr_Cs_3}
\begin{aligned}
& {\Pr}^{-} \left( {{C}_{s}}=1 \right)=\frac{{{N}^{+}_{{{C}_{s}}=1}}}{{{N}_{MCS}}}, \\
& {\Pr}^{+} \left( {{C}_{s}}=1 \right)=\frac{{{N}^{-}_{{{C}_{s}}=1}}}{{{N}_{MCS}}}, \\
& {\Pr}^{\forall} \left( {{C}_{s}}=1 \right)=\frac{{{N}^{\forall}_{{{C}_{s}}=1}}}{{{N}_{MCS}}},
\end{aligned}
\end{equation}
Refer to Eq.(\ref{inequal_N}), three normal probabilities $ {\Pr}^{-} \left( {{C}_{s}}=1 \right) $, $ {\Pr}^{+} \left( {{C}_{s}}=1 \right) $ and $ {\Pr}^{\forall} \left( {{C}_{s}}=1 \right) $ exist an inequality relation
\begin{equation}\label{Pr_inequal}
{\Pr}^{-} \left( {{C}_{s}}=1 \right) \le {\Pr}^{\forall} \left( {{C}_{s}}=1 \right) \le {\Pr}^{+} \left( {{C}_{s}}=1 \right).
\end{equation}
Therefore, the normal probability interval $ \widetilde{\Pr} \left( {{C}_{s}}=1 \right) $ is
\begin{equation}\label{Pr_Cs_1}
\widetilde{\Pr} \left( {{C}_{s}}=1 \right) = \left[{\Pr}^{-} \left( {{C}_{s}}=1 \right),{\Pr}^{+} \left( {{C}_{s}}=1 \right) \right].
\end{equation}

In summary, the pseudo code for calculating the satellite component normal probability interval is shown in the following \textbf{Algorithm} \textbf{\ref{algorithm3}}.

\begin{algorithm}[!htbp]
	\caption{Satellite component normal probability interval estimation algorithm.}
	\label{algorithm3}
	\LinesNumbered
	\KwIn{\\ 
		\qquad (1) Hyperparameter $\lambda$; \\
		\qquad (2) Working state thresholds $ \left\{T_{lim}^{C_s}|s=1,2,\cdots, n\right\} $ ; \\
		\qquad (3) The number $ N_{pre} $ of quantile levels; \\
		\qquad (4) Trained physics-informed Deep MC-QR model $\mathcal{M}\left(\bm{T}_{MP}, \bm{\tau}; \bm{\theta}\right)$; \\
		\qquad (5) $N_{MCS}$ discretized MP temperature images $\left\{ \bm{T}^{pre}_{MP,m}|m=1,2,\cdots,N_{MCS} \right\}$.
	}
	\KwOut{\\ \qquad Satellite component normal probability interval $ \left\{\widetilde{\Pr} \left( {{C}_{s}=1} \right)|s=1,2,\cdots, n \right\} $.}
	 
	\For{$m=1:N_{MCS}$}{
		Calculate the mean $ {{\hat{\bm{T}}}^{m}_{pre}} $ and the standard deviation $ \hat{\bm{\sigma}}^m_{pre} $ by the \textbf{Algorithm} \ref{algorithm2}; \\
		Calculate the interval highest temperatures $ \widetilde{T}^m_s $ by Eqs.(\ref{interval_T}) and (\ref{Ts}). \\
	}
	\For{$s=1:n$}{
		Construct the lower bound set ${\mathcal{T}}^{lower}_s= \left\{\widetilde{T}^{-m}_s|m=1,2,\cdots,N_{MCS}\right\}$; \\
		Construct the upper bound set $ {\mathcal{T}}^{upper}_s= \left\{\widetilde{T}^{+m}_s|m=1,2,\cdots,N_{MCS}\right\} $; \\
		Count the number $ {{N}^{-}_{{{C}_{s}}=1}} $ of elements in the set $ {\mathcal{T}}^{lower}_s $ that meet condition $ \widetilde{T}^{-m} < T_{lim}^{C_s} $; \\
		Count the number $ {{N}^{+}_{{{C}_{s}}=1}} $ of elements in the set $ {\mathcal{T}}^{lower}_s $ that meet condition $ \widetilde{T}^{+m} < T_{lim}^{C_s} $; \\
		Calculate the normal probability interval $ \widetilde{\Pr} \left( {{C}_{s}}=1 \right) = \left[{\Pr}^{-} \left( {{C}_{s}}=1 \right), {\Pr}^{+} \left( {{C}_{s}}=1 \right) \right] $.
	}
\end{algorithm}

\subsection{Interval multilevel BN modeling for satellite heat reliability analysis}\label{sec42}
\begin{figure*}[!htb]
	\centering
	{\includegraphics[scale=0.8]{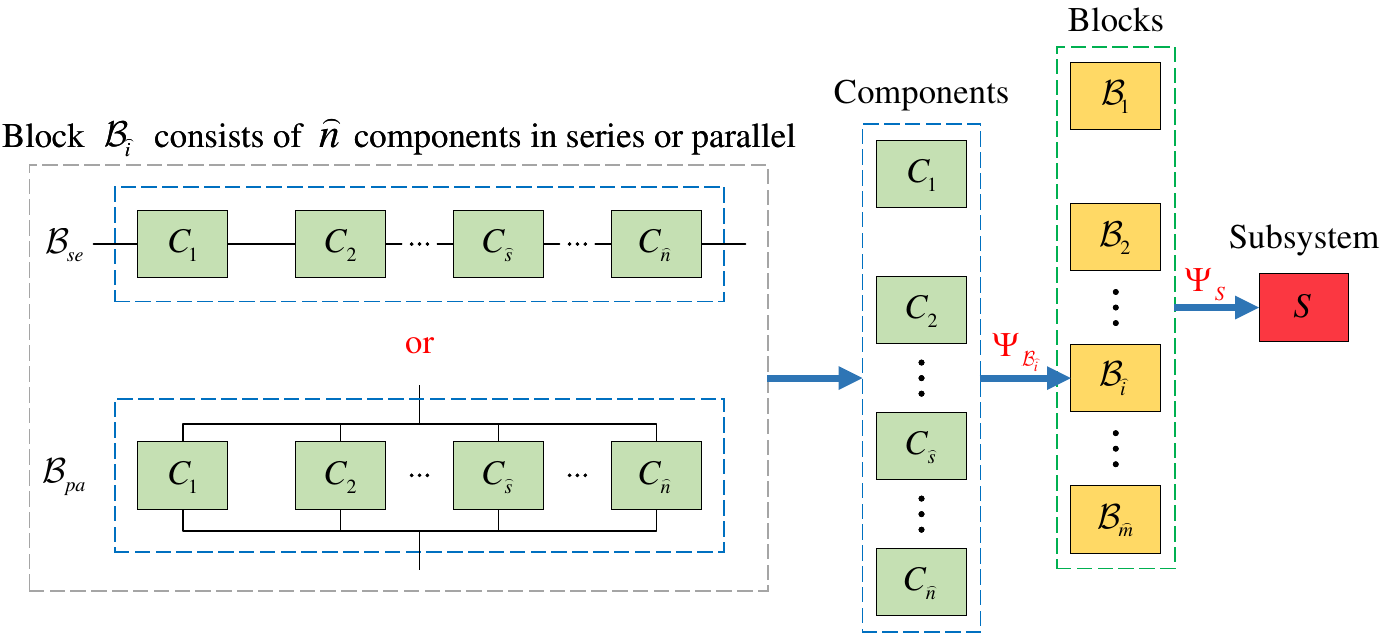}}
	\caption{The studied satellite subsystem $S$. The block $ \mathcal{B}_{\stackrel\frown{i}}$ may be series block $ \mathcal{B}_{se} $ and  parallel block $ \mathcal{B}_{pa} $ .}\label{Block}
\end{figure*}

As shown in Fig.\ref{Block}, the components of the studied satellite subsystem $ S $ in Fig.\ref{TFR} can be divided into $ \stackrel\frown{m} $ blocks $ \left\{ \mathcal{B}_{\stackrel\frown{i}}| \stackrel\frown{i}=1,2,\cdots, \stackrel\frown{m} \right\} $. In Fig.\ref{Block}, $ {{\Psi }_{S}} $ denotes the logical relationship between subsystem $ S $ and $ \stackrel\frown{m} $ blocks, and $ {{\Psi }_{\mathcal{B}_{\stackrel\frown{i}}}} $ represents the logical relationship between the block $ \mathcal{B}_{\stackrel\frown{i}} $ and its $ \stackrel\frown{n} $ components. For the studied satellite subsystem $ S $, the logical relationshipships between components mainly include series and parallel. Thus, the block $ \mathcal{B}_{\stackrel\frown{i}} $ may be series block $ \mathcal{B}_{se} $ or parallel block $ \mathcal{B}_{pa} $ as shown in Fig.\ref{Block}. Then, the blocks $ \left\{ \mathcal{B}_{\stackrel\frown{i}}| \stackrel\frown{i}=1,2,\cdots, \stackrel\frown{m} \right\} $ are connected in series or parallel to form the satellite subsystem $ S $, i.e. the logical relationship $ {{\Psi }_{S}} $ may be series or parallel.

\begin{figure*}[!htb]
	\centering
	{\includegraphics[scale=0.8]{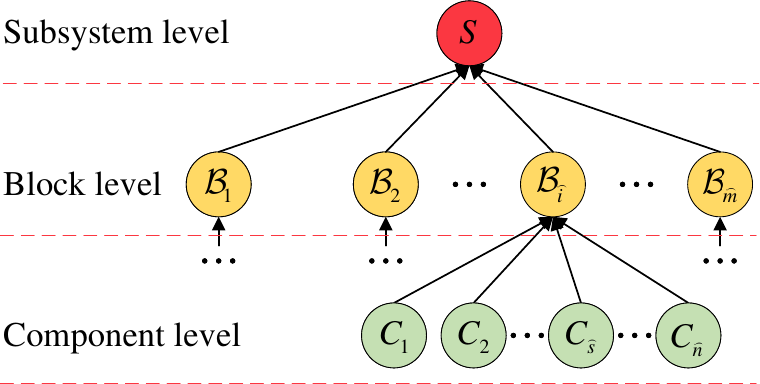}}
	\caption{An interval multilevel BN for analyzing the heat reliability of the satellite subsystem}\label{IBN}
\end{figure*}

According to Fig.\ref{Block}, this section constructs a interval multilevel BN for analyzing the satellite subsystem $ S $ reliability as shown in Fig.\ref{IBN}. For the conditional probability table (CPT) of each non-root node, all parent nodes' state combinations are determined by the format in \cite{Zheng2019, Zheng2020}. For example, the CPT $ \Pr \left( S|{\mathcal{B}_{1}},{\mathcal{B}_{2}},\cdots ,{\mathcal{B}_{\stackrel\frown{m}}} \right) $ of the leaf node $ S $ is shown in Table \ref{Pr_S_CPT}, where $S=0$ and $S=1$ respectively denote the subsystem failure and normal, the values of variables $\left\{p_{1,1}, p_{1,2},\cdots,p_{2^n,1}p_{2^n,2}\right\}$ are determined by the logical relationship $ {{\Psi }_{S}} $. By the variable elimination algorithm (VE) \cite{Dechter1998}, the interval probability distribution $\widetilde{\Pr}\left(S\right)$ of the child node $ S $ can be calculated by Eq.(\ref{Pr_S}).

\begin{table}[htbp]	
	\centering	
	\caption{The CPT $ \Pr \left( S|{\mathcal{B}_{1}},{\mathcal{B}_{2}},\cdots ,{\mathcal{B}_{\stackrel\frown{m}}} \right) $ of the leaf node $ S $}	
	\label{Pr_S_CPT}	
	\begin{tabular}{lllllcc}
		\toprule
		\multirow{2}{*}{$\mathcal{B}_{1}$} & \multirow{2}{*}{$\mathcal{B}_{2}$} & \multirow{2}{*}{$\cdots$} & \multirow{2}{*}{${\mathcal{B}_{\stackrel\frown{m}-1}}$} & \multirow{2}{*}{$\mathcal{B}_{\stackrel\frown{m}}$} & \multicolumn{2}{c}{$ \Pr \left( S|{\mathcal{B}_{1}},{\mathcal{B}_{2}},\cdots ,{\mathcal{B}_{\stackrel\frown{m}}} \right) $} \\
		\cmidrule(r){6-7}		
		& & & & & $S=0$ & $S=1$   \\		
		\midrule		
		0	&0 &$\cdots$ &0 &0	&$p_{1,1}$	&$p_{1,2}$ \\
		0	&0 &$\cdots$ &0 &1	&$p_{2,1}$	&$p_{2,2}$ \\
		0	&0 &$\cdots$ &1 &0	&$p_{3,1}$	&$p_{3,2}$ \\
		0	&0 &$\cdots$ &1 &1	&$p_{4,1}$	&$p_{4,2}$ \\
		$\vdots$ &$\vdots$ &$\vdots$ &$\vdots$ &$\vdots$	&$\vdots$	&$\vdots$ \\
		1	&1 &$\cdots$ &0 &0	&$p_{(2^{\stackrel\frown{m}}-3),1}$ &$p_{(2^{\stackrel\frown{m}}-3),2}$ \\
		1	&1 &$\cdots$ &0 &1	&$p_{(2^{\stackrel\frown{m}}-2),1}$ &$p_{(2^{\stackrel\frown{m}}-2),2}$ \\
		1	&1 &$\cdots$ &1 &0	&$p_{(2^{\stackrel\frown{m}}-1),1}$ &$p_{(2^{\stackrel\frown{m}}-1),2}$ \\
		1	&1 &$\cdots$ &1 &1	&$p_{2^{\stackrel\frown{m}},1}$ &$p_{2^{\stackrel\frown{m}},2}$ \\
		\bottomrule		
	\end{tabular}	
\end{table}

\begin{equation}\label{Pr_S}
\begin{aligned}
& {\widetilde{\Pr}} \left( S \right)=\sum\limits_{\mathcal{B}_{1}}^{2}{\cdots \sum\limits_{{\mathcal{B}_{\stackrel\frown{m}-1}}=1}^{2}{\sum\limits_{\mathcal{B}_{\stackrel\frown{m}}=1}^{2}{{\widetilde{\Pr}} \left( {\mathcal{B}_{1}},{\mathcal{B}_{2}},\cdots ,{\mathcal{B}_{\stackrel\frown{m}}},S \right)}}} \\ 
& \qquad\;\;\,=\sum\limits_{{\mathcal{B}_{1}}=1}^{2}{\cdots \sum\limits_{{\mathcal{B}_{\stackrel\frown{m}-1}}=1}^{2}{\sum\limits_{\mathcal{B}_{\stackrel\frown{m}}=1}^{2}{{\widetilde{\Pr}} \left( {\mathcal{B}_{1}} \right)\cdots {\widetilde{\Pr}} \left( {\mathcal{B}_{\stackrel\frown{m}-1}} \right){\widetilde{\Pr}} \left( {\mathcal{B}_{\stackrel\frown{m}}} \right) \Pr \left( S|{\mathcal{B}_{1}},{\mathcal{B}_{2}},\cdots ,{\mathcal{B}_{\stackrel\frown{m}}} \right)}}} \\ 
& \qquad\;\;\,=\sum\limits_{\mathcal{B}_{1}=1}^{2}{{\widetilde{\Pr}} \left( {\mathcal{B}_{1}} \right)\cdots \sum\limits_{{\mathcal{B}_{\stackrel\frown{m}-1}}=1}^{2}{{\widetilde{\Pr}} \left( {\mathcal{B}_{\stackrel\frown{m}-1}} \right)\sum\limits_{\mathcal{B}_{\stackrel\frown{m}}=1}^{2}{{\widetilde{\Pr}} \left( {\mathcal{B}_{\stackrel\frown{m}}} \right)\Pr \left( S|{\mathcal{B}_{1}},{\mathcal{B}_{2}},\cdots ,{\mathcal{B}_{\stackrel\frown{m}}} \right)}}} \\ 
\end{aligned}
\end{equation}

For $ \stackrel\frown{i}=1,2,\cdots, \stackrel\frown{m} $, the interval probability $ {\widetilde{\Pr}} \left( {\mathcal{B}_{\stackrel\frown{i}}} \right) $ in Eq.(\ref{Pr_S}) is calculated by
\begin{equation}\label{Pr_Bi_se}
{\widetilde{\Pr}} \left( {\mathcal{B}_{\stackrel\frown{i}}} \right) = \widetilde{\Pr} \left( {{\mathcal{B}}_{se}} \right)
\end{equation}
for series logical relationship or calculated by
\begin{equation}\label{Pr_Bi_pa}
{\widetilde{\Pr}} \left( {\mathcal{B}_{\stackrel\frown{i}}} \right) = \widetilde{\Pr} \left( {{\mathcal{B}}_{pa}} \right)
\end{equation}
for parallel logical relationship, where $ \widetilde{\Pr} \left( {{\mathcal{B}}_{se}} \right) $ and $ \widetilde{\Pr} \left( {{\mathcal{B}}_{pa}} \right) $ are calculated as follows.

For the series block $ \mathcal{B}_{se} $ in Fig.\ref{Block}, its normal probability interval $ \widetilde{\Pr}\left(\mathcal{B}_{se}=1\right) $ is
\begin{equation}\label{Pr_se_1_all}
\widetilde{\Pr} \left( {{\mathcal{B}}_{se}}=1 \right)=\prod\limits_{\mathcal{K}=1}^{\stackrel\frown{n}}{\widetilde{\Pr} \left( {{C}_{\mathcal{K}}}=1 \right)}.
\end{equation}

For the parallel block $ \mathcal{B}_{pa} $  in Fig.\ref{Block}, its normal probability interval $\widetilde{\Pr} \left( {{\mathcal{B}}_{pa}}=1 \right)$ (See the \ref{appendix_A} for the detailed derivation process.) is
\begin{equation}\label{Pr_pa_0_all}
\widetilde{\Pr} \left( {{\mathcal{B}}_{pa}}=1 \right)=1-\prod\limits_{{\mathcal{K}}=1}^{\stackrel\frown{n}}{\left[1-\widetilde{\Pr} \left( {{C}_{\mathcal{K}}}=1 \right)\right]}.
\end{equation}

Like the calculations of $ \widetilde{\Pr} \left( {{\mathcal{B}}_{se}} \right) $ and $ \widetilde{\Pr} \left( {{\mathcal{B}}_{pa}} \right) $, the interval probability $ \widetilde{\Pr} \left( S=1 \right) $ can be calculated based on Eqs.(\ref{Pr_S}), (\ref{Pr_se_1_all}), and (\ref{Pr_pa_0_all}).

\section{Case study}\label{sec5}
In this section, two case studies are used for verifying the effectiveness of the proposed methods. Two U-nets in the physics-informed Deep MC-QR method are modeled by PyTorch\footnote{https://pytorch.org/get-started/locally/}. The first case study is used to validate that the proposed physics-informed Deep MC-QR method can accurately reconstruct the temperature field and precisely quantify the aleatoric uncertainty caused by data noise. The second case study analyzes the heat reliability of a satellite subsystem. For two case studies, the hyperparameters $ {\alpha }_{\tau} $, $ {\alpha }_{LE} $, $ {\alpha }_{BC} $, and $ {\alpha }_{TV} $ are equal to $1\times10^5$, $1\times10^2$, $1\times10^2$ and $1\times10^4$, respectively. The relevant codes of the physics-informed Deep MC-QR method by Python are available on this website\footnote{https://github.com/Xiaohu-Zheng/Physics-informed-Deep-MC-QR}. In this paper, the unit of temperature is K. In order to facilitate training model, the MP temperature value is normalized by ${(\bm{T}_{MP}-298)}/{50}\;$. In the prediction, the parameter $N_{pre}$ is 200 for two cases.

\subsection{Case 1: Physics-informed Deep MC-QR method validation}\label{sec51}
\begin{figure*}[!htb]
	\centering
	{\includegraphics[scale=1]{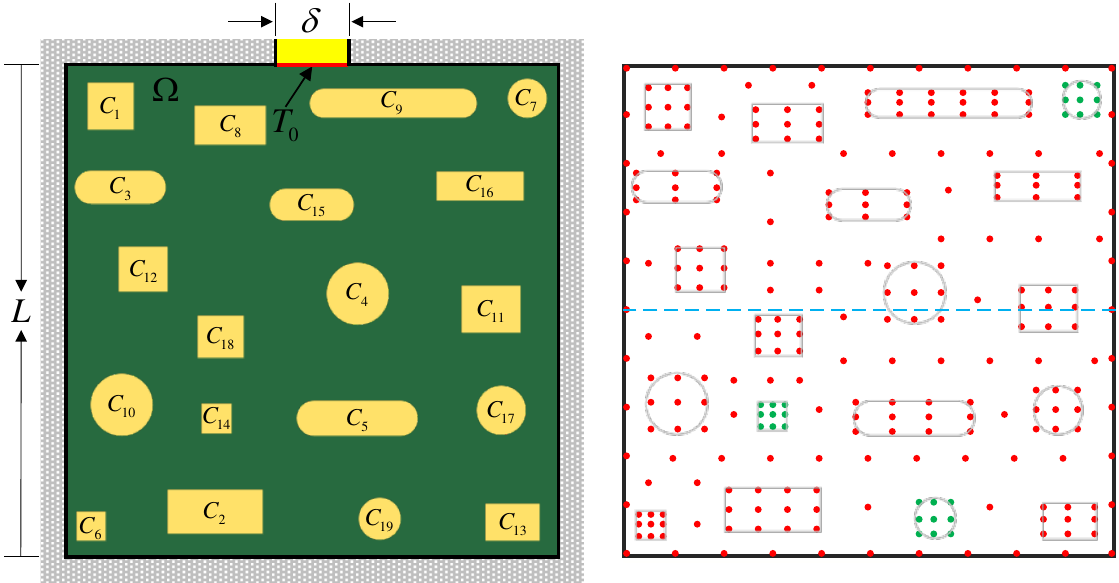}}
	\caption{A hypothetical subsystem with 19 components in case 1.}\label{example_case1}
\end{figure*}

The first case uses a hypothetical subsystem with 19 components, as shown in Fig.\ref{example_case1}. The length $ L $ is equal to 0.1m, and the width $\delta$ of the heat sink is 0.01m. The probability distributions of all component powers are shown in Table \ref{Com_PD_case1}. This case uses 280 sensors to monitor the point temperature, where the positions of all sensors are shown in Fig.\ref{example_case1}.

\begin{table}[htb]
	\centering
	\caption{All component powers' probability distributions in case 1.}
	\begin{tabular}{llcl}
		\toprule
		Component & Mean  & Standard deviation & Distribution \\
		\midrule
		$C_1$    & 30000 & 6000  & Gumbel \\
		$C_2$    & 38900 & 9092  & Lognormal \\
		$C_3$    & 35000 & 5000  & Normal \\
		$C_4$    & 41200 & 9140  & Gumbel \\
		$C_5$    & 36000 & 8024  & Lognormal \\
		$C_6$    & 12000 & 1590  & Normal \\
		$C_8$    & 31200 & 7150  & Gumbel \\
		$C_9$    & 40000 & 5521  & Normal \\
		$C_{11}$ & 31000 & 4311  & Normal \\
		$C_{12}$ & 28070 & 6100  & Lognormal \\
		$C_{13}$ & 29800 & 4321  & Normal \\
		$C_{16}$ & 28900 & 7216  & Gumbel \\
		$C_{17}$ & 22100 & 5021  & Lognormal \\
		$C_{18}$ & 22000 & 2521  & Normal \\
		\midrule
		Component & Lower boundary  & Upper boundary & Distribution \\
		\midrule
		$C_7$    & 100   & 12000 & Uniform \\
		$C_{10}$ & 2000  & 30000 & Uniform \\
		$C_{14}$ & 0     & 10000 & Uniform \\
		$C_{15}$ & 500   & 28000 & Uniform \\
		$C_{19}$ & 100   & 13000 & Uniform \\
		\bottomrule
	\end{tabular}
	\label{Com_PD_case1}
\end{table}

\subsubsection{Preparing monitoring point temperature data}\label{sedc511}
In this paper, the two-dimensional square area $\Omega$ is divided into a $200 \times 200$ grid $\bm{G}$. The values of grid cells in the area ${\Omega}_{C_{s}}$ ($s=1,2,\cdots,19$) are equal to the power $P_s$ of the component $C_s$, and the values of grid cells in the area $ {\Omega}_{NC} $ are equal to zero. According to the component powers' distributions, 20000 grids $\left\{\bm{G}_1, \bm{G}_2, \cdots ,\bm{G}_{20000}\right\}$ are sampled by the Latin Hypercube Sampling. Then, 20000 temperature fields $\left\{{\bm{T}}_1, {\bm{T}}_2, \cdots ,{\bm{T}}_{20000}\right\}$ are generated by the recon-data-generator\footnote{https://github.com/shendu-sw/recon-data-generator}. Refer to all sensors' positions, their values are taken from the temperature fields $\left\{{\bm{T}}_1, {\bm{T}}_2, \cdots ,{\bm{T}}_{20000}\right\}$ to form 20000 MP temperatures $\left\{\bm{T}^1_{MP}, \bm{T}^2_{MP}, \cdots ,\bm{T}^{20000}_{MP}\right\}$. Based on these MP temperatures, this case prepares three kinds of datasets with Gaussian noises, i.e., $\mathcal{D}_{\varepsilon_1}$, $\mathcal{D}_{\varepsilon_2}$, and $\mathcal{D}^b_{\varepsilon_1}$, to validate that the proposed physics-informed Deep MC-QR method can accurately quantify the aleatoric uncertainty caused by data noise. The Gaussian noises $ {\varepsilon}_1 \sim N\left ( 0, {0.25}^2  \right )  $ and $ {\varepsilon}_2 \sim N\left ( 0, {0.50}^2  \right )  $ are added to the temperature sensors values of green points in Fig.\ref{example_case1} to generate the first two datasets $\mathcal{D}_{\varepsilon_1}$ and $\mathcal{D}_{\varepsilon_2}$, respectively. The Gaussian noise $ {\varepsilon}_1 \sim N\left ( 0, {0.25}^2  \right )  $ is added to the temperature values of the measuring points below the blue line in Fig.\ref{example_case1} to generate the last dataset $\mathcal{D}^b_{\varepsilon_1}$, respectively. For 20000 MP temperatures $\left\{\bm{T}^1_{MP}, \bm{T}^2_{MP}, \cdots ,\bm{\textsc{T}}^{20000}_{MP}\right\}$, 10800 MP temperatures are used to train the physics-informed Deep MC-QR model $\mathcal{M}\left(\bm{T}_{MP}, \bm{\tau}; \bm{\theta}\right)$, 1200  MP temperatures are used to be the validate data, and the remaining 8000 MP temperatures are used to test the trained model $\mathcal{M}\left(\bm{T}_{MP}, \bm{\tau}; \bm{\theta}\right)$.

\subsubsection{Result analysis}\label{sedc512}
\paragraph{\textbf{Model accuracy analysis}}
In this case, the maximum training epoch and the learning rate are set to be 400 and 0.1, respectively. Based on the prepared three kinds of training datasets, three models $\mathcal{M}^{{\varepsilon}_1}\left(\bm{T}_{MP}, \bm{\tau}; \bm{\theta}\right)$, $\mathcal{M}^{{\varepsilon}_2}\left(\bm{T}_{MP}, \bm{\tau}; \bm{\theta}\right)$ and $\mathcal{M}^{b{\varepsilon}_1}\left(\bm{T}_{MP}, \bm{\tau}; \bm{\theta}\right)$ are trained by the \textbf{Algorithm} \ref{algorithm1}. Then, the temperature fields corresponding for 8000 MP temperatures are reconstructed by the \textbf{Algorithm} \ref{algorithm2}. According to four indicators in section \ref{sec34}, the prediction performance of three trained models $\mathcal{M}^{{\varepsilon}_1}\left(\bm{T}_{MP}, \bm{\tau}; \bm{\theta}\right)$, $\mathcal{M}^{{\varepsilon}_2}\left(\bm{T}_{MP}, \bm{\tau}; \bm{\theta}\right)$ and $\mathcal{M}^{b{\varepsilon}_1}\left(\bm{T}_{MP}, \bm{\tau}; \bm{\theta}\right)$ are shown in Table \ref{model_acc_case1}. For three trained models, all three kinds of average errors, i.e., $ \overline{RMSE} $, $ \overline{MAE} $ and $ \overline{MRE} $, are sufficiently small. Besides, the $\overline{R^2}$ values of three trained models are almost close to 1. Therefore, the proposed physics-informed Deep MC-QR method can construct an accurate surrogate model to reconstruct the temperature fields of the hypothetical subsystem.

\begin{table}[!htb]
	\centering
	\caption{The prediction performance of three trained models $\mathcal{M}^{{\varepsilon}_1}\left(\bm{T}_{MP}, \bm{\tau}; \bm{\theta}\right)$, $\mathcal{M}^{{\varepsilon}_2}\left(\bm{T}_{MP}, \bm{\tau}; \bm{\theta}\right)$ and $\mathcal{M}^{b{\varepsilon}_1}\left(\bm{T}_{MP}, \bm{\tau}; \bm{\theta}\right)$ in case 1.}
	\begin{tabular}{lllllll}
		\toprule
		Dataset & Noise variance & Model & $ \overline{RMSE} $  & $ \overline{MAE} $ & $ \overline{MRE} $ & $\overline{R^2}$ \\
		\midrule
		$\mathcal{D}_{\varepsilon_1}$ & $0.25^2$ & $\mathcal{M}^{{\varepsilon}_1}\left(\bm{T}_{MP}, \bm{\tau}; \bm{\theta}\right)$ & 0.0511 & 0.0324 & 0.00009 & 0.9999 \\
		$\mathcal{D}_{\varepsilon_2}$ & $0.50^2$ & $\mathcal{M}^{{\varepsilon}_2}\left(\bm{T}_{MP}, \bm{\tau}; \bm{\theta}\right)$ & 0.0490 & 0.0306 & 0.00009 & 0.9999 \\
		$\mathcal{D}^b_{\varepsilon_1}$ & $0.25^2$ & $\mathcal{M}^{b{\varepsilon}_1}\left(\bm{T}_{MP}, \bm{\tau}; \bm{\theta}\right)$ & 0.0688 & 0.0486 & 0.00014 & 0.9999 \\
		\bottomrule
	\end{tabular}
	\label{model_acc_case1}
\end{table}

\begin{figure*}[htb]
	\centering
	\subfigure [MP temperature $ \bm{T}^{1*}_{MP} $]
	{\includegraphics[scale=0.38]{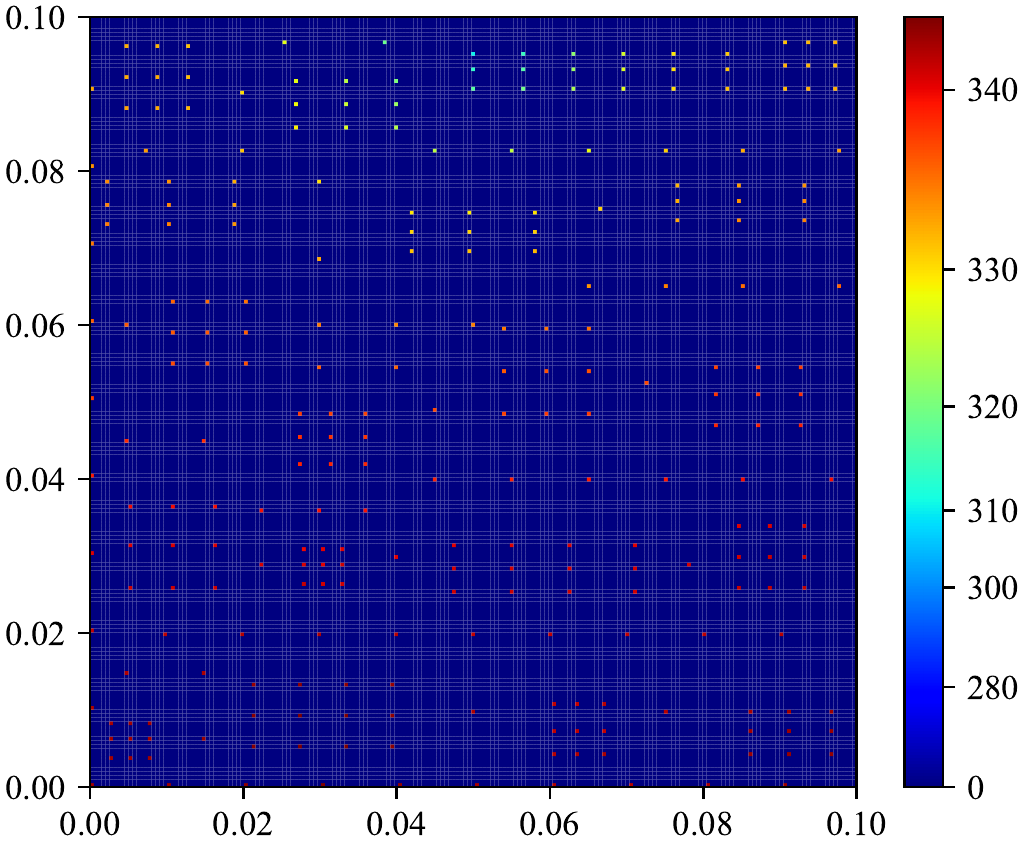}}
	\subfigure [MP temperature $ \bm{T}^{2*}_{MP} $]
	{\includegraphics[scale=0.38]{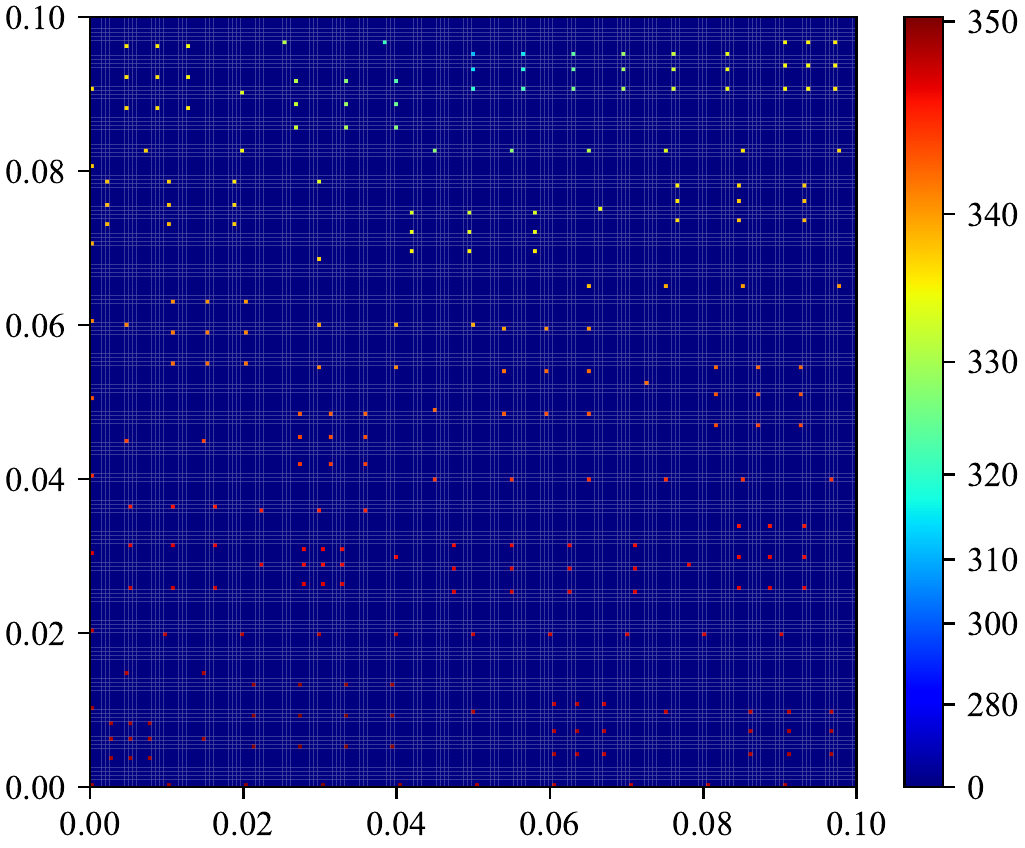}}
	\subfigure [MP temperature $ \bm{T}^{3*}_{MP} $]
	{\includegraphics[scale=0.38]{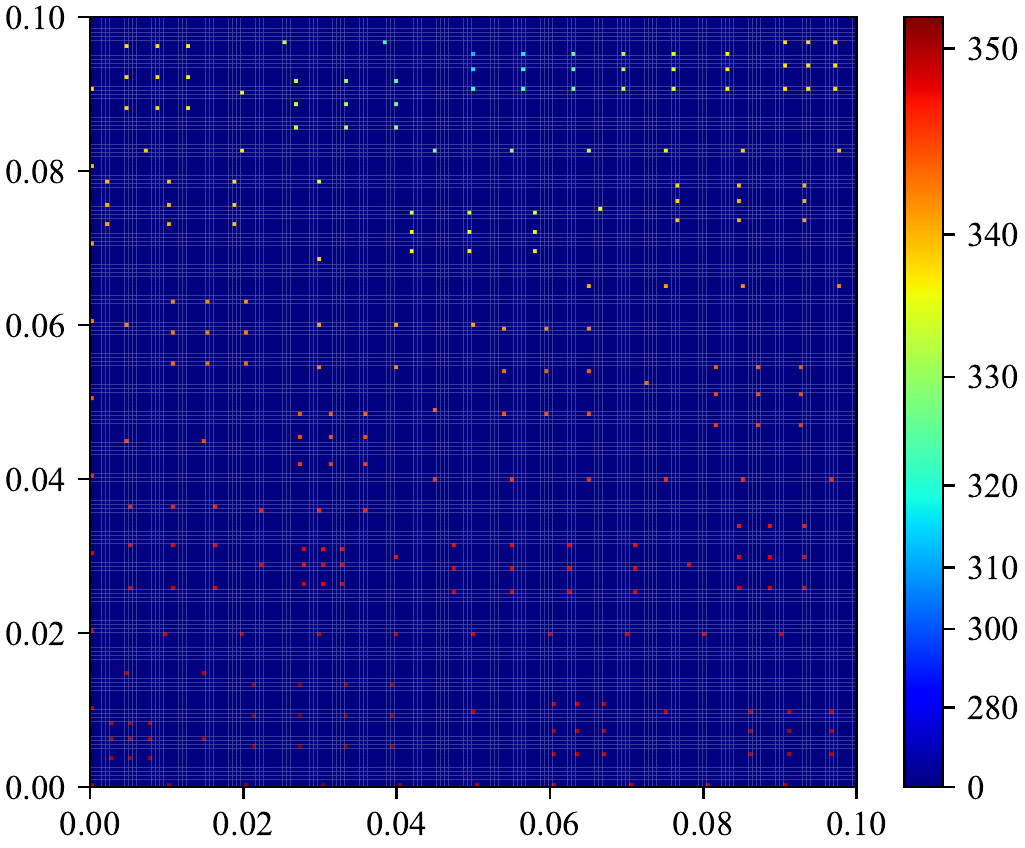}}
	\subfigure [MP temperature $ \bm{T}^{4*}_{MP} $]
	{\includegraphics[scale=0.38]{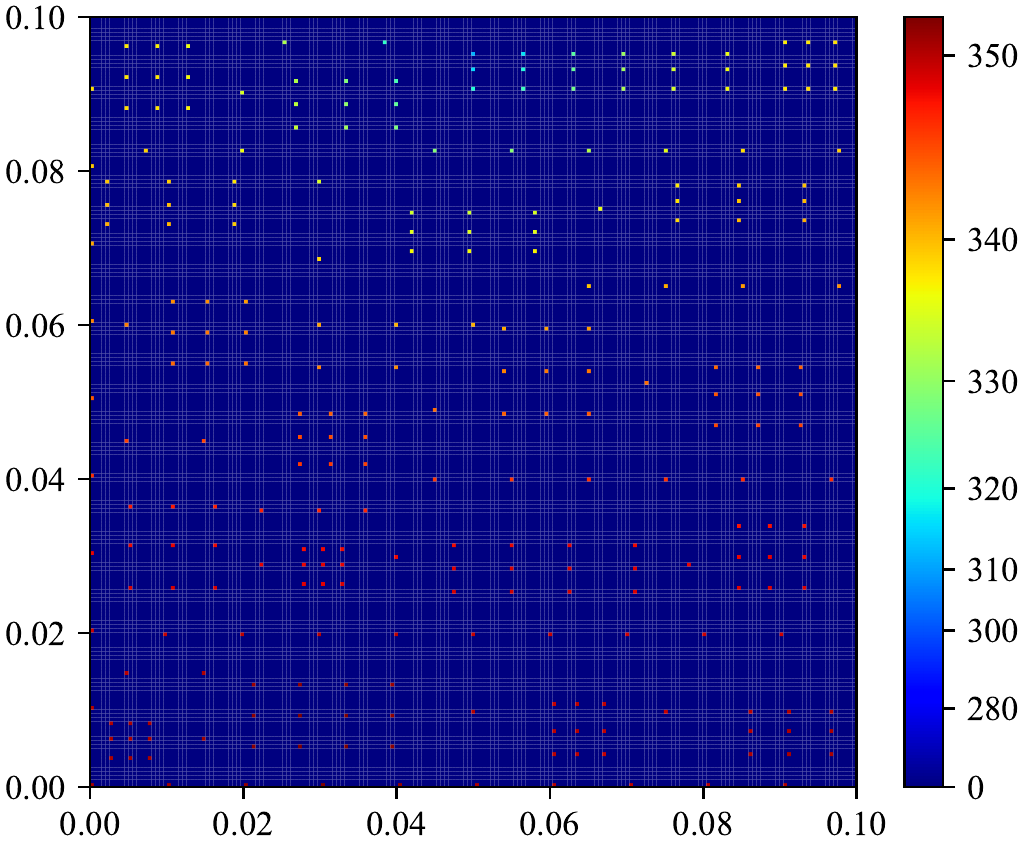}}	
	\caption{Four MP temperatures $ \left\{\bm{T}^{1*}_{MP}, \bm{T}^{2*}_{MP}, \bm{T}^{3*}_{MP}, \bm{T}^{4*}_{MP}\right\} $.}\label{MP_case1}
\end{figure*}

\begin{figure*}[!htb]
	\centering
	{\includegraphics[scale=0.40]{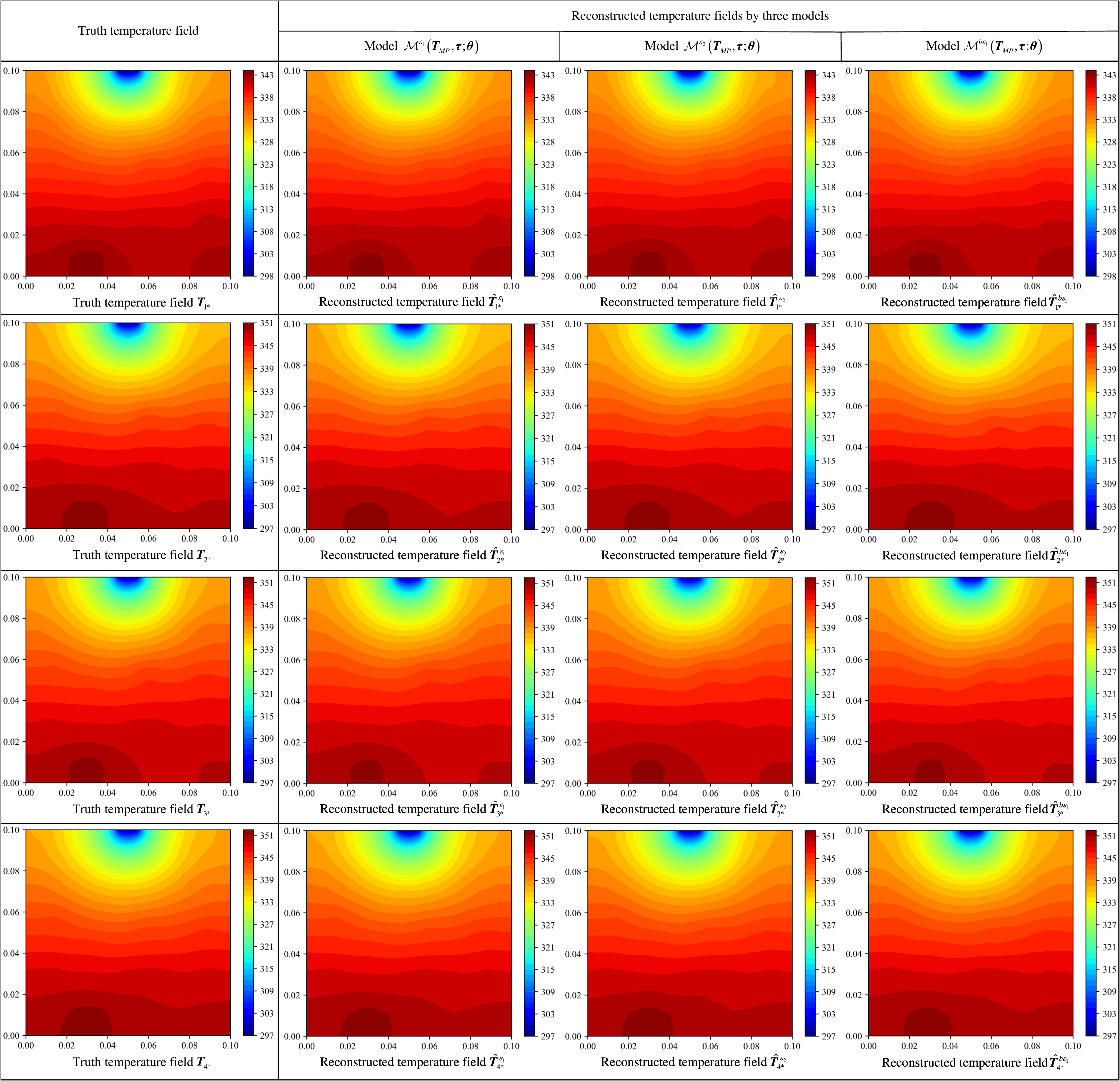}}
	\caption{Four MP temperatures $ \left\{\bm{T}^{1*}_{MP}, \bm{T}^{2*}_{MP}, \bm{T}^{3*}_{MP}, \bm{T}^{4*}_{MP}\right\} $ and their truth temperature fields $ \left\{\bm{T}_{1*}, \bm{T}_{2*}, \bm{T}_{3*}, \bm{T}_{4*}\right\} $. The first column on the left is the truth temperature fields corresponding to four MP temperatures $ \left\{\bm{T}^{1*}_{MP}, \bm{T}^{2*}_{MP}, \bm{T}^{3*}_{MP}, \bm{T}^{4*}_{MP}\right\} $. For each row, the last three figures show the temperature fields reconstructed by the three models $\mathcal{M}^{{\varepsilon}_1}\left(\bm{T}_{MP}, \bm{\tau}; \bm{\theta}\right)$, $\mathcal{M}^{{\varepsilon}_2}\left(\bm{T}_{MP}, \bm{\tau}; \bm{\theta}\right)$ and $\mathcal{M}^{b{\varepsilon}_1}\left(\bm{T}_{MP}, \bm{\tau}; \bm{\theta}\right)$ for the same MP temperature.}\label{MP_pre_case1}
\end{figure*}

As shown in Fig.\ref{MP_case1}, this case randomly chooses four MP temperatures $ \left\{\bm{T}^{1*}_{MP}, \bm{T}^{2*}_{MP}, \bm{T}^{3*}_{MP}, \bm{T}^{4*}_{MP}\right\} $ to show the prediction effectiveness of three trained models $\mathcal{M}^{{\varepsilon}_1}\left(\bm{T}_{MP}, \bm{\tau}; \bm{\theta}\right)$, $\mathcal{M}^{{\varepsilon}_2}\left(\bm{T}_{MP}, \bm{\tau}; \bm{\theta}\right)$ and $\mathcal{M}^{b{\varepsilon}_1}\left(\bm{T}_{MP}, \bm{\tau}; \bm{\theta}\right)$. For four MP temperatures $ \left\{\bm{T}^{1*}_{MP}, \bm{T}^{2*}_{MP}, \bm{T}^{3*}_{MP}, \bm{T}^{4*}_{MP}\right\} $, their truth temperature fields $ \left\{\bm{T}_{1*}, \bm{T}_{2*}, \bm{T}_{3*}, \bm{T}_{4*}\right\}$ and reconstructed temperature fields by three trained models are presented in Fig.\ref{MP_pre_case1}. For each row, the last three reconstructed temperature fields are basically consistent with the first truth temperature field.

\begin{figure*}[!htb]
	\centering
	\subfigure [Aleatoric uncertainty $\sigma_{1}^{{\varepsilon}_1}$]
	{\includegraphics[scale=0.383]{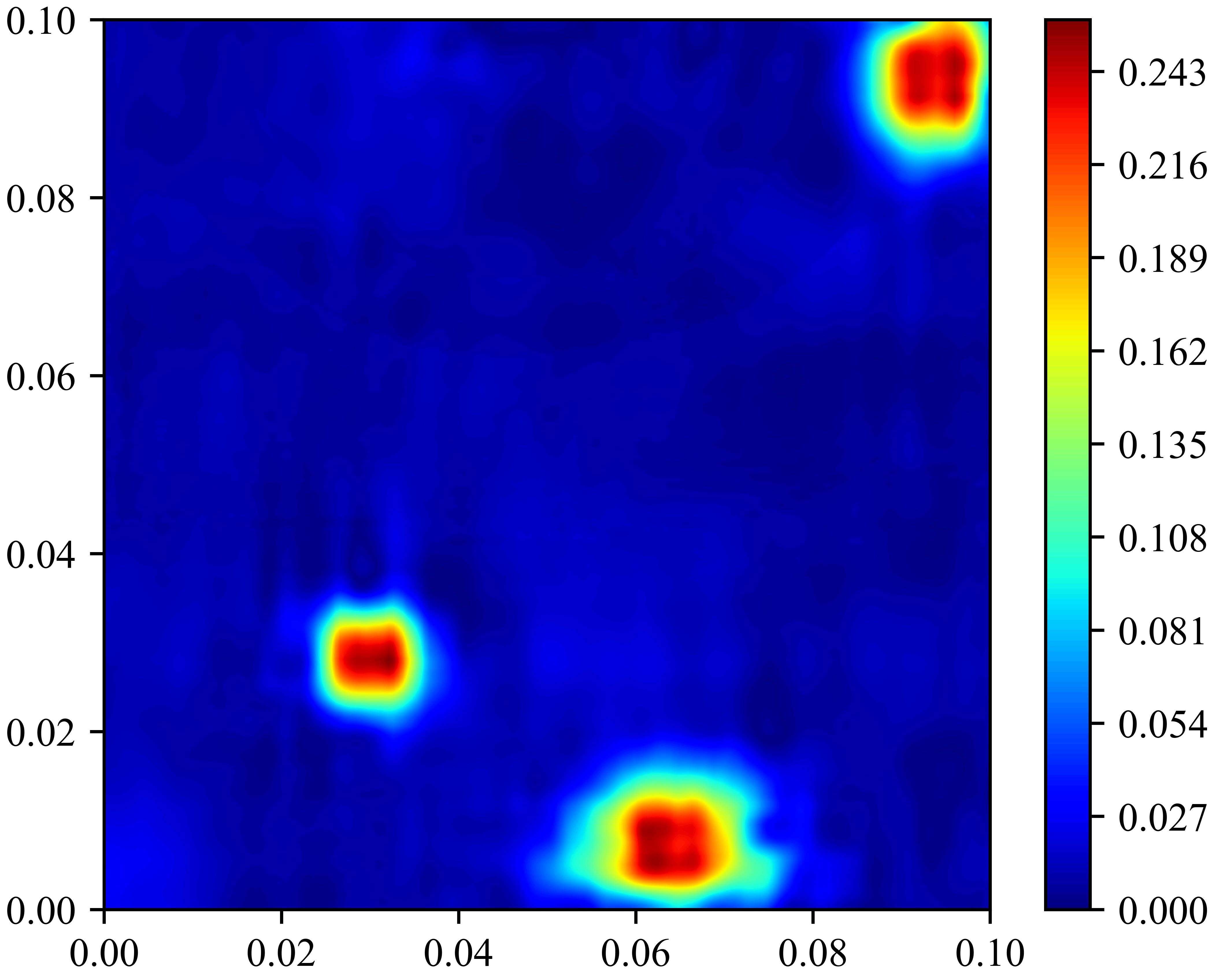}}
	\subfigure [Aleatoric uncertainty $\sigma_{2}^{{\varepsilon}_1}$]
	{\includegraphics[scale=0.383]{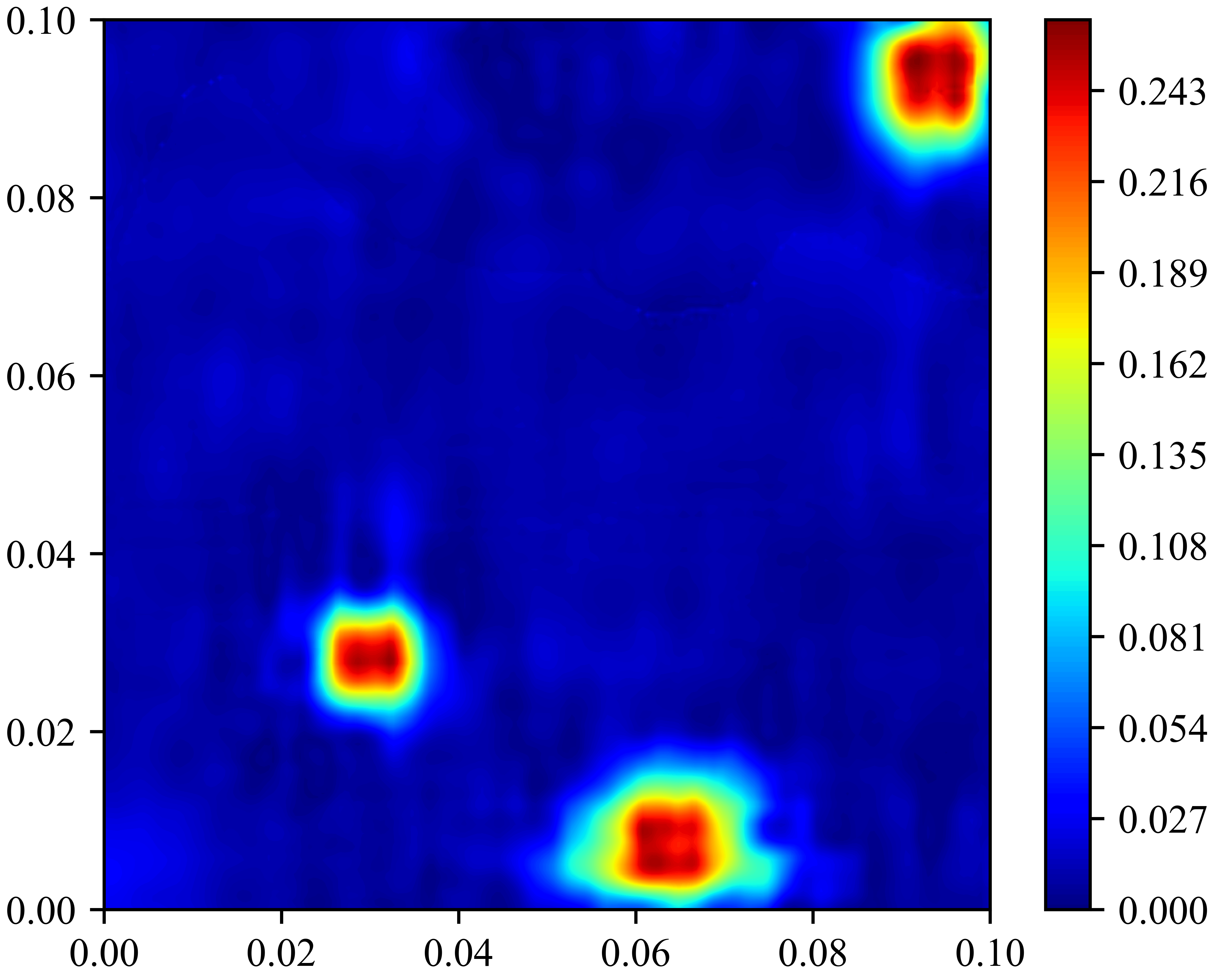}}
	\subfigure [Aleatoric uncertainty $\sigma_{3}^{{\varepsilon}_1}$]
	{\includegraphics[scale=0.383]{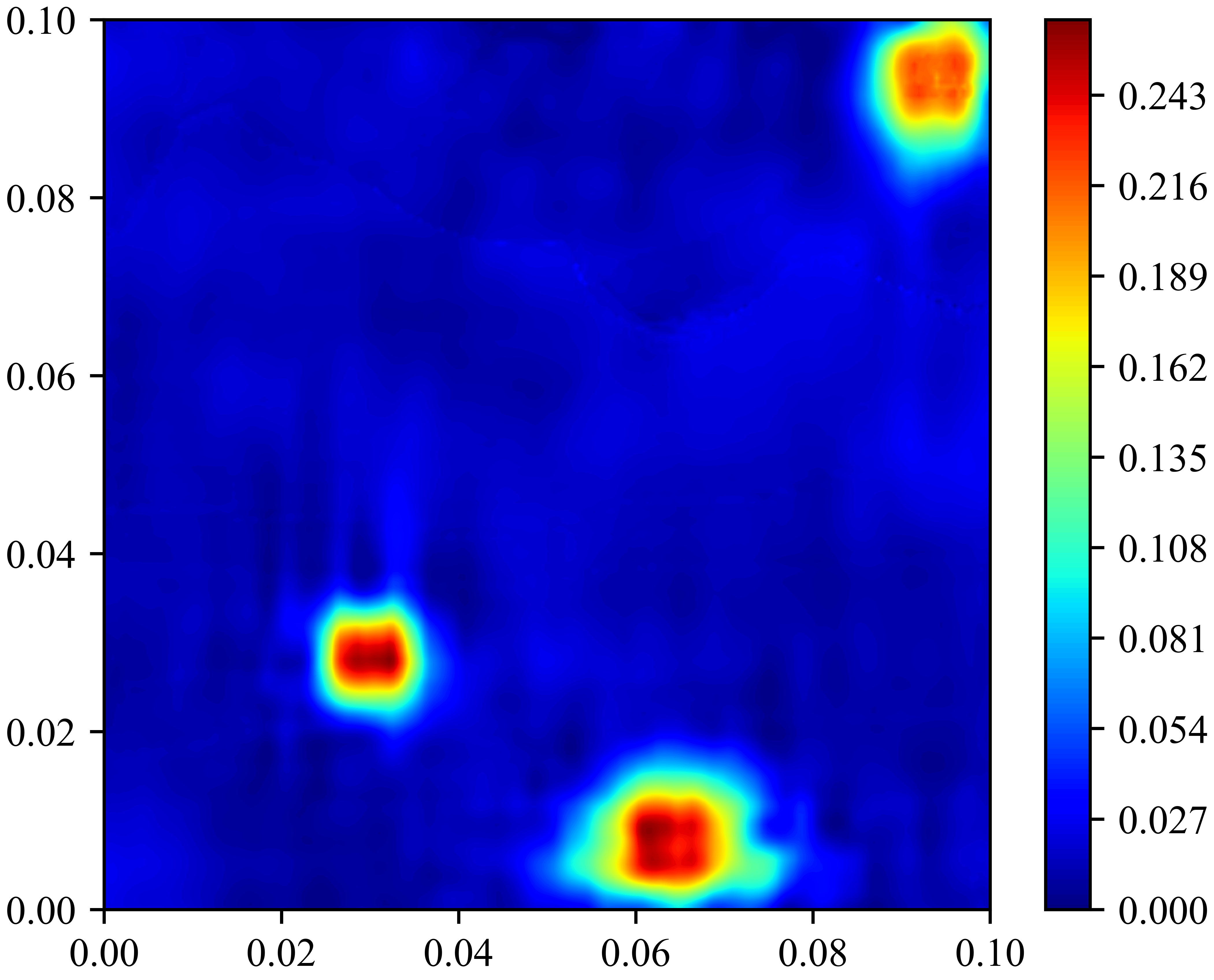}}
	\subfigure [Aleatoric uncertainty $\sigma_{4}^{{\varepsilon}_1}$]
	{\includegraphics[scale=0.383]{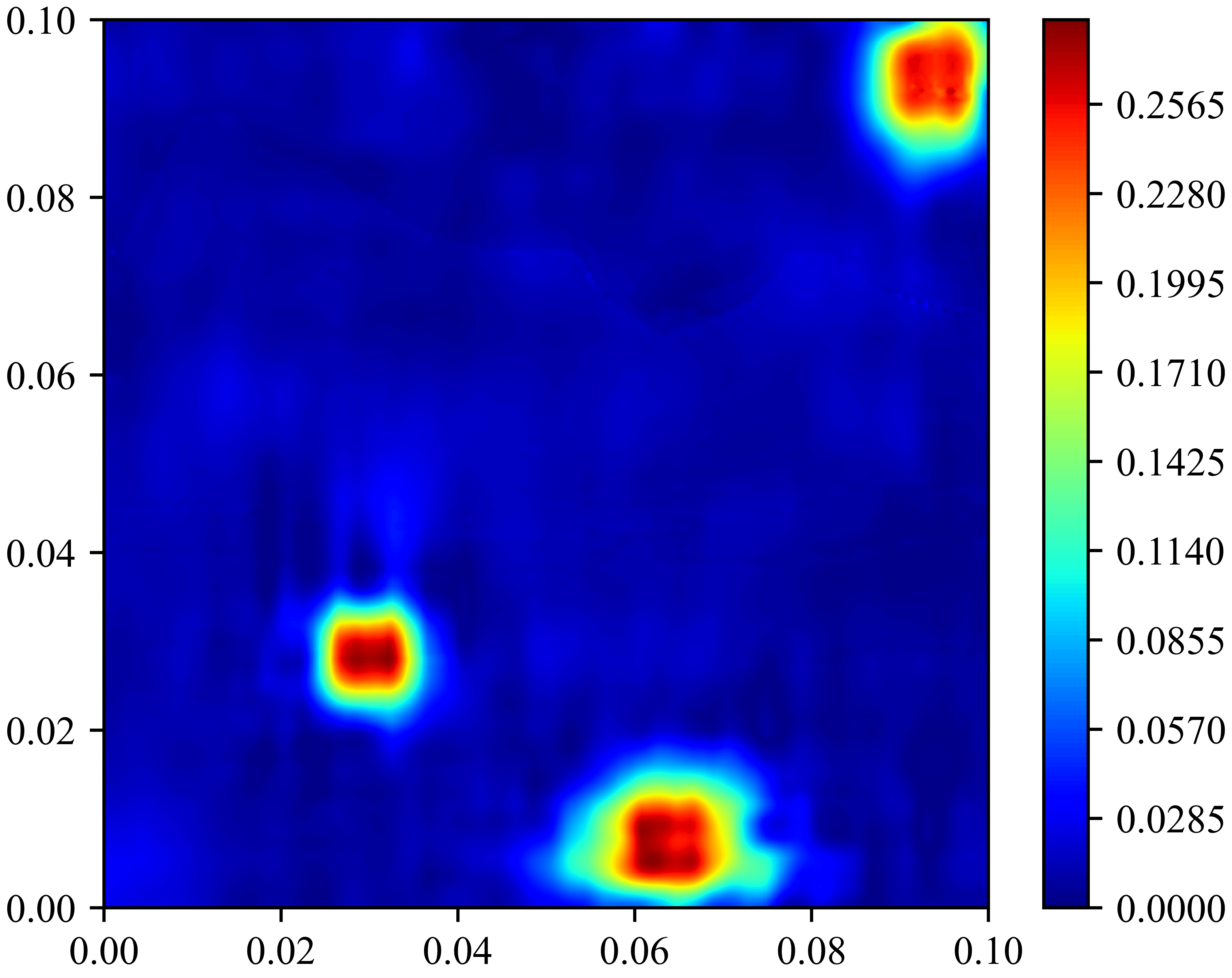}}
	\caption{The quantified aleatoric uncertainties $\left\{\sigma_{1}^{{\varepsilon}_1}, \sigma_{2}^{{\varepsilon}_1}, \sigma_{3}^{{\varepsilon}_1}, \sigma_{4}^{{\varepsilon}_1}\right\}$ for four MP temperatures $ \left\{\bm{T}^{1*}_{MP}, \bm{T}^{2*}_{MP}, \bm{T}^{3*}_{MP}, \bm{T}^{4*}_{MP}\right\} $ with the noise $ {\varepsilon}_1 \sim N\left ( 0, {0.25}^2  \right ) $ based on the model $\mathcal{M}^{{\varepsilon}_1}\left(\bm{T}_{MP}, \bm{\tau}; \bm{\theta}\right)$. }\label{alea005_case1}
\end{figure*}

\begin{figure*}[!htb]
	\centering
	\subfigure [Aleatoric uncertainty $\sigma_{1}^{{\varepsilon}_2}$]
	{\includegraphics[scale=0.383]{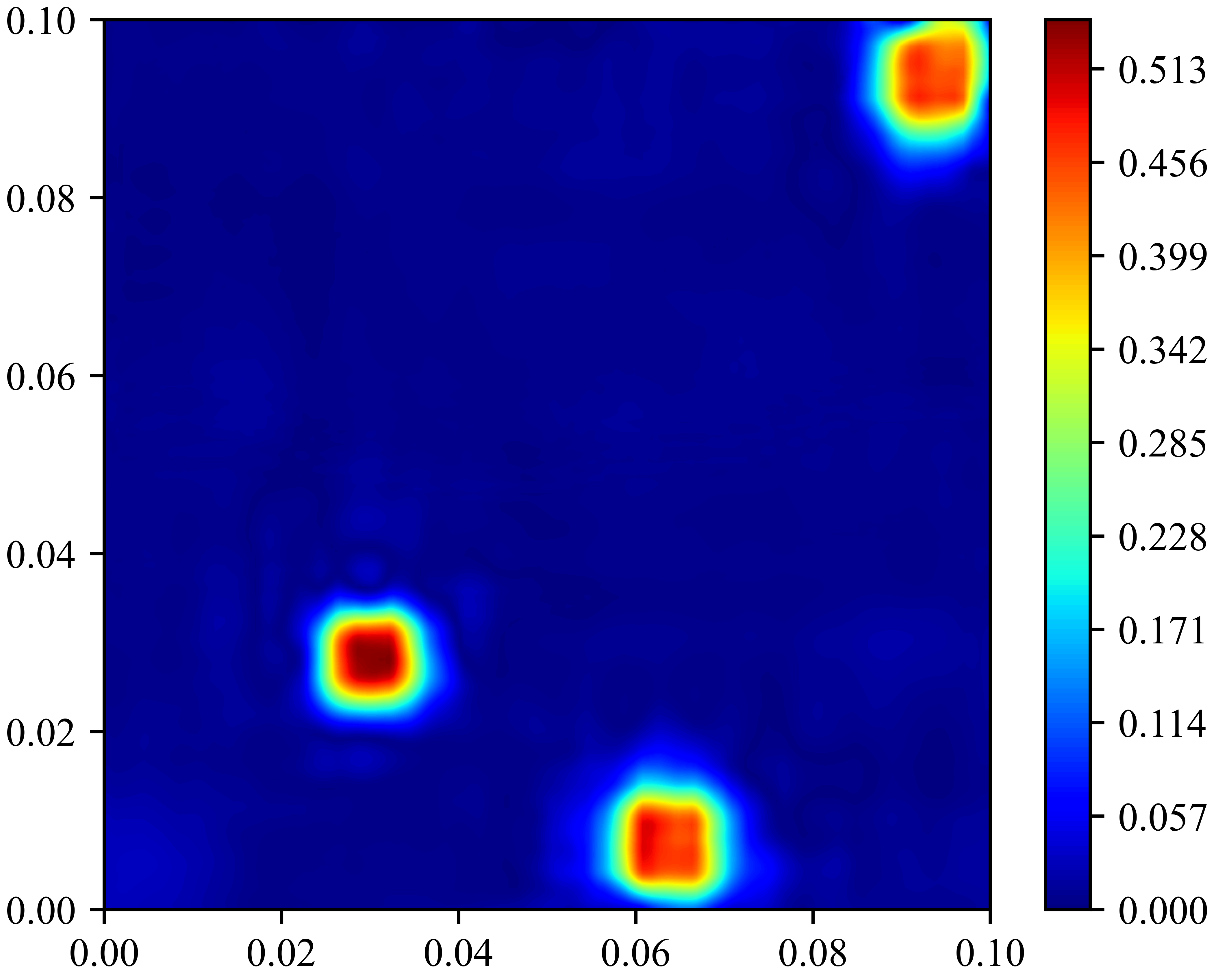}}
	\subfigure [Aleatoric uncertainty $\sigma_{2}^{{\varepsilon}_2}$]
	{\includegraphics[scale=0.383]{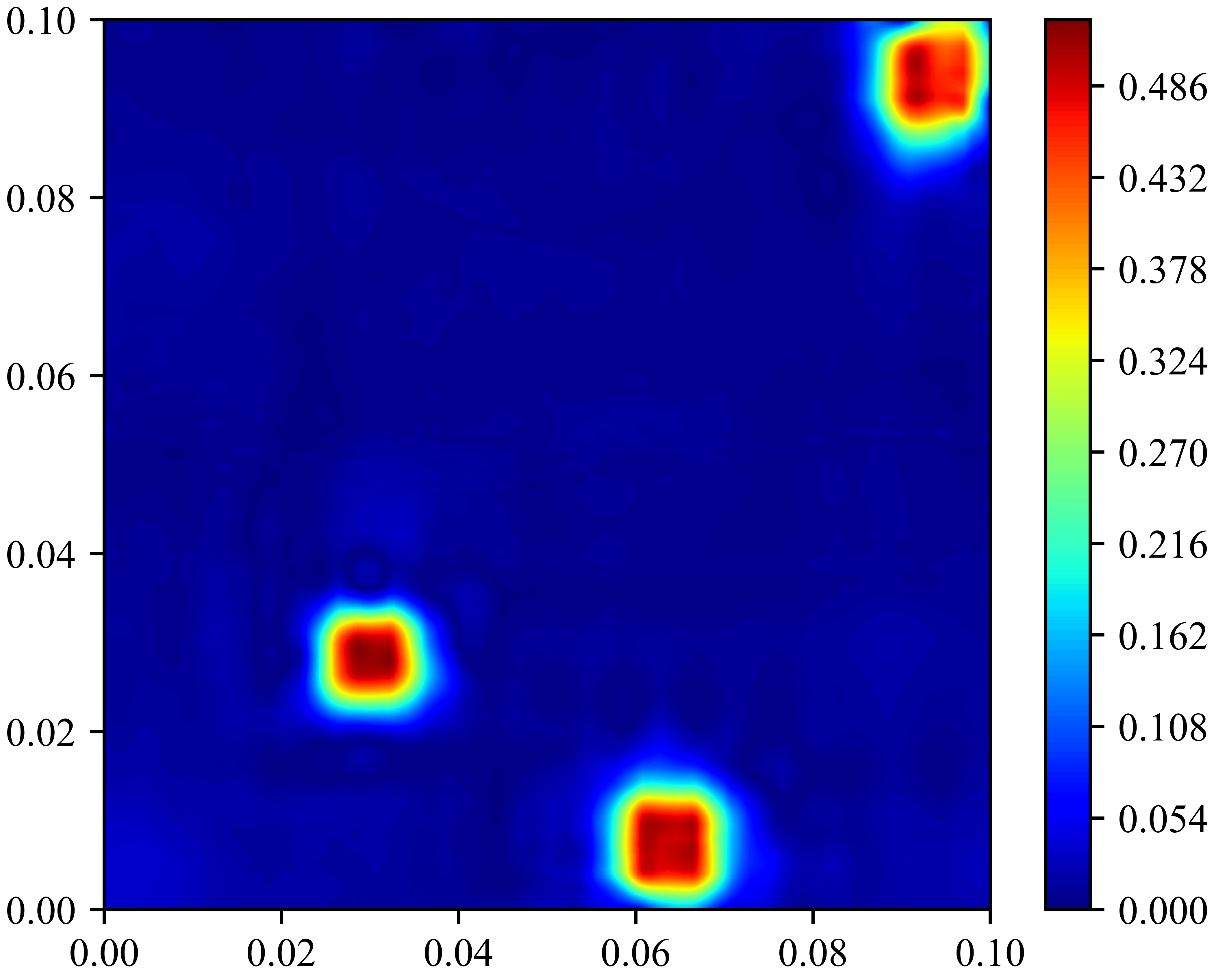}}
	\subfigure [Aleatoric uncertainty $\sigma_{3}^{{\varepsilon}_2}$]
	{\includegraphics[scale=0.383]{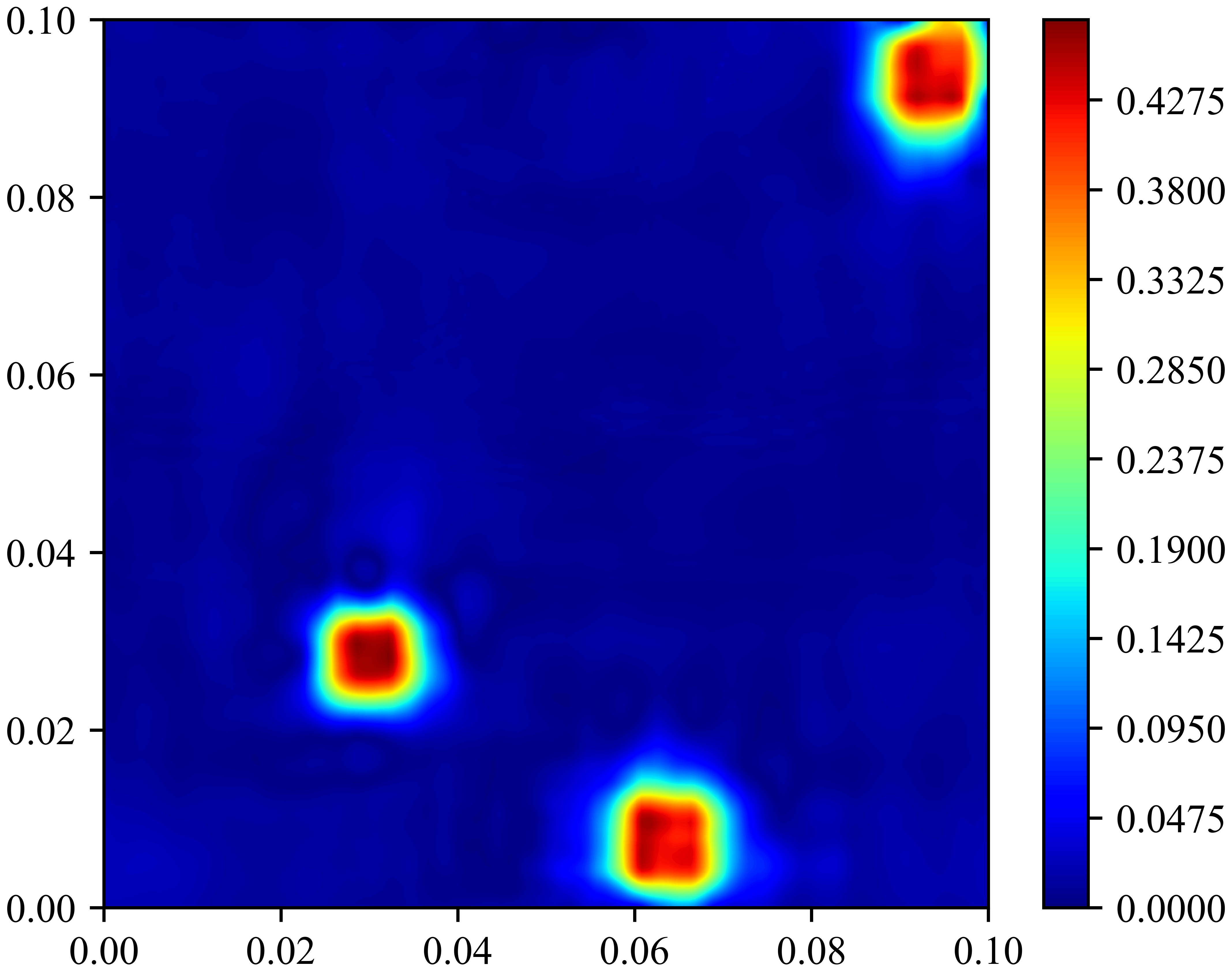}}
	\subfigure [Aleatoric uncertainty $\sigma_{4}^{{\varepsilon}_2}$]
	{\includegraphics[scale=0.383]{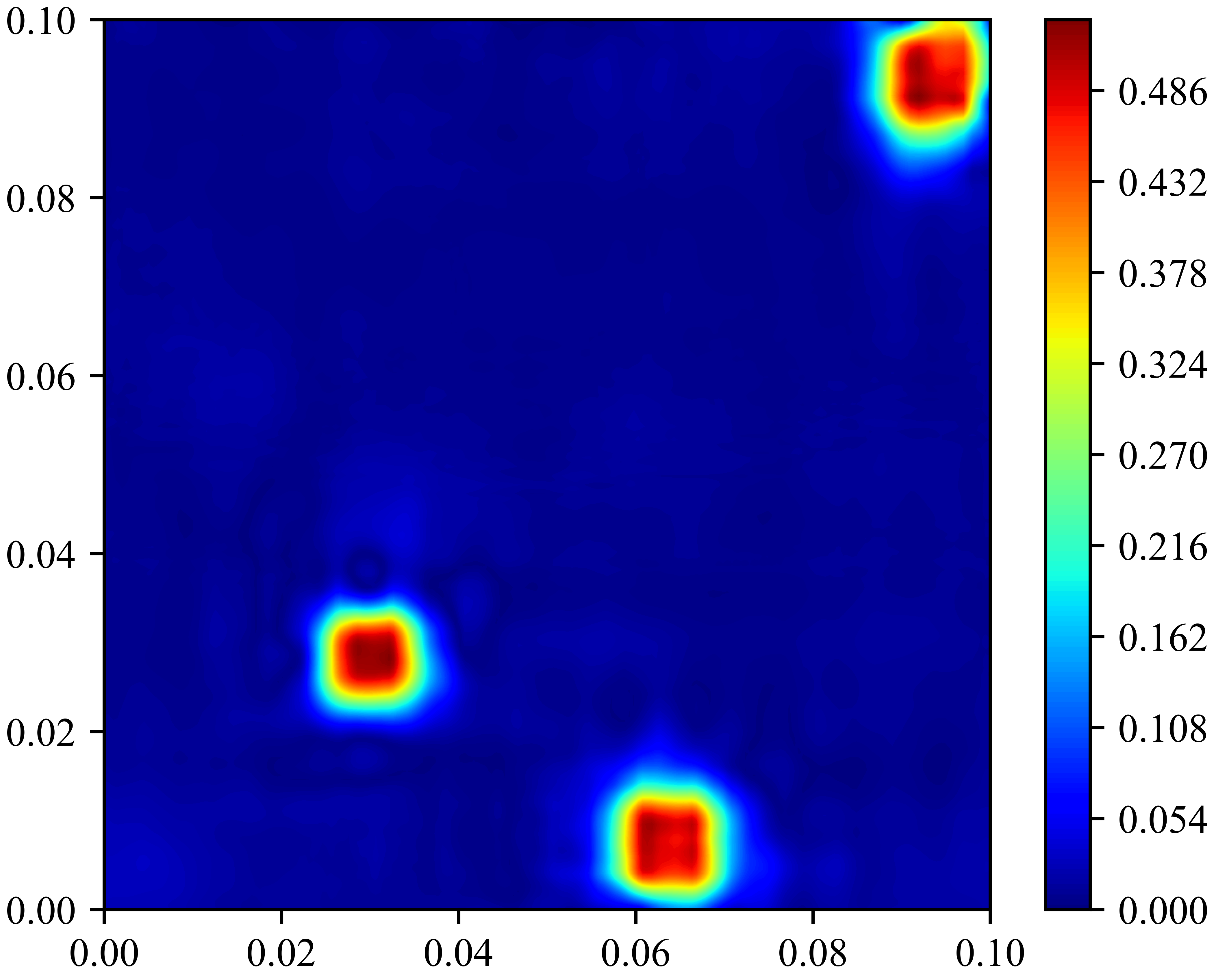}}
	\caption{The quantified aleatoric uncertainties $\left\{\sigma_{1}^{{\varepsilon}_2}, \sigma_{2}^{{\varepsilon}_2}, \sigma_{3}^{{\varepsilon}_2}, \sigma_{4}^{{\varepsilon}_2}\right\}$ for four MP temperatures $ \left\{\bm{T}^{1*}_{MP}, \bm{T}^{2*}_{MP}, \bm{T}^{3*}_{MP}, \bm{T}^{4*}_{MP}\right\} $ with the noise $ {\varepsilon}_1 \sim N\left ( 0, {0.5}^2  \right ) $ based on the model $\mathcal{M}^{{\varepsilon}_2}\left(\bm{T}_{MP}, \bm{\tau}; \bm{\theta}\right)$. }\label{alea01_case1}
\end{figure*}

\paragraph{\textbf{Aleatoric uncertainty quantification result analysis}} For four MP temperatures $ \left\{\bm{T}^{1*}_{MP}, \bm{T}^{2*}_{MP}, \bm{T}^{3*}_{MP}, \bm{T}^{4*}_{MP}\right\} $ (Fig.\ref{MP_case1}) with two kinds of Gaussian noises $ {\varepsilon}_1 \sim N\left ( 0, {0.25}^2  \right )  $ and $ {\varepsilon}_2 \sim N\left ( 0, {0.50}^2  \right )  $, the corresponding aleatoric uncertainties are quantified based on the models $\mathcal{M}^{{\varepsilon}_1}\left(\bm{T}_{MP}, \bm{\tau}; \bm{\theta}\right)$ and $\mathcal{M}^{{\varepsilon}_2}\left(\bm{T}_{MP}, \bm{\tau}; \bm{\theta}\right)$ as shown in Figs.\ref{alea005_case1} and \ref{alea01_case1}, respectively. According to section \ref{sedc511}, the models $\mathcal{M}^{{\varepsilon}_1}\left(\bm{T}_{MP}, \bm{\tau}; \bm{\theta}\right)$ and $\mathcal{M}^{{\varepsilon}_2}\left(\bm{T}_{MP}, \bm{\tau}; \bm{\theta}\right)$ are trained by the datasets that include the Gaussian noises ($ {\varepsilon}_1 \sim N\left ( 0, {0.25}^2  \right )  $, $ {\varepsilon}_2 \sim N\left ( 0, {0.50}^2  \right )  $) in the green points in Fig.\ref{example_case1}. Refer to Figs.\ref{alea005_case1} and \ref{alea01_case1}, the aleatoric uncertainties corresponding to the areas of the green points are much larger than the other areas. For the magnitudes of aleatoric uncertainties in Figs.\ref{alea005_case1} and \ref{alea01_case1}, the maximum values are approximately equal to 0.25 ($\mathcal{M}^{{\varepsilon}_1}\left(\bm{T}_{MP}, \bm{\tau}; \bm{\theta}\right)$) and 0.5 ($\mathcal{M}^{{\varepsilon}_2}\left(\bm{T}_{MP}, \bm{\tau}; \bm{\theta}\right)$), respectively. Apparently, the magnitudes of two aleatoric uncertainties are basically the same as the magnitudes of two Gaussian noises. Therefore, the proposed physics-informed Deep MC-QR method can quantify the aleatoric uncertainty accurately.

\begin{figure*}[!htb]
	\centering
	\subfigure [Aleatoric uncertainty $\sigma_{1}^{b{\varepsilon}_1}$]
	{\includegraphics[scale=0.383]{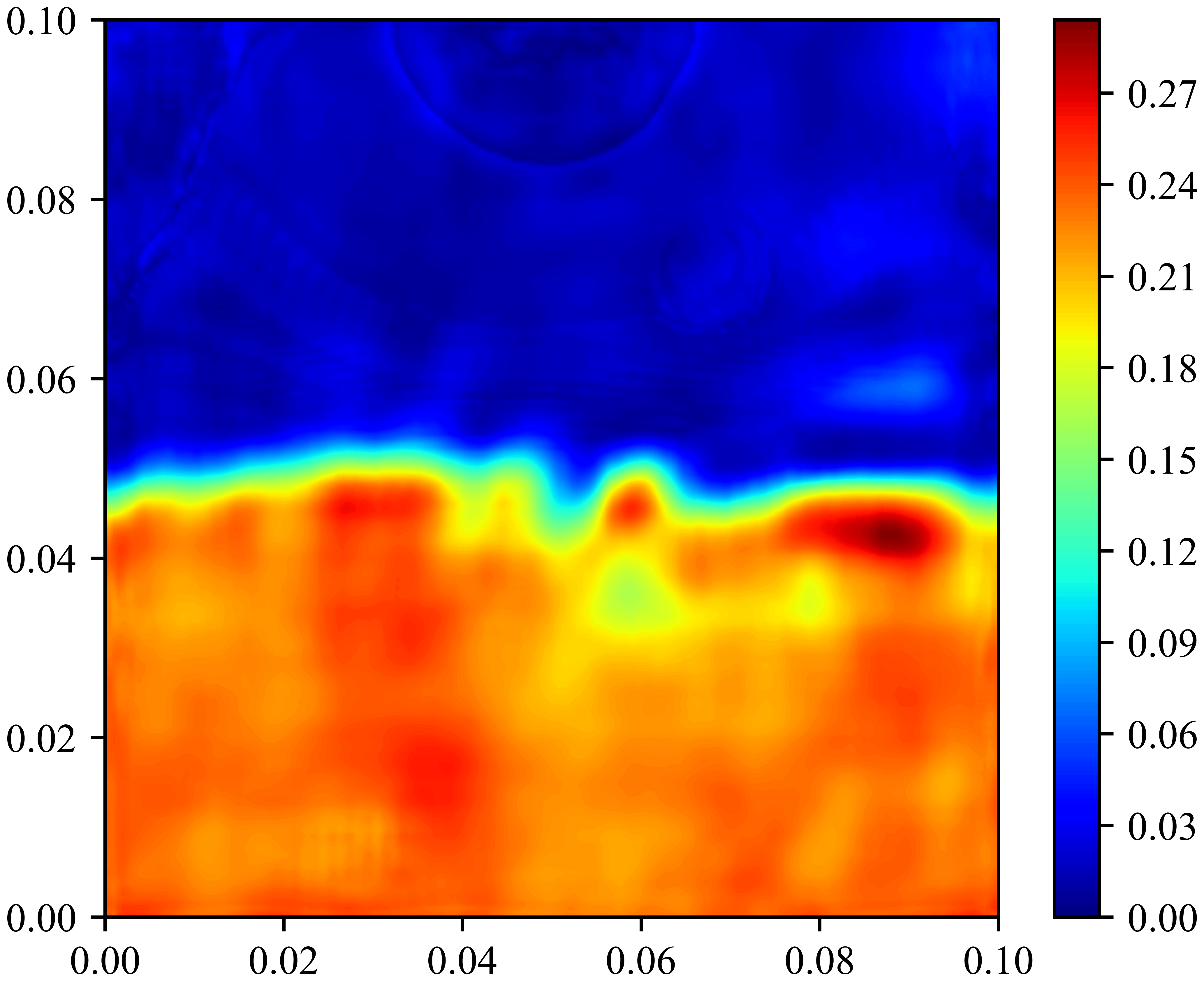}\label{buttom005_1}}
	\subfigure [Aleatoric uncertainty $\sigma_{2}^{b{\varepsilon}_1}$]
	{\includegraphics[scale=0.383]{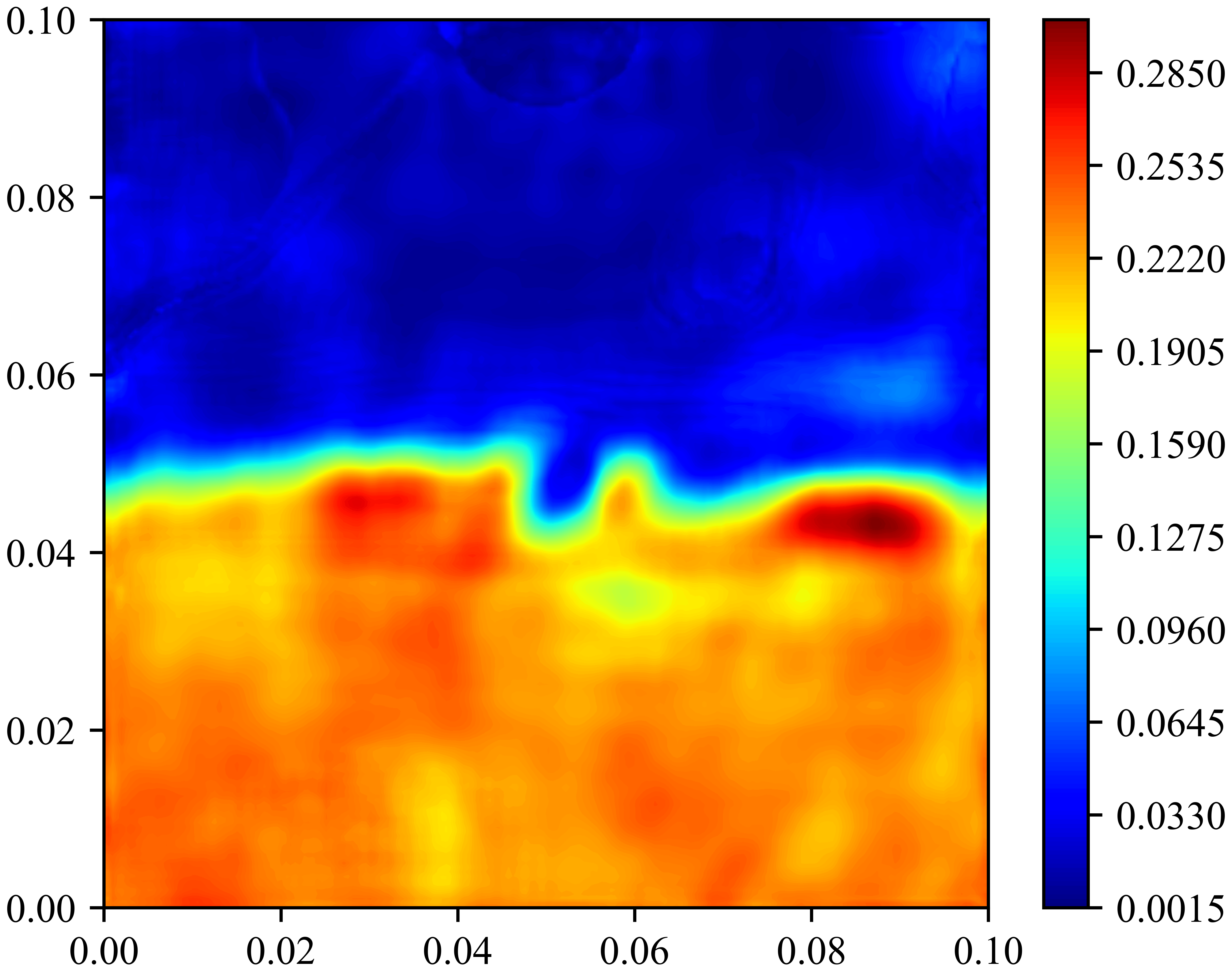}}
	\subfigure [Aleatoric uncertainty $\sigma_{3}^{b{\varepsilon}_1}$]
	{\includegraphics[scale=0.383]{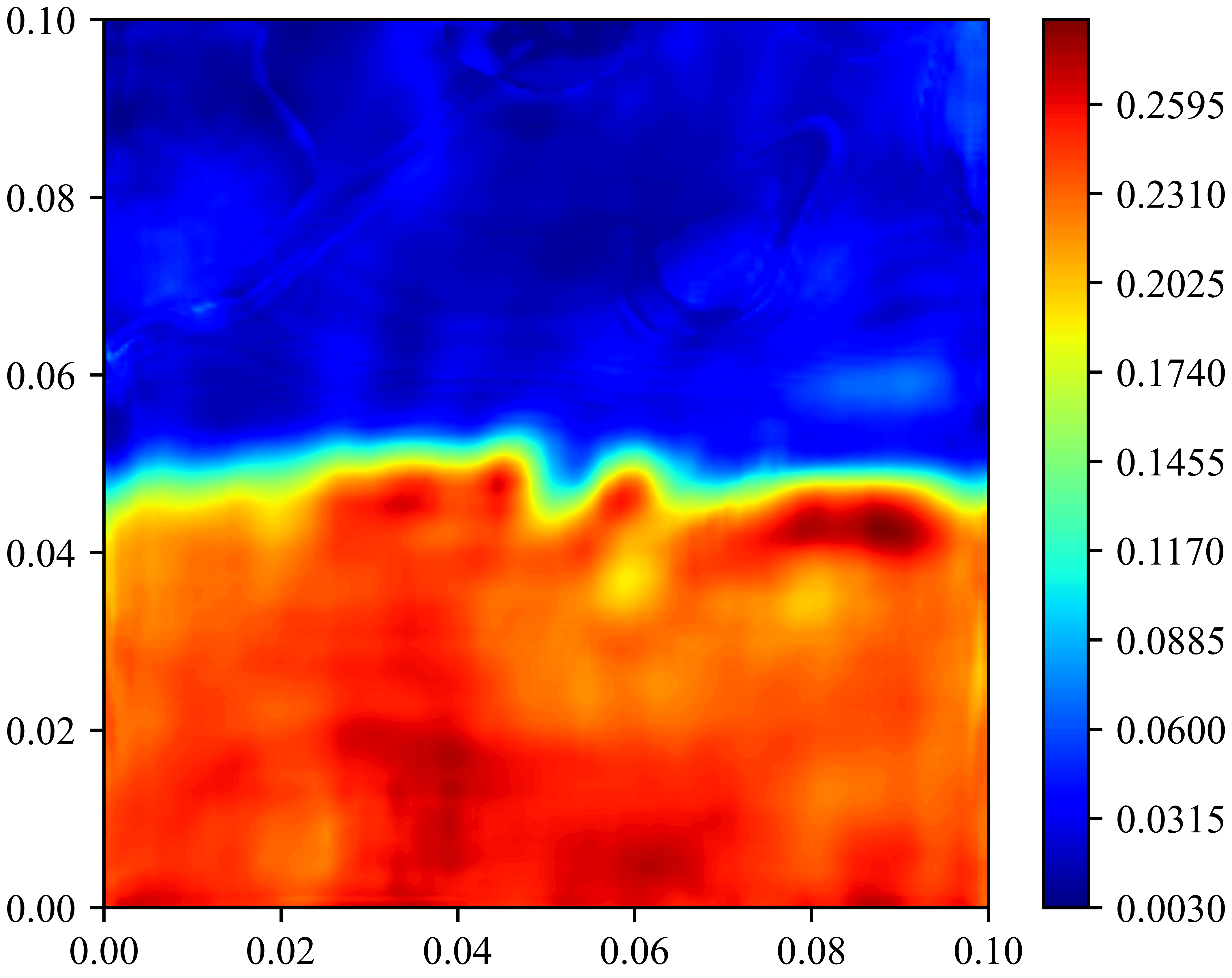}}
	\subfigure [Aleatoric uncertainty $\sigma_{4}^{b{\varepsilon}_1}$]
	{\includegraphics[scale=0.383]{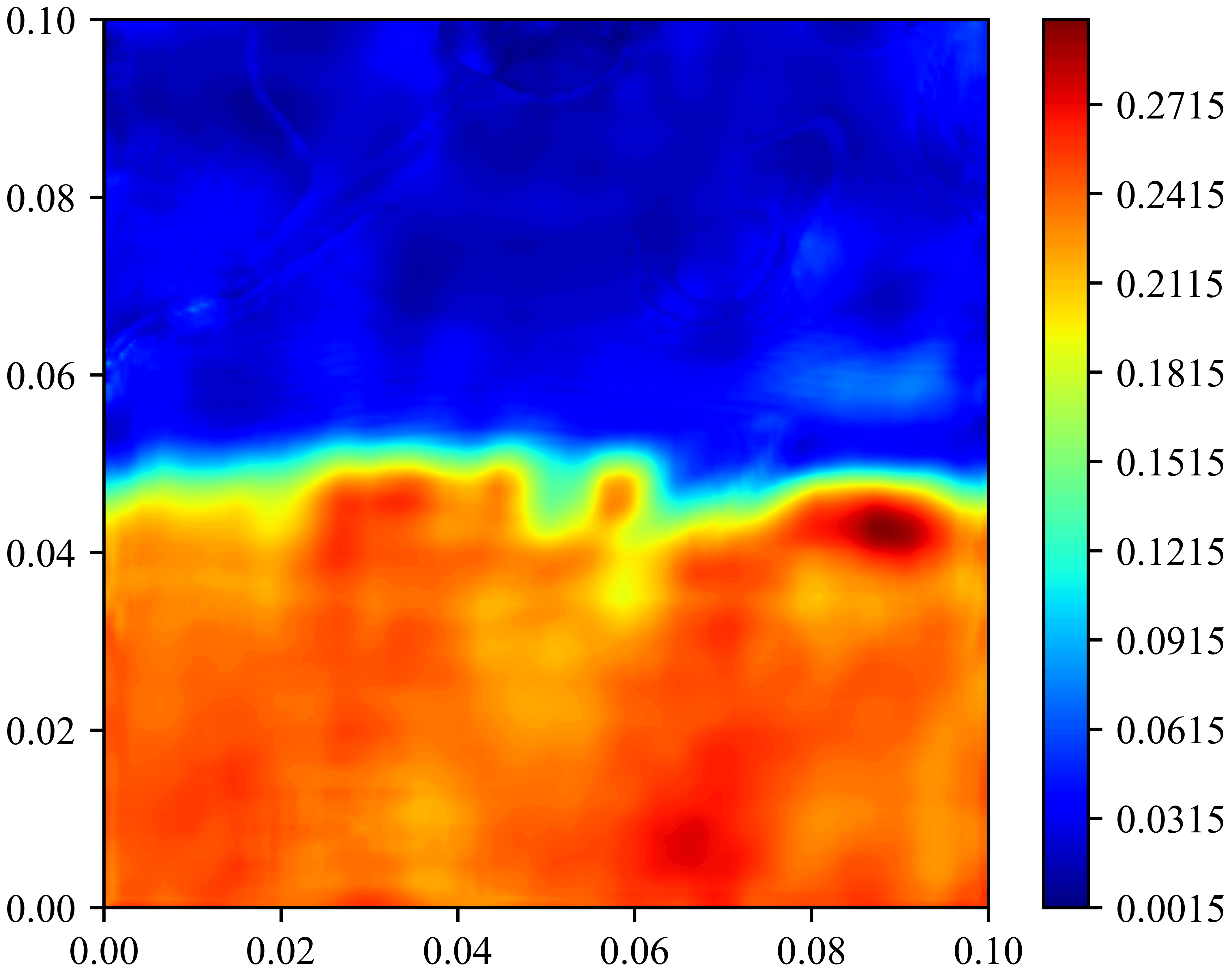}}
	\caption{The quantified aleatoric uncertainties $\left\{\sigma_{1}^{b{\varepsilon}_1}, \sigma_{2}^{b{\varepsilon}_1}, \sigma_{3}^{b{\varepsilon}_1}, \sigma_{4}^{b{\varepsilon}_1}\right\}$ for four MP temperatures $ \left\{\bm{T}^{1*}_{MP}, \bm{T}^{2*}_{MP}, \bm{T}^{3*}_{MP}, \bm{T}^{4*}_{MP}\right\} $ with the noise $ {\varepsilon}_1 \sim N\left ( 0, {0.25}^2  \right ) $ based on the model $\mathcal{M}^{b{\varepsilon}_1}\left(\bm{T}_{MP}, \bm{\tau}; \bm{\theta}\right)$. }\label{alea005buttom_case1}
\end{figure*}

\paragraph{\textbf{Necessity analysis of aleatoric uncertainty quantification}} For four MP temperatures (Fig.\ref{MP_case1}) whose monitoring points below the blue line in Fig.\ref{example_case1} include the Gaussian noise $ {\varepsilon}_1 \sim N\left ( 0, {0.25}^2  \right )  $, the corresponding aleatoric uncertainties are quantified based on the model $\mathcal{M}^{b{\varepsilon}_1}\left(\bm{T}_{MP}, \bm{\tau}; \bm{\theta}\right)$ as shown in Figs.\ref{alea005buttom_case1}. According to section \ref{sedc511}, the model $\mathcal{M}^{b{\varepsilon}_1}\left(\bm{T}_{MP}, \bm{\tau}; \bm{\theta}\right)$ is trained by the dataset that includes the Gaussian noise $ {\varepsilon}_1 \sim N\left ( 0, {0.25}^2  \right )  $ in the monitoring points below the blue line in Fig.\ref{example_case1}. 

Refer to Fig.\ref{alea005buttom_case1}, the aleatoric uncertainties corresponding to the area below the blue line are much larger than the area above the blue line. There are three pseudo-abnormal phenomenons in Fig.\ref{alea005buttom_case1} as follows:

(1) \textbf{The maximum aleatoric uncertainty is abnormal.} As shown in Fig.\ref{alea005buttom_case1}, the max value of quantified aleatoric uncertainty is approximately equal to 0.3. However, the value of actual Gaussian noise is equal to 0.25.

\begin{figure*}[!htb]
	\centering
	{\includegraphics[scale=0.55]{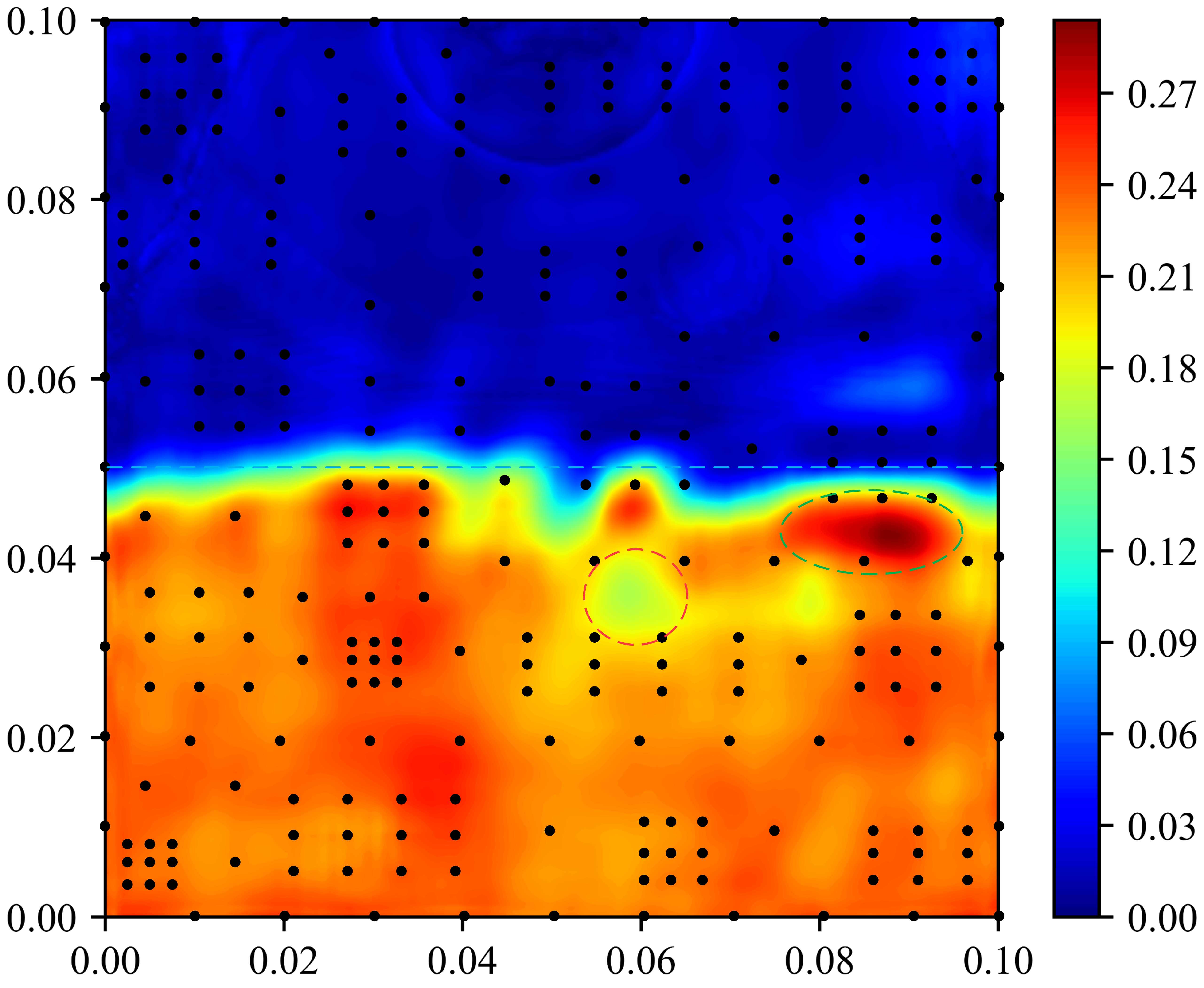}}
	\caption{The fusion figure between the monitoring point position (Fig.\ref{example_case1}) and the quantified aleatoric uncertainty $\sigma_{1}^{b{\varepsilon}_1}$ (Fig. \ref{buttom005_1}). }\label{alea_MP}
\end{figure*}

(2) \textbf{The aleatoric uncertainties of the non-monitoring point area below the blue line are abnormal.} It is noteworthy that the aleatoric uncertainties of the non-monitoring point area below the blue line are also relatively large, as shown in Fig.\ref{alea_MP}. However, the non-monitoring point area below the blue line does not contain Gaussian noise. 

(3) \textbf{It is abnormal for the magnitude of aleatoric uncertainty in some non-monitoring point areas to have large differences.} As shown in Fig.\ref{alea_MP}, the aleatoric uncertainty in the green dashed circle is larger than in the red dashed circle for the non-monitoring point area. However, there is no Gaussian noise in these two areas.

In fact, the above three pseudo-abnormal phenomenons are caused by the spread of data noise deviation. For the first pseudo-abnormal phenomenon, the spread of data noise leads to the accumulation of aleatoric uncertainty. Thereby, the maximum aleatoric uncertainty is larger than the actual Gaussian noise.  For the second pseudo-abnormal phenomenon, the spread of data noise will transfer the aleatoric uncertainty to the surrounding area. Thus, the non-monitoring point area below the blue line contains aleatoric uncertainty. When the sensor measures temperature, the measured value is sometimes relatively large (positive Gaussian noise), and sometimes the measured value is relatively small (negative Gaussian noise). Thus for the third pseudo-abnormal phenomenon, the accumulation of positive or negative Gaussian noise will lead to the magnitude of aleatoric uncertainty in some non-monitoring point areas to large differences. For one thing, suppose the Gaussian noise of the measuring point in the area around the non-monitoring point is only positive or negative. In that case, the aleatoric uncertainty of the non-monitoring point area will be relatively large. For the other thing, suppose the Gaussian noise of the measuring point in the area around the non-monitoring point contains both positive and negative. In that case, the aleatoric uncertainty of the non-monitoring point area will be relatively small due to partial noise cancellation.

In summary, the data noise of the monitoring point not only causes aleatoric uncertainty in its temperature prediction value but also leads to aleatoric uncertainty in the temperature prediction value of the surrounding area. Therefore, aleatoric uncertainty quantification is essential to ensure the credibility of the reconstructed temperature field for the satellite TFR problem. 

\subsection{Case 2: A satellite subsystem heat reliability analysis}\label{sec52}
\subsubsection{Case background}
\begin{figure*}[!htb]
	\centering
	{\includegraphics[scale=1]{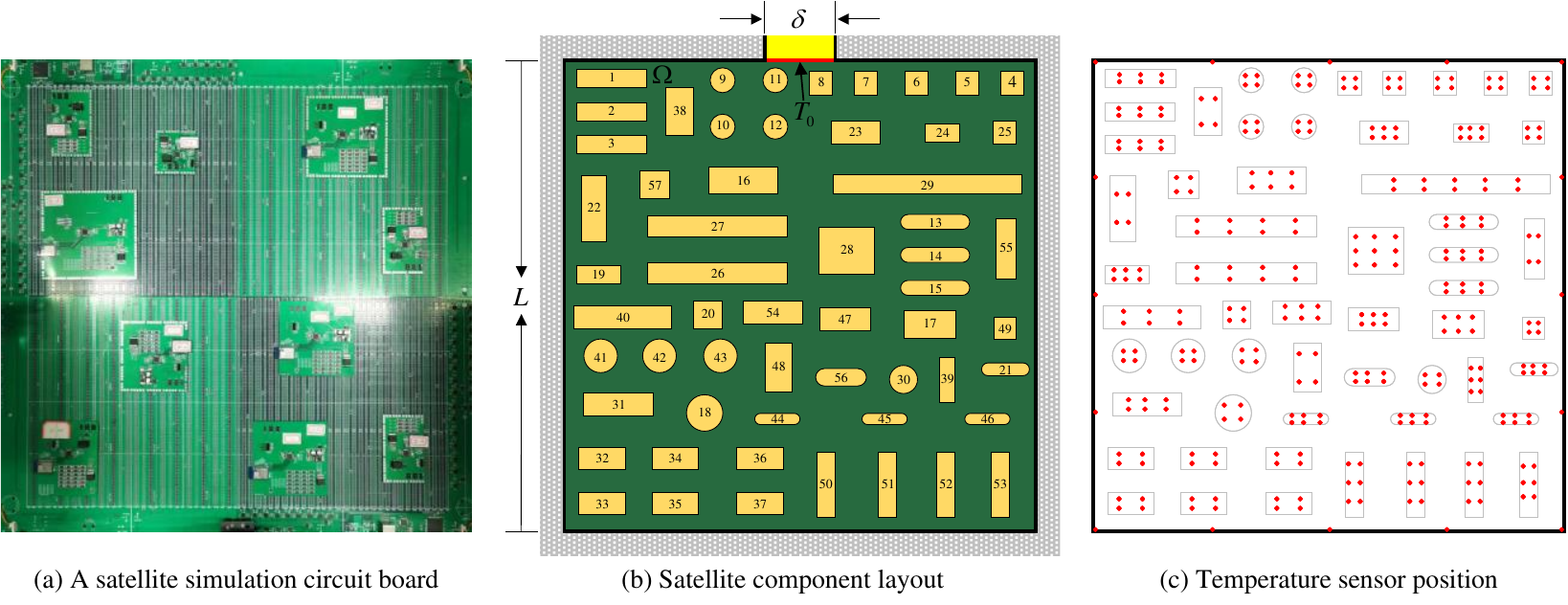}}
	\caption{A satellite subsystem with 57 components in case 2.}\label{Layout_SAT}
\end{figure*}

\begin{figure*}[!htb]
	\centering
	{\includegraphics[scale=0.8]{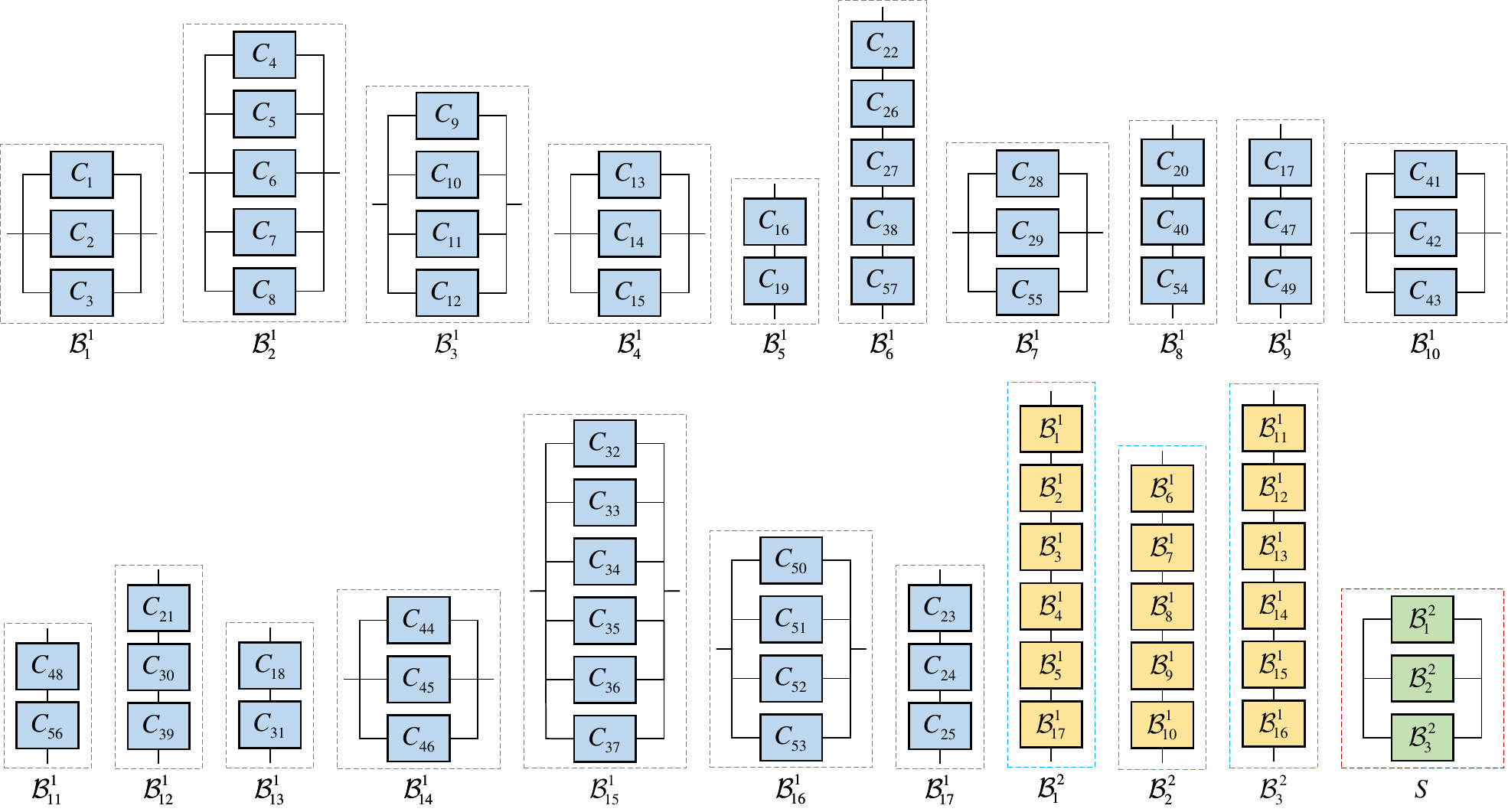}}
	\caption{The logical relationships between satellite components ($ C $), blocks ($\mathcal{B}$), and satellite subsystem ($S$).}\label{Case2_com_frame}
\end{figure*}

The second case analyzes the heat reliability of a satellite subsystem. As shown in Fig.\ref{Layout_SAT}, Fig.\ref{Layout_SAT} (a) is the satellite subsystem circuit board for installing 57 components, and Fig.\ref{Layout_SAT} (b) is the satellite component layout. The length $ L $ is equal to 0.8m and the width $\delta$ of the heat sink is 0.112m. A total of 313 temperature monitoring points are arranged in this satellite subsystem, where the positions of all temperature monitoring points are shown in Fig.\ref{Com_PD_case2} (c). As shown in Fig.\ref{Case2_com_frame}, the satellite subsystem $S$ includes 20 blocks, and each block is composed of some satellite components in a certain logical relationship. For example, the block $\mathcal{B}_{1}^{1}$ consists of components $\left\{C_1, C_2, C_3\right\}$ by the parallel relationship, the block $\mathcal{B}_{5}^{1}$ is composed of components $\left\{C_{16}, C_{19}\right\}$ by the series relationships. Especially, the blocks $\mathcal{B}_{1}^{2}$, $\mathcal{B}_{2}^{2}$, and $\mathcal{B}_{3}^{2}$ are composed of blocks $\left\{\mathcal{B}_{1}^{1}, \mathcal{B}_{2}^{1}, \mathcal{B}_{3}^{1}, \mathcal{B}_{4}^{1}, \mathcal{B}_{5}^{1}, \mathcal{B}_{17}^{1}\right\}$, $\left\{\mathcal{B}_{6}^{1}, \mathcal{B}_{7}^{1}, \mathcal{B}_{8}^{1}, \mathcal{B}_{9}^{1}, \mathcal{B}_{10}^{1}\right\}$,  and $\left\{\mathcal{B}_{11}^{1}, \mathcal{B}_{12}^{1}, \mathcal{B}_{13}^{1}, \mathcal{B}_{14}^{1}, \mathcal{B}_{15}^{1}, \mathcal{B}_{16}^{1}\right\}$ in the series relationship, respectively. The satellite subsystem $S$ consists of three blocks $\mathcal{B}_{1}^{2}$, $\mathcal{B}_{2}^{2}$, and $\mathcal{B}_{3}^{2}$ by the parallel relationship. The probability distributions of satellite component powers are shown in Table \ref{Com_PD_case2}. Besides, the working state thresholds of 57 satellite components are also presented in Table \ref{Com_PD_case2}.

\begin{table}[!htb]
	\centering
	\caption{All satellite component powers' probability distributions and working state thresholds in case 2.}
	\begin{tabular}{cllll|cllll}
		\toprule
		S.C. & Mean  & S.D. & Distribution & Threshold & S.C. & Mean  & S.D. & Distribution & Threshold \\
		\midrule
		$C_1$    & 195 & 8.3  & Normal    & 322.60 & $C_{31}$ & 180 & 6.5   & Normal  & 330.10 \\
		$C_2$    & 195 & 8.3  & Normal    & 322.90 &  $C_{32}$ & 197 & 13.8  & Normal & 330.80 \\
		$C_3$    & 195 & 8.3  & Normal    & 323.60 &  $C_{33}$ & 197 & 13.8  & Normal & 333.10 \\
		$C_4$    & 136 & 5.7  & Normal    & 321.05 &  $C_{34}$ & 197 & 13.8  & Normal & 330.70 \\
		$C_5$    & 136 & 5.7  & Normal    & 320.30 &  $C_{35}$ & 197 & 13.8  & Normal & 331.00 \\
		$C_6$    & 136 & 5.7  & Normal    & 318.10 &  $C_{36}$ & 197 & 13.8  & Normal & 330.30 \\
		$C_7$    & 136 & 5.7  & Normal    & 313.82 &  $C_{37}$ & 197 & 13.8  & Normal & 330.60 \\
		$C_8$    & 136 & 5.7  & Normal    & 308.65 &  $C_{38}$ & 164 & 5.8   & Normal & 321.00 \\
		$C_{13}$ & 186 & 8.8  & Lognormal & 324.30 &  $C_{39}$ & 167 & 5.1   & Gumbel & 328.60 \\
		$C_{14}$ & 186 & 8.8  & Lognormal & 325.40 &  $C_{41}$ & 138 & 2.9   & Normal & 329.20 \\
		$C_{15}$ & 186 & 8.8  & Lognormal & 326.40 &  $C_{42}$ & 138 & 2.9   & Normal & 329.00 \\
		$C_{16}$ & 172 & 7.8  & Gumbel    & 322.30 &  $C_{43}$ & 138 & 2.9   & Normal & 328.90 \\
		$C_{17}$ & 172 & 7.8  & Gumbel    & 327.50 &  $C_{47}$ & 194 & 6.4   & Normal & 327.30 \\
		$C_{18}$ & 167 & 6.3  & Normal    & 330.00 &  $C_{48}$ & 87  & 5.7   & Normal & 329.00 \\
		$C_{20}$ & 139 & 5.4  & Lognormal & 328.00 &  $C_{49}$ & 134 & 2.8   & Normal & 327.50 \\
		$C_{21}$ & 135 & 4.1  & Normal    & 328.30 &  $C_{50}$ & 193 & 17.1  & Lognormal & 330.20 \\
		$C_{22}$ & 96  & 5.2  & Gumbel    & 325.80 &  $C_{51}$ & 193 & 17.1  & Lognormal & 330.00 \\
		$C_{23}$ & 168 & 4.9  & Lognormal & 317.90 &  $C_{52}$ & 193 & 17.1  & Lognormal & 329.90 \\
		$C_{25}$ & 141 & 3.9  & Normal    & 321.90 &  $C_{53}$ & 193 & 17.1  & Lognormal & 329.70 \\
		$C_{26}$ & 218 & 6.6  & Normal    & 326.80 &  $C_{54}$ & 159 & 3.5   & Normal & 327.40 \\
		$C_{27}$ & 218 & 6.6   & Normal   & 325.30 &  $C_{55}$ & 147 & 4.2   & Normal & 326.00 \\
		$C_{28}$ & 202 & 8.9   & Normal   & 325.50 &  $C_{57}$ & 137 & 6.4   & Gumbel & 324.00 \\
		$C_{29}$ & 313 & 9.7   & Gumbel   & 323.50 &   &  &   &  &  \\
		\midrule
		S.C. & L.B.  & U.B. & Distribution & Threshold & S.C. & L.B.  & U.B. & Distribution & Threshold \\
		\midrule
		$C_{9}$  & 82  & 121 & Uniform & 316.10 & $C_{30}$  & 60  & 133  & Uniform & 328.60 \\
		$C_{10}$ & 82  & 121 & Uniform & 318.30 & $C_{40}$  & 110 & 180  & Uniform & 328.40 \\
		$C_{11}$ & 82  & 121 & Uniform & 309.03 & $C_{44}$  & 78  & 143  & Uniform & 329.60 \\
		$C_{12}$ & 82  & 121 & Uniform & 315.50 & $C_{45}$  & 78  & 143  & Uniform & 329.20 \\
		$C_{19}$ & 67  & 139 & Uniform & 327.00 & $C_{46}$  & 78  & 143  & Uniform & 329.00 \\
		$C_{24}$ & 90  & 147 & Uniform & 321.00 & $C_{56}$  & 121 & 160  & Uniform & 328.70 \\
		\bottomrule
		\multicolumn{10}{l}{S.C. = Satellite component \qquad S.D. = Standard deviation \qquad L.B. = Lower boundary} \\
		\multicolumn{10}{l}{U.B. = Upper boundary}
	\end{tabular}
	\label{Com_PD_case2}
\end{table}

\subsubsection{Deep MC-QR model training and results analysis}
\paragraph{\textbf{Preparing data and training model}} According to the component powers' probability distributions, this case samples 10000 sets of component powers by the Latin Hypercube Sampling, and based on this, 10000 satellite subsystem operating conditions are set. Then,  10000 ground simulation experiments are carried out based on a satellite simulation system. For each simulation experiment, these temperature values of all points in Fig.\ref{Com_PD_case2} (c) are measured by 313 sensors when the satellite simulation system arrives at steady-state. Finally, the ground simulation experiment can obtain 10000 MP temperatures $\left\{\bm{T}^1_{MP}, \bm{T}^2_{MP}, \cdots ,\bm{T}^{10000}_{MP}\right\}$. The sensor measurement deviation is about $\pm0.15 \text{K}$. In this case, 8100 MP temperatures are used to train the physics-informed Deep MC-QR model $\mathcal{M}\left(\bm{T}_{MP}, \bm{\tau}; \bm{\theta}\right)$, 900  MP temperatures are used to be the validate data, and the remaining 1000 MP temperatures are used to test the trained model $\mathcal{M}\left(\bm{T}_{MP}, \bm{\tau}; \bm{\theta}\right)$. For the 1000 testing MP temperatures, the corresponding approximate truth temperature fields can be obtained by the infrared camera. In this case, the maximum training epoch and the learning rate are set to be 80 and 0.1, respectively. Based on the prepared training dataset, the trained model $\mathcal{M}\left(\bm{T}_{MP}\right)$ is obtained by the \textbf{Algorithm} \ref{algorithm1}. 

\begin{figure*}[!htb]
	\centering
	\subfigure [MP temperature]
	{\includegraphics[scale=0.4]{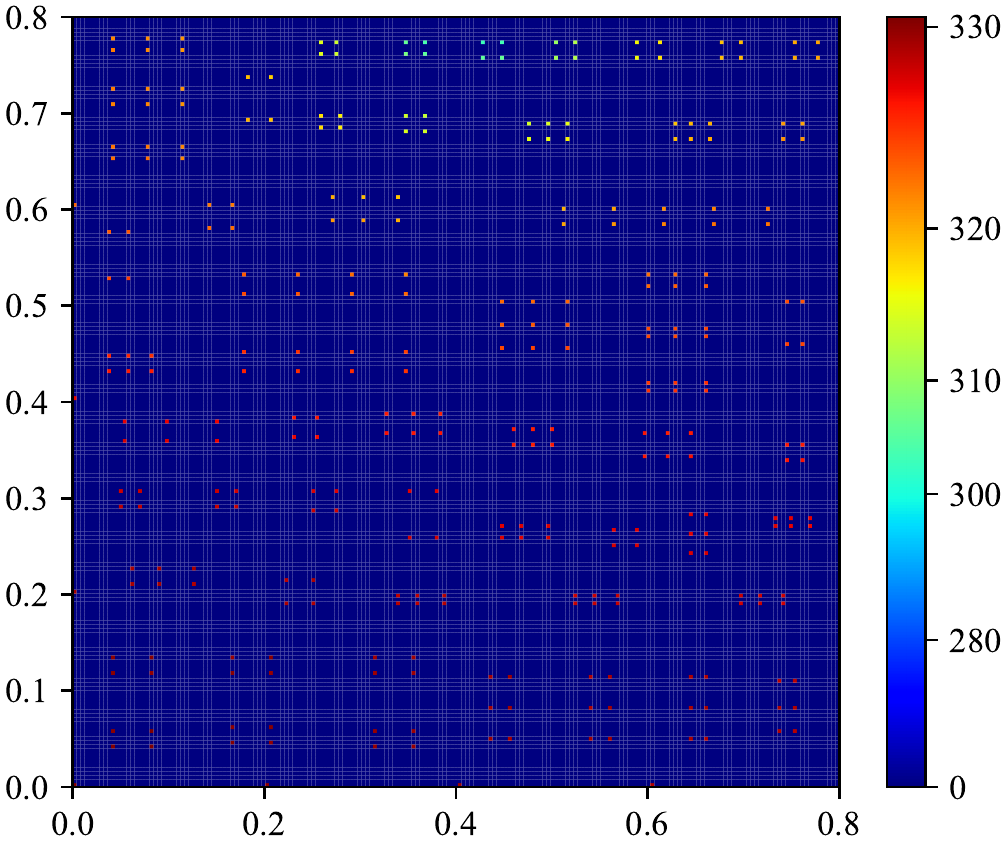}\label{MP_case2}}
	\subfigure [Truth temperature field]
	{\includegraphics[scale=0.4]{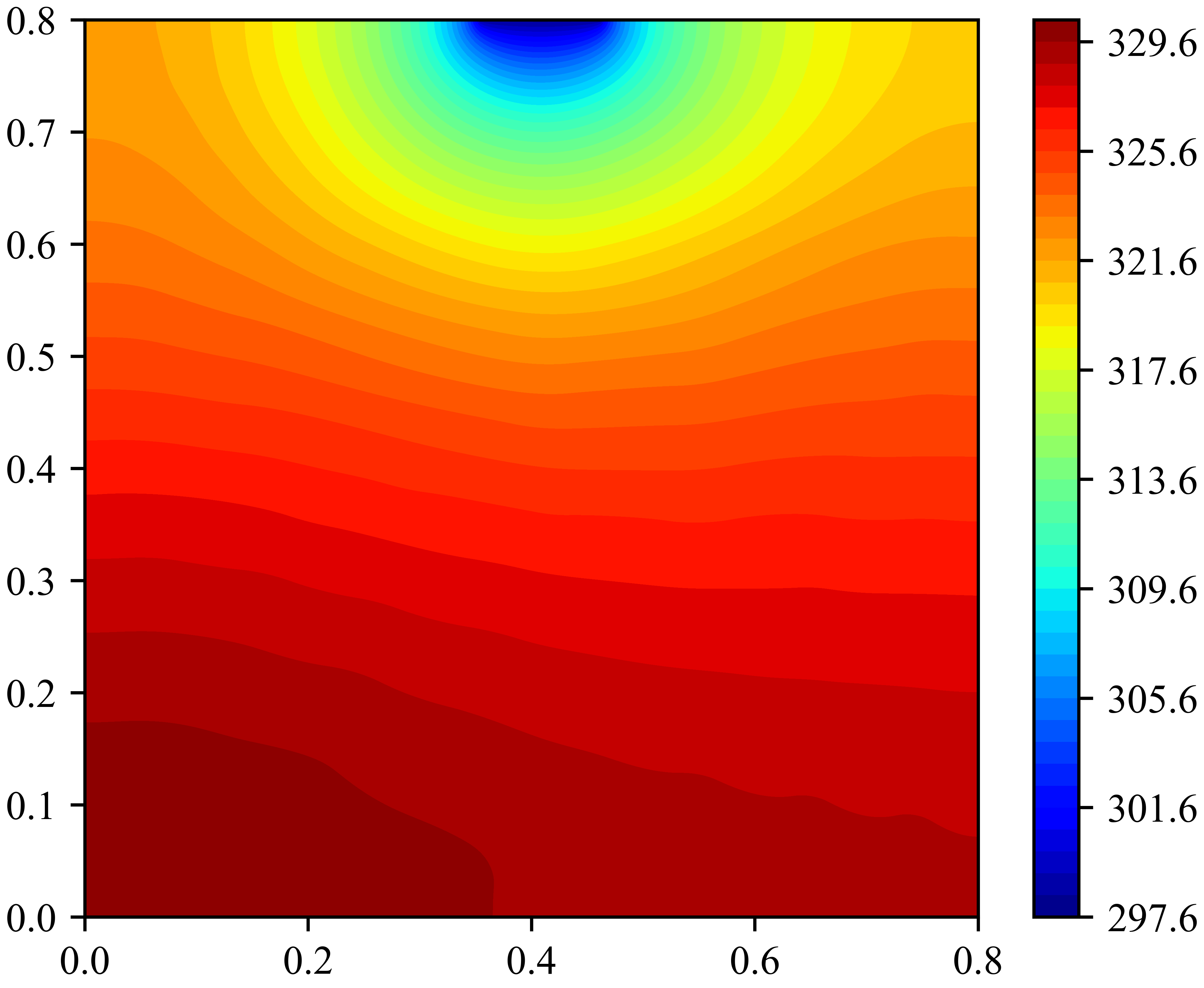}\label{truth_case2}}
	\subfigure [Reconstructed temperature field]
	{\includegraphics[scale=0.4]{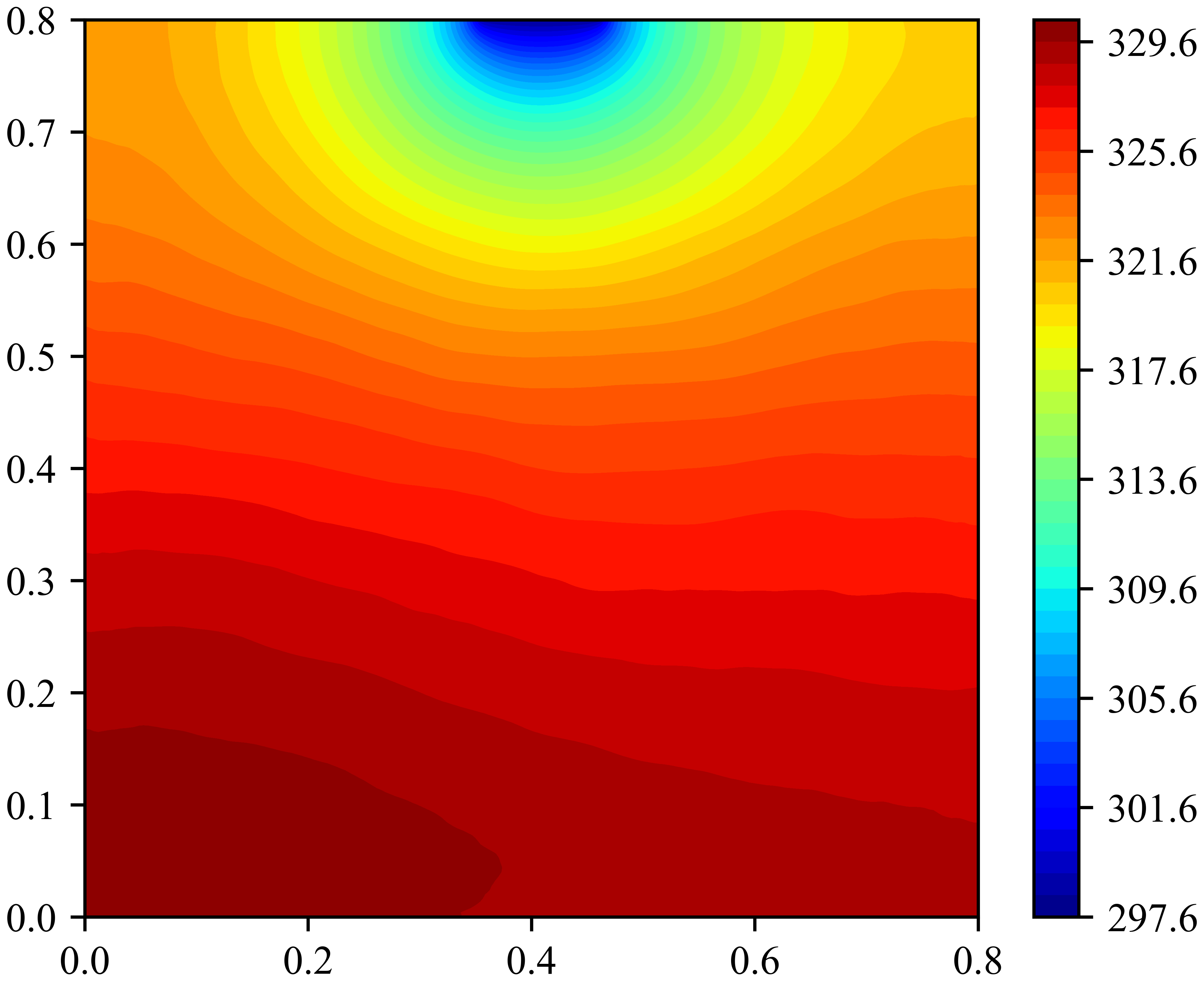}\label{reconstructed_case2}}
	\subfigure [Aleatoric uncertainty]
	{\includegraphics[scale=0.4]{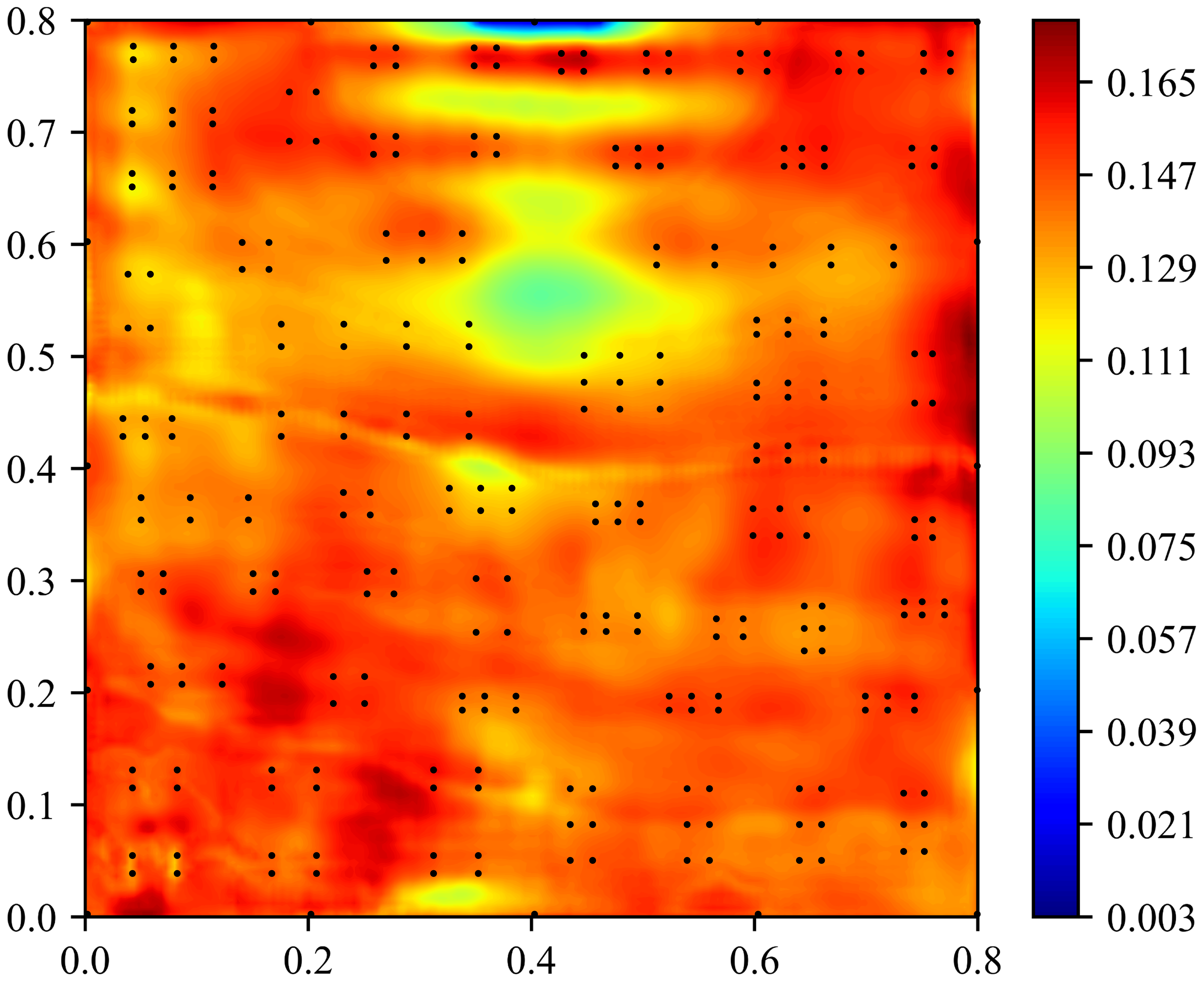}\label{alea_case2}}
	\caption{A MP temperature and its truth temperature field, reconstructed temperature field, and aleatoric uncertainty. The black points in (d) are the temperature monitoring points.}\label{test_case2}
\end{figure*}

\paragraph{\textbf{Predicted results analysis}} According to four indicators in section \ref{sec34}, three kinds of average errors $ \overline{RMSE} $, $ \overline{MAE} $ and $ \overline{MRE} $ are 0.0524, 0.0414, and 0.0001, respectively. Besides, the $\overline{R^2}$ value of the trained model is 0.9999. For the MP temperature in Fig.\ref{MP_case2}, its truth temperature field, reconstructed temperature field, and aleatoric uncertainty are shown in Fig.\ref{truth_case2}, Fig.\ref{reconstructed_case2}, and Fig.\ref{alea_case2}, respectively. Comparing Fig.\ref{truth_case2} and Fig.\ref{reconstructed_case2}, the reconstructed temperature field is basically consistent with the truth temperature field. For Fig.\ref{alea_case2}, the maximum value of quantified aleatoric uncertainty is approximately equal to 0.17. Thereby, the maximum difference between the quantified aleatoric uncertainty and the sensor measurement absolute deviation (0.15) is about 0.02. According to the necessity analysis of aleatoric uncertainty quantification in section \ref{sedc512}, the difference of 0.02 is mainly caused by the spread of data noise deviation. Besides, it is noteworthy that the quantified aleatoric uncertainties in the top heat sink area ($\Omega_{BC}$) and its surrounding non-monitoring point areas are almost equal to 0. Because the loss function (Eq.(\ref{PI_Loss})) of the training model uses boundary condition (Eq.(\ref{BC})), the temperature value of area $\Omega_{BC}$ is not affected by data noise. Therefore, the aleatoric uncertainties in area $\Omega_{BC}$ and its surrounding non-monitoring point areas are very small.

In summary, the proposed physics-informed Deep MC-QR method can construct an accurate surrogate model to reconstruct the temperature field of the satellite subsystem and can quantify the aleatoric uncertainty precisely. Thus, the reconstructed temperature field and the quantified aleatoric uncertainty are used to analyze the heat reliability of the satellite subsystem (Fig.\ref{Layout_SAT}) based on the interval multilevel BN in the following sections.

\subsubsection{Interval 4-level BN modeling}
\begin{figure*}[!htb]
	\centering
	\subfigure [Satellite component $C_{1}$]
	{\includegraphics[scale=0.6]{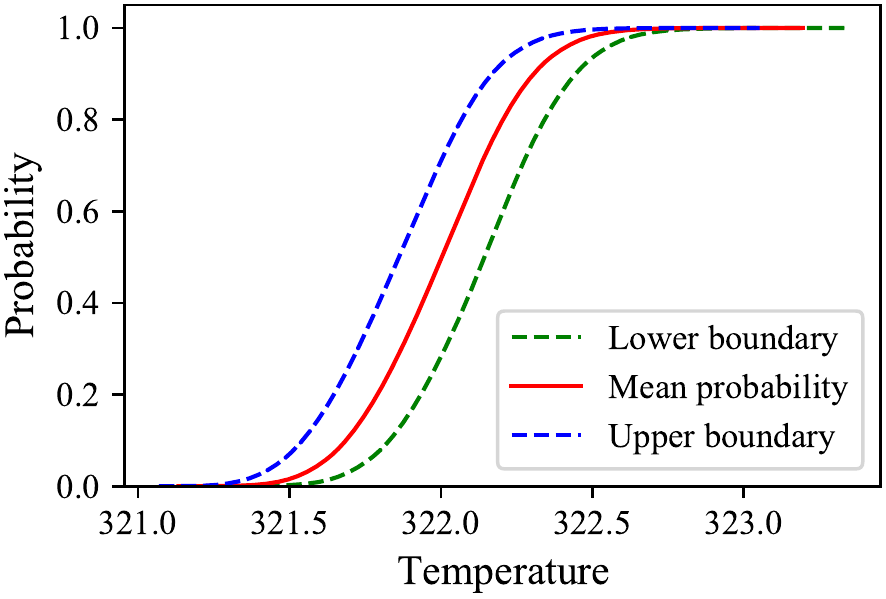}\label{cdf_com1}}
	\subfigure [Satellite component $C_{10}$]
	{\includegraphics[scale=0.6]{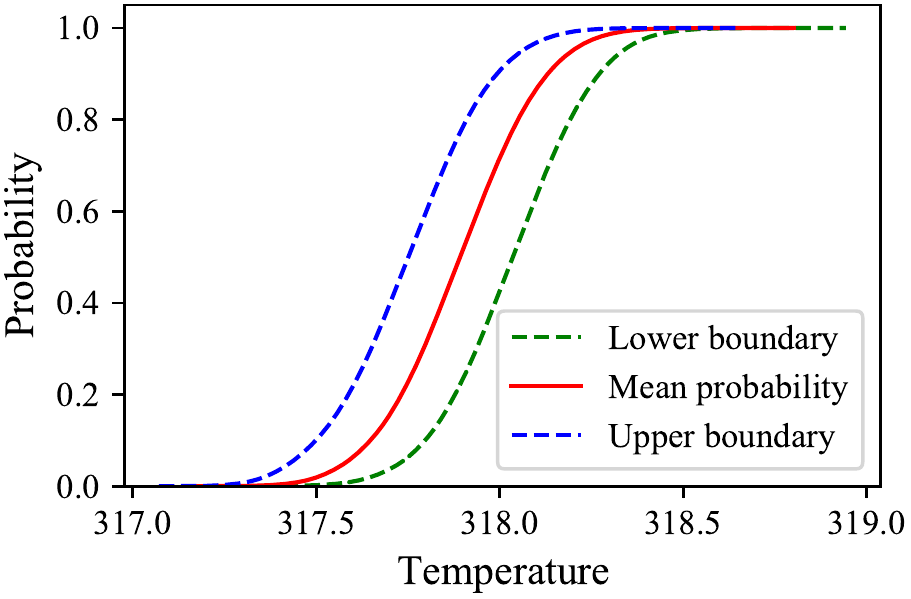}\label{cdf_com10}}
	\subfigure [Satellite component $C_{20}$]
	{\includegraphics[scale=0.6]{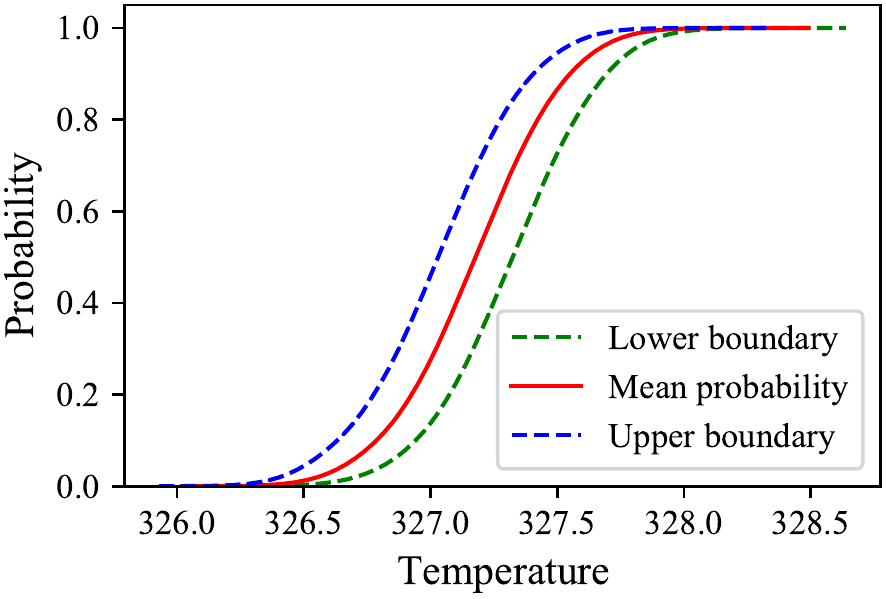}\label{cdf_com20}}
	\subfigure [Satellite component $C_{30}$]
	{\includegraphics[scale=0.6]{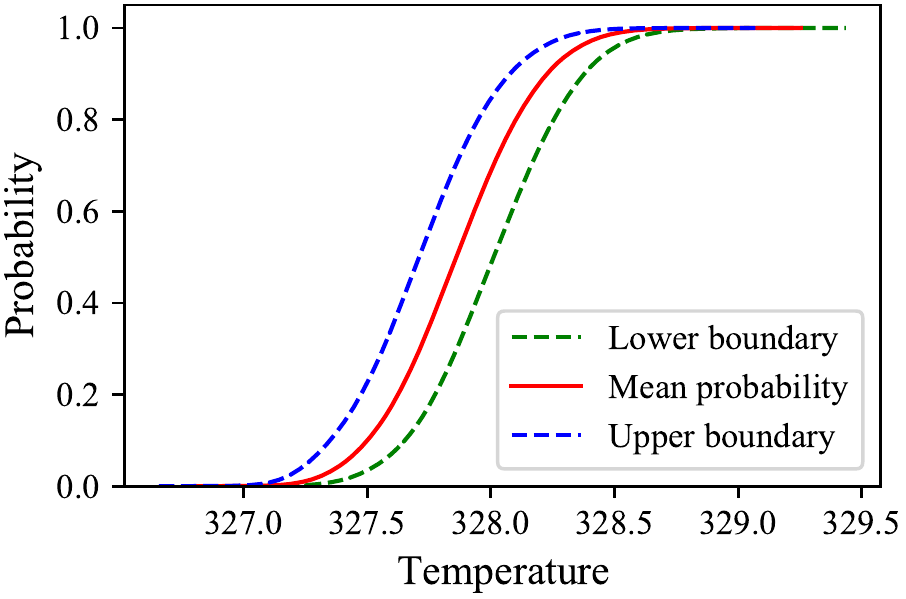}\label{cdf_com30}}
	\subfigure [Satellite component $C_{40}$]
	{\includegraphics[scale=0.6]{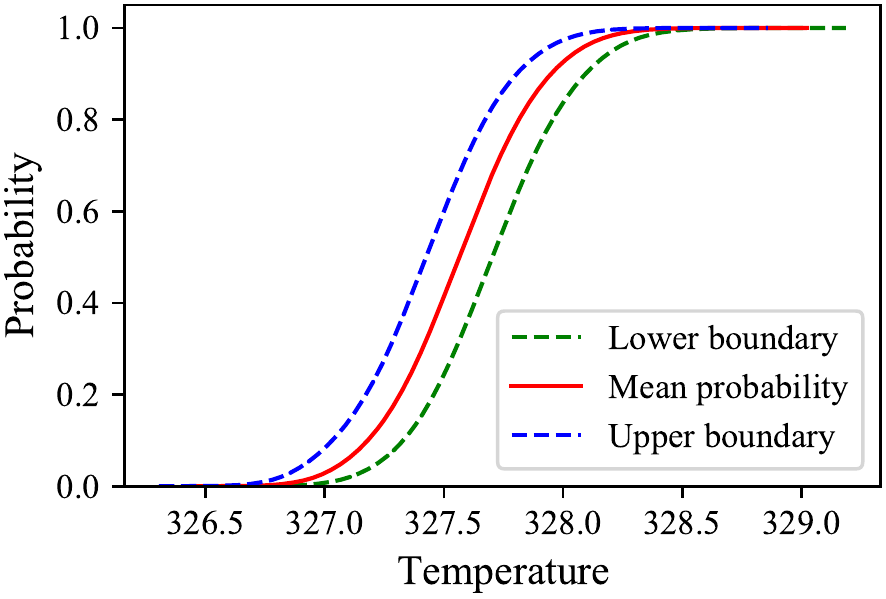}\label{cdf_com40}}
	\subfigure [Satellite component $C_{57}$]
	{\includegraphics[scale=0.6]{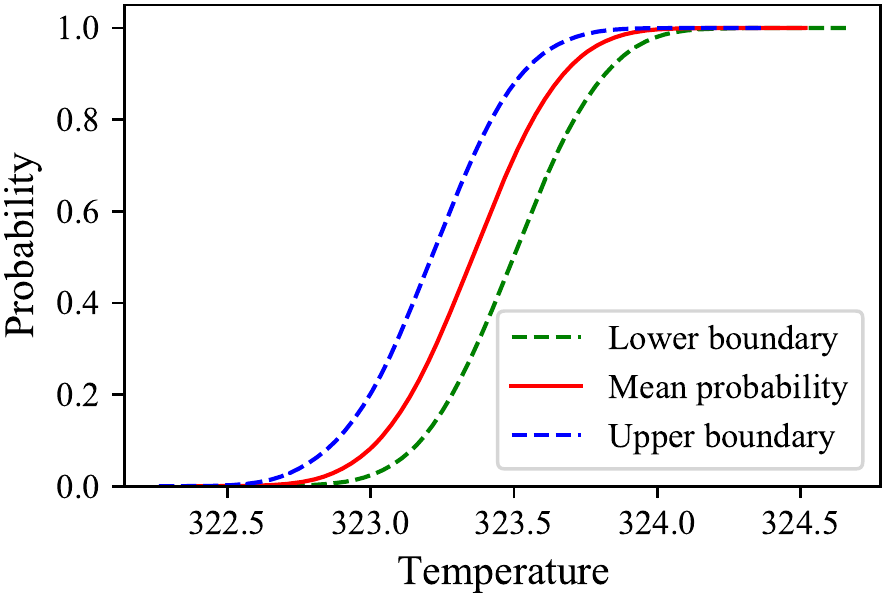}\label{cdf_com57}}
	\caption{The cumulative distribution functions of satellite components $C_{1}$, $C_{10}$, $C_{20}$, $C_{30}$, $C_{40}$, and $C_{57}$.}\label{cdf_case2}
\end{figure*}

For this case, the hyperparameter $\lambda$ is set to be 1.0, i.e., $\lambda=1.0$, and $1\times10^5$ discretized MP temperature images are used to estimate the probability distributions of satellite components, i.e., $N_{MCS}=1\times10^5$. Based on $1\times10^5$ reconstructed temperature fields, the cumulative distribution function of each satellite component can be estimated. For example, the cumulative distribution functions of satellite components $C_{1}$, $C_{10}$, $C_{20}$, $C_{30}$, $C_{40}$, and $C_{57}$ are shown in Fig.\ref{cdf_case2}. By \textbf{Algorithm} \ref{algorithm3}, the satellite components' normal probability intervals are estimated as shown in Table \ref{Com_prob_case2}. If the aleatoric uncertainty of the reconstructed temperature field is ignored, i.e., $\lambda=0$, the satellite components' normal probabilities $\left\{Pr_{\lambda=0}^{C_{s}=1}|s=1,2,\cdots,57\right\}$ are calculated based on the reconstructed temperature fields $\left\{{\hat{\bm{T}}}_{pre}^{m}|m=1,2,\cdots,10^5\right\}$ and the working state thresholds, as shown in Table \ref{Com_prob_case2}.

\begin{table}[!htb]
	\centering
	\caption{The estimated satellite components' (interval) normal probabilities in case 2.}
	\begin{tabular}{lll|lll|lll}
		\toprule
		S.C. & $Pr_{\lambda=0}^{C_{s}=1}$ & N.P. interval & S.C. & $Pr_{\lambda=0}^{C_{s}=1}$ & N.P. interval & S.C. & $Pr_{\lambda=0}^{C_{s}=1}$ & N.P. interval \\
		\midrule
		$C_{1}$  & 0.9944 & $\left[0.9746, 0.9991 \right]$ & $C_{20}$ & 0.9987 & $\left[ 0.9926, 0.9998 \right]$ & $C_{39}$ & 0.9758 & $\left[ 0.9246, 0.9942 \right]$ \\
		$C_{2}$  & 0.9888 & $\left[ 0.9550, 0.9981 \right]$ & $C_{21}$ & 0.9967 & $\left[ 0.9850, 0.9995 \right]$ & $C_{40}$ & 0.9976 & $\left[ 0.9893, 0.9997 \right]$ \\
		$C_{3}$  & 0.9987 & $\left[ 0.9913, 0.9998 \right]$ & $C_{22}$ & 0.9984 & $\left[ 0.9890, 0.9997 \right]$ & $C_{41}$ & 0.9846 & $\left[ 0.9507, 0.9965 \right]$ \\
		$C_{4}$  & 0.9973 & $\left[ 0.9819, 0.9996 \right]$ & $C_{23}$ & 0.9973 & $\left[ 0.9778, 0.9997 \right]$ & $C_{42}$ & 0.9778 & $\left[ 0.9363, 0.9945 \right]$ \\
		$C_{5}$  & 0.9966 & $\left[ 0.9764, 0.9996 \right]$ & $C_{24}$ & 0.9980 & $\left[ 0.9857, 0.9998 \right]$ & $C_{43}$ & 0.9959 & $\left[ 0.9836, 0.9994 \right]$ \\
		$C_{6}$  & 0.9907 & $\left[ 0.9437, 0.9990 \right]$ & $C_{25}$ & 0.9879 & $\left[ 0.9437, 0.9981 \right]$ & $C_{44}$ & 0.9950 & $\left[ 0.9800, 0.9992 \right]$ \\
		$C_{7}$  & 0.9924 & $\left[ 0.9399, 0.9994 \right]$ & $C_{26}$ & 0.9916 & $\left[ 0.9685, 0.9986 \right]$ & $C_{45}$ & 0.9948 & $\left[ 0.9800, 0.9991 \right]$ \\
		$C_{8}$  & 0.9985 & $\left[ 0.9666, 0.9999 \right]$ & $C_{27}$ & 0.9975 & $\left[ 0.9850, 0.9998 \right]$ & $C_{46}$ & 0.9903 & $\left[ 0.9681, 0.9979 \right]$ \\
		$C_{9}$  & 0.9959 & $\left[ 0.9599, 0.9998 \right]$ & $C_{28}$ & 0.9980 & $\left[ 0.9885, 0.9997 \right]$ & $C_{47}$ & 0.9959 & $\left[ 0.9820, 0.9993 \right]$ \\
		$C_{10}$ & 0.9870 & $\left[ 0.9267, 0.9987 \right]$ & $C_{29}$ & 0.9911 & $\left[ 0.9644, 0.9983 \right]$ & $C_{48}$ & 0.9957 & $\left[ 0.9824, 0.9993 \right]$ \\
		$C_{11}$ & 0.9967 & $\left[ 0.9410, 0.9999 \right]$ & $C_{30}$ & 0.9957 & $\left[ 0.9818, 0.9993 \right]$ & $C_{49}$ & 0.9940 & $\left[ 0.9780, 0.9987 \right]$ \\
		$C_{12}$ & 0.9960 & $\left[ 0.9657, 0.9997 \right]$ & $C_{31}$ & 0.9930 & $\left[ 0.9759, 0.9986 \right]$ & $C_{50}$ & 0.9903 & $\left[ 0.9676, 0.9978 \right]$ \\
		$C_{13}$ & 0.9868 & $\left[ 0.9516, 0.9977 \right]$ & $C_{32}$ & 0.9915 & $\left[ 0.9699, 0.9980 \right]$ & $C_{51}$ & 0.9906 & $\left[ 0.9671, 0.9982 \right]$ \\
		$C_{14}$ & 0.9912 & $\left[ 0.9616, 0.9986 \right]$ & $C_{33}$ & 0.9952 & $\left[ 0.9833, 0.9991 \right]$ & $C_{52}$ & 0.9884 & $\left[ 0.9642, 0.9972 \right]$ \\
		$C_{15}$ & 0.9917 & $\left[ 0.9684, 0.9981 \right]$ & $C_{34}$ & 0.9968 & $\left[ 0.9865, 0.9994 \right]$ & $C_{53}$ & 0.9722 & $\left[ 0.9273, 0.9908 \right]$ \\
		$C_{16}$ & 0.9906 & $\left[ 0.9504, 0.9990 \right]$ & $C_{35}$ & 0.9978 & $\left[ 0.9904, 0.9997 \right]$ & $C_{54}$ & 0.9881 & $\left[ 0.9568, 0.9979 \right]$ \\
		$C_{17}$ & 0.9959 & $\left[ 0.9827, 0.9993 \right]$ & $C_{36}$ & 0.9915 & $\left[ 0.9711, 0.9981 \right]$ & $C_{55}$ & 0.9966 & $\left[ 0.9858, 0.9991 \right]$ \\
		$C_{18}$ & 0.9944 & $\left[ 0.9788, 0.9990 \right]$ & $C_{37}$ & 0.9943 & $\left[ 0.9768, 0.9991 \right]$ & $C_{56}$ & 0.9975 & $\left[ 0.9881, 0.9996 \right]$ \\
		$C_{19}$ & 0.9895 & $\left[ 0.9625, 0.9981 \right]$ & $C_{38}$ & 0.9951 & $\left[ 0.9703, 0.9995 \right]$ & $C_{57}$ & 0.9969 & $\left[ 0.9816, 0.9997 \right]$ \\
		\bottomrule
	\end{tabular}
	\label{Com_prob_case2}
\end{table}

\begin{figure*}[!htbp]
	\centering
	{\includegraphics[scale=0.65]{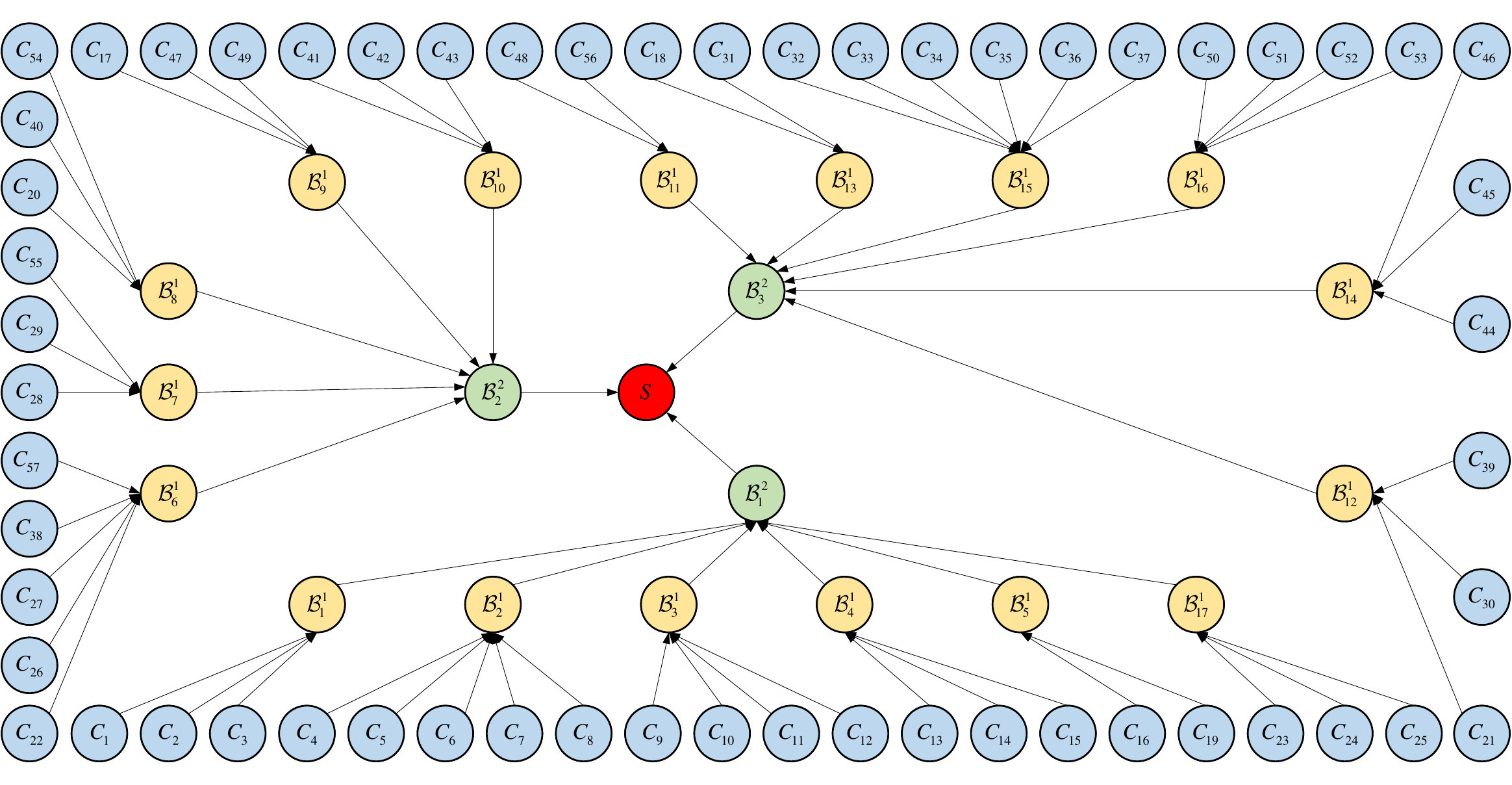}}
	\caption{Interval 4-level BN for satellite system heat reliability analysis.}
	\label{Case2_BN}
\end{figure*}

According to the logical relationships between satellite components, blocks, and satellite subsystem in Fig.\ref{Case2_com_frame}, this case model an interval 4-level BN for satellite system heat reliability analysis as shown in Fig.\ref{Case2_BN}. The interval probability distributions of the root nodes are shown in Table \ref{Com_prob_case2}. For the child nodes, their CPTs can be obtained by the series or parallel relationships. For example, the CPT of a child node with parallel parent nodes is shown in Table \ref{Pr_pa_CPT}. Thereby, the satellite subsystem reliability interval can be inferred based on the interval 4-level BN.

\subsubsection{Heat reliability analysis}
Inferred by the interval 4-level BN, the normal probability intervals of blocks $\mathcal{B}_1^2$, $\mathcal{B}_2^2$ and $\mathcal{B}_3^2$ are shown in Table \ref{block_prob_case2}, and the satellite subsystem interval heat reliability $\widetilde{R}$ is $\left[0.99416447, 0.99999952\right]$, i.e., $\widetilde{R}=\left[0.99416447, 0.99999952\right]$. If the aleatoric uncertainty of the reconstructed temperature field is ignored ($\lambda=0$), the satellite subsystem heat reliability $R_{\lambda=0}$ is 0.99991071, i.e., $R_{\lambda=0}=0.99991071$. Apparently, $R_{\lambda=0}\in\widetilde{R}$. Suppose that the engineers use the result $R_{\lambda=0}=0.99991071$ to evaluate the satellite subsystem heat reliability. This may lead to the following problems. If the temperature measured by the sensor is a relatively large truth temperature, the result $R_{\lambda=0}=0.99991071$ will be less than the truth heat reliability, i.e., underestimating the satellite subsystem heat reliability. Otherwise, the result $R_{\lambda=0}=0.99991071$ will be more than the truth heat reliability, i.e., overestimating the satellite subsystem heat reliability. However, the above problems will not occur when engineers use $\widetilde{R}=\left[0.99416447, 0.99999952\right]$ to evaluate the satellite subsystem heat reliability. Besides, $\widetilde{R}=\left[0.99416447, 0.99999952\right]$ contains the lowest and highest heat reliability of the satellite subsystem. Therefore, the interval heat reliability $\widetilde{R}$ can provide engineers with more reliable information to design this satellite subsystem.

\begin{table}[!htb]
	\centering
	\caption{The normal probability intervals of blocks $\mathcal{B}_1^2$, $\mathcal{B}_1^2$ and $\mathcal{B}_1^2$ in case 2.}
	\begin{tabular}{cccc}
		\toprule
		Block & $\mathcal{B}_1^2$ & $\mathcal{B}_2^2$ & $\mathcal{B}_2^2$ \\	
		\midrule
		N.P. interval & $\left[0.8320, 0.9947 \right]$ & $\left[ 0.7968, 0.9920 \right]$ & $\left[ 0.8291, 0.9894 \right]$ \\
		\bottomrule
	\end{tabular}
	\label{block_prob_case2}
\end{table}

As shown in Table \ref{relia_case2}, this case studies the influence of the block $\mathcal{B}_1^2$ ( or $\mathcal{B}_2^2$, or $\mathcal{B}_3^2$) failure on satellite subsystem (interval) heat reliability $R^E_{\lambda=0}$ ($\widetilde{R}^E$). For $\lambda=0$, the heat reliability $R^E_{\lambda=0}$ under any piece of evidence in Table \ref{relia_case2} is less than $R_{\lambda=0}=0.99991071$. Compared with $\widetilde{R}=\left[0.99416447, 0.99999952\right]$, the failure of block $\mathcal{B}_1^2$ ( or $\mathcal{B}_2^2$, or $\mathcal{B}_3^2$) reduces the satellite subsystem interval heat reliability $\widetilde{R}^E$. For one thing, according to the interval number size comparison formula \cite{Nakahara1992}, the order of three interval numbers in Table \ref{relia_case2} is $\left[0.96527106, 0.99991542 \right] < \left[ 0.96585417, 0.99995774 \right] < \left[ 0.97128373, 0.99994403 \right]$, which means that the failure of block $\mathcal{B}_2^2$ has the lowest impact on the satellite subsystem heat reliability. For another thing, judging from the values of $R^E_{\lambda=0}$, the satellite subsystem heat reliability is least affected by the failure of block $\mathcal{B}_3^2$. Apparently, the above two conclusions are contradictory. Detailed comparison shows that $R^{E\left(\mathcal{B}_2^2=0\right)}_{\lambda=0} = 0.99818277$ is very close to $R^{E\left(\mathcal{B}_3^2=0\right)}_{\lambda=0} = 0.99822152$. Due to the presence of data noise, the above second conclusion may not be robust. Thus, it is more believable to use interval numbers to analyze the influence of block failure on satellite subsystem heat reliability.

\begin{table}[!htb]
	\centering
	\caption{The satellite subsystem (interval) heat reliability $R^E_{\lambda=0}$ ($\widetilde{R}^E$) under different evidences in case 2.}
	\begin{tabular}{cccc}
		\toprule
 		Evidence & $\mathcal{B}_1^2=0$ & $\mathcal{B}_2^2=0$ & $\mathcal{B}_3^2=0$ \\	
		 \midrule
		 $R^E_{\lambda=0}$ & 0.99753356 & 0.99818277 & 0.99822152 \\
		 $\widetilde{R}^E$ & $\left[ 0.96527106, 0.99991542 \right]$ & $\left[0.97128373, 0.99994403 \right]$ & $\left[0.96585417, 0.99995774 \right]$ \\
		\bottomrule
	\end{tabular}
	\label{relia_case2}
\end{table}

\section{Conclusions}
This paper proposes the physics-informed Deep MC-QR method for reconstructing temperature field and quantifying the aleatoric uncertainty. For one thing, the proposed method combines a deep convolutional neural network with the known physics knowledge to reconstruct an accurate temperature field using some unlabeled monitoring point temperatures. For another thing, the proposed method can quantify the aleatoric uncertainty by the Monte Carlo quantile regression. Based on the reconstructed temperature field and the quantified aleatoric uncertainty, the normal probability interval of each satellite component is obtained by implementing the Monte Carlo simulation on the trained physics-informed Deep MC-QR model. Thereby, this paper models an interval multilevel BN to analyze satellite heat reliability. Two case studies are used for verifying the effectiveness of the proposed methods. The results of the first case study show that the proposed physics-informed Deep MC-QR method can accurately reconstruct the temperature field and precisely quantify the aleatoric uncertainty caused by data noise. The second case study validates that interval heat reliability can provide engineers with more reliable information to design a satellite subsystem.

In summary, this paper has the following three innovations: (1) The proposed physics-informed Deep MC-QR method can reconstruct an accurate temperature field using only monitoring point temperatures combined with the known physics knowledge; (2) Based on the Monte Carlo quantile regression, the proposed physics-informed Deep MC-QR method can precisely quantify the aleatoric uncertainty caused by data noise; (3) The interval multilevel BN can infer the satellite heat reliability using the reconstructed temperature field and the quantified aleatoric uncertainty. Generally, the positions of monitoring points are very important for the TFR problem. Therefore, the authors will explore how to choose the monitoring points' positions to improve further the accuracy of the reconstructed temperature field in future research.

\section*{Acknowledgments}
This work was supported by the Postgraduate Scientific Research Innovation Project of Hunan Province (No.CX20200006) and the National Natural Science Foundation of China (Nos.11725211).

\appendix
\setcounter{table}{0}
\section{Parallel block's normal probability interval}\label{appendix_A}
\setcounter{table}{0}
For the parallel block $ \mathcal{B}_{pa} $ in Fig.\ref{Block}, the state combinations of all components $\left\{{{C}_{\stackrel\frown{s}}}|{\stackrel\frown{s}}=1,2,\cdots,\stackrel\frown{n}\right\}$ are determined by the format in these references \cite{Zheng2019, Zheng2020}. Therefore, the CPT $ \Pr \left( \mathcal{B}_{pa}|{C_{1}},{C_{2}},\cdots ,{C_{\stackrel\frown{n}}} \right) $ of parallel block $ \mathcal{B}_{pa} $ is shown in Table \ref{Pr_pa_CPT}.
 
\begin{table}[htbp]	
	\centering	
	\caption{The CPT $ \Pr \left( \mathcal{B}_{pa}|{C_{1}},{C_{2}},\cdots ,{C_{\stackrel\frown{n}}} \right) $ of the parallel block $ \mathcal{B}_{pa} $.}	
	\label{Pr_pa_CPT}	
	\begin{tabular}{lllllcc}
		\toprule
		\multirow{2}{*}{$C_{1}$} & \multirow{2}{*}{$C_{2}$} & \multirow{2}{*}{$\cdots$} & \multirow{2}{*}{${C_{\stackrel\frown{n}-1}}$} & \multirow{2}{*}{$C_{\stackrel\frown{n}}$} & \multicolumn{2}{c}{$ \Pr \left( \mathcal{B}_{pa}|{C_{1}},{C_{2}},\cdots ,{C_{\stackrel\frown{n}}} \right) $} \\
		\cmidrule(r){6-7}		
		& & & & & $\mathcal{B}_{pa}=0$ & $\mathcal{B}_{pa}=1$   \\		
		\midrule		
		0	&0 &$\cdots$ &0 &0	&1	&0 \\
		0	&0 &$\cdots$ &0 &1	&0	&1 \\
		0	&0 &$\cdots$ &1 &0	&0	&1 \\
		0	&0 &$\cdots$ &1 &1	&0	&1 \\
		$\vdots$ &$\vdots$ &$\vdots$ &$\vdots$ &$\vdots$	&$\vdots$	&$\vdots$ \\
		1	&1 &$\cdots$ &0 &0	&0 &1 \\
		1	&1 &$\cdots$ &0 &1	&0 &1 \\
		1	&1 &$\cdots$ &1 &0	&0 &1 \\
		1	&1 &$\cdots$ &1 &1	&0 &1 \\
		\bottomrule		
	\end{tabular}	
\end{table}

Refer to Table \ref{Pr_pa_CPT}, the normal conditional probability $ \Pr \left( \mathcal{B}_{pa}=1|{C_{1}},{C_{2}},\cdots ,{C_{\stackrel\frown{n}}} \right) $ is denoted as
\begin{equation}\label{Pr_pa_CPT_1}
\Pr \left( \mathcal{B}_{pa}=1|{C_{1}},{C_{2}},\cdots ,{C_{\stackrel\frown{n}}} \right)=\left\{0, \underbrace{1,\cdots 1}_{{{2}^{\stackrel\frown{n}}}-1}\right\}.
\end{equation}
For the parallel block $ \mathcal{B}_{pa} $ in Fig.\ref{Block}, the calculation formulas ($ {\stackrel\frown{n}}=2 $ and $ {\stackrel\frown{n}}\ge3 $) of the normal probability interval $ \widetilde{\Pr}\left(\mathcal{B}_{pa}=1\right) $ are presented as follows.

If $ {\stackrel\frown{n}}=2 $,
\begin{equation}\label{Pr_B_pa_1_c2}
\begin{aligned}
& \widetilde{\Pr} \left( {{\mathcal{B}}_{pa}}=1 \right)=\widetilde{\Pr} \left( {{C}_{1}}=0 \right)\widetilde{\Pr} \left( {{C}_{2}}=1 \right) + \widetilde{\Pr} \left( {{C}_{1}}=1 \right) \\
& \qquad\qquad\quad\;\, =\left[1-\widetilde{\Pr} \left( {{C}_{1}}=1 \right)\right]\widetilde{\Pr} \left( {{C}_{2}}=1 \right) + \widetilde{\Pr} \left( {{C}_{1}}=1 \right) \\
& \qquad\qquad\quad\;\, =  \left[1-\widetilde{\Pr} \left( {{C}_{1}}=1 \right)\right]\widetilde{\Pr} \left( {{C}_{2}}=1 \right) - \left[1-\widetilde{\Pr} \left( {{C}_{1}}=1 \right)\right] + 1 \\
& \qquad\qquad\quad\;\, = 1-\left[1-\widetilde{\Pr} \left( {{C}_{1}}=1 \right)\right]\left[1-\widetilde{\Pr} \left( {{C}_{2}}=1 \right)\right] \\
& \qquad\qquad\quad\;\, = 1-\prod\limits_{{\mathcal{K}}=1}^{2}{\left[1-\widetilde{\Pr} \left( {{C}_{\mathcal{K}}}=1 \right)\right]}.
\end{aligned}
\end{equation}

If $ {\stackrel\frown{n}}\ge 3 $, the normal probability interval $ \widetilde{\Pr}\left(\mathcal{B}_{pa}=1\right) $ of the block $ \mathcal{B}_{pa} $ (Fig.\ref{Block}) is calculated by VE, i.e.,
\begin{equation}\label{Pr_B_pa_1}
\begin{aligned}
& \widetilde{\Pr} \left( {{\mathcal{B}}_{pa}}=1 \right)=\sum\limits_{{{C}_{1}}=1}^{2}{\widetilde{\Pr} \left( {{C}_{1}} \right)\cdots \sum\limits_{{{C}_{{\stackrel\frown{n}}-1}}=1}^{2}{\widetilde{\Pr} \left( {{C}_{{\stackrel\frown{n}}-1}} \right)\sum\limits_{{{C}_{\stackrel\frown{n}}}=1}^{2}{\widetilde{\Pr} \left( {{C}_{\stackrel\frown{n}}} \right){\Pr} \left( {{\mathcal{B}}_{pa}}=1|{{C}_{1}},{{C}_{2}},\cdots ,{{C}_{\stackrel\frown{n}}} \right)}}} \\ 
& \qquad\qquad\quad\;\;\;\; \vdots  \\ 
& \qquad\qquad\quad\;\; =\sum\limits_{{{C}_{1}}=1}^{2}{\widetilde{\Pr} \left( {{C}_{1}} \right)\cdots \sum\limits_{{{C}_{\stackrel\frown{s}}}=1}^{2}{\widetilde{\Pr} \left( {{C}_{\stackrel\frown{s}}} \right)\widetilde{\Pr} \left( {{\mathcal{B}}_{pa}}=1|{{C}_{1}},{{C}_{2}},\cdots ,{{C}_{\stackrel\frown{s}}} \right)}}, \\ 
\end{aligned}
\end{equation}
where
\begin{equation}\label{Pr_Bpa_1_Cs}
\widetilde{\Pr} \left( {{\mathcal{B}}_{pa}}=1|{{C}_{1}},{{C}_{2}},\cdots ,{{C}_{\stackrel\frown{s}}} \right)=\left\{ \mathcal{A}_{\stackrel\frown{s}}^{pa}, \underbrace{1,\cdots ,1}_{{{2}^{\stackrel\frown{s}}}-1} \right\},
\end{equation}
and 
\begin{equation}\label{As_pa}
\begin{aligned}
& \mathcal{A}_{\stackrel\frown{s}}^{pa}=\sum\limits_{\beta =\stackrel\frown{s}+1}^{\stackrel\frown{n}-\stackrel\frown{s}}{\left[ \widetilde{\Pr} \left( {{C}_{\beta+1 }}=1 \right)\left( \prod\limits_{\mathcal{K}=\stackrel\frown{s}+1}^{\beta}{\left[1-\widetilde{\Pr} \left( {{C}_{\mathcal{K}}}=1 \right)\right]} \right) \right]}+\widetilde{\Pr} \left( {{C}_{\stackrel\frown{s}+1}}=1 \right). \\
\end{aligned}
\end{equation}

By eliminating nodes $\left\{{{C}_{1}},{{C}_{2}},\cdots ,{{C}_{\stackrel\frown{s}}}\right\}$, Eq.(\ref{Pr_B_pa_1}) can be further converted to
\begin{equation}\label{Pr_B_pa_1_simpled}
\begin{aligned}
& \widetilde{\Pr} \left( {{\mathcal{B}}_{pa}}=1 \right)=\sum\limits_{{{C}_{1}}=1}^{2}{\widetilde{\Pr} \left( {{C}_{1}} \right)\cdots \sum\limits_{{{C}_{\stackrel\frown{s}}}=1}^{2}{\widetilde{\Pr} \left( {{C}_{\stackrel\frown{s}}} \right)\widetilde{\Pr} \left( {{\mathcal{B}}_{pa}}=1|{{C}_{1}},{{C}_{2}},\cdots ,{{C}_{\stackrel\frown{s}}} \right)}} \\ 
& \qquad\qquad\quad\;\;\;\; \vdots  \\ 
& \qquad\qquad\quad\;\;=\sum\limits_{{{C}_{1}}=1}^{2}{\widetilde{\Pr} \left( {{C}_{1}} \right)\widetilde{\Pr} \left( {{\mathcal{B}}_{pa}}=1|{{C}_{1}} \right)} \\ 
& \qquad\qquad\quad\;\;=\left[1-\widetilde{\Pr} \left( {{C}_{1}}=1 \right)\right]\mathcal{A}_{1}^{pa}+\widetilde{\Pr} \left( {{C}_{1}}=1 \right) \\
&\qquad\qquad\quad\;\; = \sum\limits_{\beta =1}^{\stackrel\frown{n}-1}{\left\{ \widetilde{\Pr} \left( {{C}_{\beta + 1}}=1 \right) \prod\limits_{\mathcal{K}=1}^{\beta }{\left[1-\widetilde{\Pr} \left( {{C}_{\mathcal{K}}}=1 \right)\right]}\right\}} +\widetilde{\Pr} \left( {{C}_{1}}=1 \right),
\end{aligned}
\end{equation}
where
\begin{equation}\label{Pr_pa_c1_1}
\widetilde{\Pr} \left( {{\mathcal{B}}_{pa}}=1|{{C}_{1}} \right)=\left\{ \mathcal{A}_{1}^{pa},1 \right\},
\end{equation}
and
\begin{equation}\label{A1_pa}
\begin{aligned}
& \mathcal{A}_{1}^{pa}=\sum\limits_{\beta =2}^{\stackrel\frown{n}-1}{\left\{ \widetilde{\Pr} \left( {{C}_{\beta + 1}}=1 \right) \prod\limits_{\mathcal{K}=2}^{\beta }{\left[1-\widetilde{\Pr} \left( {{C}_{\mathcal{K}}}=1 \right)\right]}\right\}} +\widetilde{\Pr} \left( {{C}_{2}}=1 \right). \\
\end{aligned}
\end{equation}

Then, Eq.(\ref{Pr_B_pa_1_simpled}) can be further converted to
\begin{equation}\label{Pr_B_pa_1_simpled_1}
\begin{aligned}
& \widetilde{\Pr} \left( {{\mathcal{B}}_{pa}}=1 \right)=\sum\limits_{\beta =1}^{\stackrel\frown{n}-1}{\left\{ \widetilde{\Pr} \left( {{C}_{\beta + 1}}=1 \right) \prod\limits_{\mathcal{K}=1}^{\beta }{\left[1-\widetilde{\Pr} \left( {{C}_{\mathcal{K}}}=1 \right)\right]}\right\}} +\widetilde{\Pr} \left( {{C}_{1}}=1 \right) \\
& \qquad\qquad\quad\;\, = \sum\limits_{\beta =2}^{\stackrel\frown{n}-1}{\left\{ \widetilde{\Pr} \left( {{C}_{\beta + 1}}=1 \right) \prod\limits_{\mathcal{K}=1}^{\beta }{\left[1-\widetilde{\Pr} \left( {{C}_{\mathcal{K}}}=1 \right)\right]}\right\}} \\
& \qquad\qquad\quad\;\;\;\;\;\,+\widetilde{\Pr} \left( {{C}_{2}}=1 \right)\left[1-\widetilde{\Pr} \left( {{C}_{1}}=1 \right)\right] -\left[1-\widetilde{\Pr} \left( {{C}_{1}}=1 \right)\right] + 1 \\
& \qquad\qquad\quad\;\, = \sum\limits_{\beta =2}^{\stackrel\frown{n}-1}{\left\{ \widetilde{\Pr} \left( {{C}_{\beta + 1}}=1 \right) \prod\limits_{\mathcal{K}=1}^{\beta }{\left[1-\widetilde{\Pr} \left( {{C}_{\mathcal{K}}}=1 \right)\right]}\right\}} - \prod\limits_{\mathcal{K}=1}^{2 }{\left[1-\widetilde{\Pr} \left( {{C}_{\mathcal{K}}}=1 \right)\right]} +1 \\
& \qquad\qquad\quad\;\;\;\; \vdots \\
& \qquad\qquad\quad\;\, = \sum\limits_{\beta =\stackrel\frown{s}}^{\stackrel\frown{n}-1}{\left\{ \widetilde{\Pr} \left( {{C}_{\beta + 1}}=1 \right) \prod\limits_{\mathcal{K}=1}^{\beta }{\left[1-\widetilde{\Pr} \left( {{C}_{\mathcal{K}}}=1 \right)\right]}\right\}} - \prod\limits_{\mathcal{K}=1}^{\stackrel\frown{s} }{\left[1-\widetilde{\Pr} \left( {{C}_{\mathcal{K}}}=1 \right)\right]} +1 \\
& \qquad\qquad\quad\;\, = \sum\limits_{\beta =\stackrel\frown{s}+1}^{\stackrel\frown{n}-1}{\left\{ \widetilde{\Pr} \left( {{C}_{\beta + 1}}=1 \right) \prod\limits_{\mathcal{K}=1}^{\beta }{\left[1-\widetilde{\Pr} \left( {{C}_{\mathcal{K}}}=1 \right)\right]}\right\}} \\
& \qquad\qquad\quad\;\;\;\;\;\, +  \widetilde{\Pr} \left( {{C}_{\stackrel\frown{s}+1}}=1 \right) \prod\limits_{\mathcal{K}=1}^{\stackrel\frown{s} }{\left[1-\widetilde{\Pr} \left( {{C}_{\mathcal{K}}}=1 \right)\right]} - \prod\limits_{\mathcal{K}=1}^{\stackrel\frown{s} }{\left[1-\widetilde{\Pr} \left( {{C}_{\mathcal{K}}}=1 \right)\right]} +1 \\
& \qquad\qquad\quad\;\, = \sum\limits_{\beta =\stackrel\frown{s}+1}^{\stackrel\frown{n}-1}{\left\{ \widetilde{\Pr} \left( {{C}_{\beta + 1}}=1 \right) \prod\limits_{\mathcal{K}=1}^{\beta }{\left[1-\widetilde{\Pr} \left( {{C}_{\mathcal{K}}}=1 \right)\right]}\right\}} - \prod\limits_{\mathcal{K}=1}^{\stackrel\frown{s}+1}{\left[1-\widetilde{\Pr} \left( {{C}_{\mathcal{K}}}=1 \right)\right]} +1 \\
& \qquad\qquad\quad\;\;\;\; \vdots \\
& \qquad\qquad\quad\;\, =  \widetilde{\Pr} \left( {{C}_{\stackrel\frown{n}}}=1 \right) \prod\limits_{\mathcal{K}=1}^{\stackrel\frown{n}-1}{\left[1-\widetilde{\Pr} \left( {{C}_{\mathcal{K}}}=1 \right)\right]} - \prod\limits_{\mathcal{K}=1}^{\stackrel\frown{n}-1}{\left[1-\widetilde{\Pr} \left( {{C}_{\mathcal{K}}}=1 \right)\right]} +1 \\
& \qquad\qquad\quad\;\, =  1 - \prod\limits_{\mathcal{K}=1}^{\stackrel\frown{n}}{\left[1-\widetilde{\Pr} \left( {{C}_{\mathcal{K}}}=1 \right)\right]}.
\end{aligned}
\end{equation}

In summary, the normal probability interval $ \widetilde{\Pr}\left(\mathcal{B}_{pa}=1\right) $ of the parallel block $ \mathcal{B}_{pa} $ ($ {\stackrel\frown{n}}\ge2 $) is calculated by 
\begin{equation}\label{Pr_B_pa_1_diff}
\widetilde{\Pr} \left( {{\mathcal{B}}_{pa}}=1 \right) = 1 - \prod\limits_{\mathcal{K}=1}^{\stackrel\frown{n}}{\left[1-\widetilde{\Pr} \left( {{C}_{\mathcal{K}}}=1 \right)\right]}.
\end{equation}

\bibliography{mybibfile}

\end{document}